\documentclass{article}

\PassOptionsToPackage{numbers, compress}{natbib}
\usepackage{microtype}
\usepackage{graphicx}
\usepackage{overpic}
\usepackage{subfig}
\usepackage{booktabs} 

\usepackage{tikz}
\usepackage{tikz-cd}
\usepackage{caption}
\usepackage{soul}
\usepackage{enumitem}
\usepackage{amsmath,mathtools}
\usepackage{amssymb}
\usepackage{amsthm}
\PassOptionsToPackage{table,dvipsnames}{xcolor}
\usepackage{xcolor}
\usepackage{colortbl}
\definecolor{RoyalBlue}{rgb}{0.254, 0.41, 0.882}

\usepackage{hyperref}

\usepackage[capitalize,noabbrev]{cleveref}

\hypersetup{
    colorlinks=true,
    linkcolor=RoyalBlue,
    citecolor=RoyalBlue,
    urlcolor=RoyalBlue
}

\usepackage{multirow} 

\usepackage{iac_pkg}
\usepackage{lipsum}
\usepackage{bbm}
\usepackage{xspace}
\usepackage{stmaryrd}

\usepackage{algorithm}
\usepackage{algorithmic}
\usepackage{varwidth}

\usepackage{wrapfig}
\usepackage{placeins}

\theoremstyle{plain}

\theoremstyle{definition}

\theoremstyle{remark}

\definecolor{lightred}{RGB}{180,57,65}
\definecolor{cell_bisque}{rgb}{1.0, 0.89, 0.77}
\definecolor{cell_blond}{rgb}{0.98, 0.94, 0.75}
\definecolor{cell_blue}{RGB}{155, 187, 228}
\definecolor{princetonorange}{rgb}{1.0, 0.56, 0.0}
\definecolor{pinkpearl}{rgb}{0.91, 0.67, 0.81}
\definecolor{mossgreen}{RGB}{145, 174, 129}
\definecolor{parula_1}{rgb}{0.2422,0.1504,0.6603}
\definecolor{parula_2}{rgb}{0.2803, 0.2782, 0.9221}
\definecolor{parula_3}{rgb}{0.2440, 0.4358, 0.9988}
\definecolor{parula_4}{rgb}{0.1540, 0.5902, 0.9218}
\definecolor{parula_5}{rgb}{0.0297, 0.7082, 0.8163}
\definecolor{parula_6}{rgb}{0.1938, 0.7758, 0.6251}
\definecolor{parula_7}{rgb}{0.5044, 0.7993, 0.3480}
\definecolor{parula_8}{rgb}{0.8634, 0.7406, 0.1596}
\definecolor{parula_9}{rgb}{0.9892, 0.8136, 0.1885}
\definecolor{parula_10}{rgb}{0.9769, 0.9839, 0.0805}
\definecolor{headerbg}{RGB}{230, 230, 250} 
\definecolor{headerbg}{gray}{0.95}
\definecolor{rowbg}{RGB}{245, 245, 255}   
\usepackage[textsize=footnotesize]{todonotes}

\usepackage[preprint]{neurips_2025}

\DeclareRobustCommand{\fwdarrow}{%
  \tikz[baseline=-0.3ex]{\draw[->, thick, black, opacity=0.7, >=stealth] (0,0) -- (1em,0);}%
}
\DeclareRobustCommand{\revarrow}{%
  \tikz[baseline=-0.3ex]{\draw[dashed,->, thick, black, opacity=0.7, >=stealth] (0,0) -- (1em,0);}%
}
\usepackage[utf8]{inputenc} 
\usepackage[T1]{fontenc}    
\usepackage{url}            
\usepackage{amsfonts}       
\usepackage{nicefrac}       

 \title{What is Adversarial Training for Diffusion Models?}

\author{%
  Maria Rosaria Briglia$^\dagger$\\
  \And
  Mujtaba Hussain Mirza$^\dagger$
  \And
  Giuseppe Lisanti$^{\star}$\\
  \And
  Iacopo Masi$^\dagger$\\
}

\begin{document}

\maketitle
{\vspace{-20pt}
\centering
 \methodimagetag{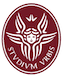}~~Sapienza University of Rome   $^{\dagger}$~~ \methodimagetag{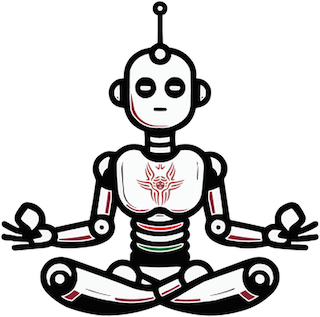} \href{https://omnai.di.uniroma1.it}{OmnAI Lab}$^{\dagger}$
\methodimagetag{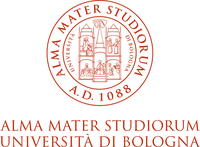}~~University of Bologna $^{\star}$~ 

}{\centering

}

\begin{abstract}
We answer the question in the title showing that adversarial training (AT) for diffusion models (DMs) fundamentally differs from classifiers: while AT in classifiers enforces output \emph{invariance}, AT in DMs requires \emph{equivariance} to keep the diffusion process aligned with the data distribution.
We define AT as a way to enforce smoothness in the diffusion flow, improving robustness to outliers and corrupted data.
Unlike prior art, our method makes no assumptions on the noise model and integrates seamlessly into diffusion training by adding either random noise--similar to randomized smoothing--or adversarial noise--akin to AT. 
This enables intrinsic capabilities such as handling noisy data, dealing with extreme variability such as outliers, preventing memorization, and obviously improving robustness.
We rigorously evaluate our approach with proof-of-concept datasets with \emph{known} distributions in low- and high-dimensional space, thereby taking a perfect measure of errors; we further evaluate on standard benchmarks such as CIFAR-10, CelebA and LSUN Bedroom, showing strong performance under severe noise, data corruption and iterative adversarial attacks. Code is available at \href{https://github.com/OmnAI-Lab/Adversarial-Training-DM.git}{github.com/OmnAI-Lab/Adversarial-Training-DM}.
\end{abstract}    
\section{Introduction}\label{sec:intro}

When scaling up Diffusion Model (DM) training,  dealing with noisy data is unavoidable.
Large-scale datasets, currently the key of AI success, often contain different forms of noise. These include \textit{inlier noise}, i.e., small perturbations within the expected distribution, \textit{outlier noise}, where samples deviate significantly from it, \textit{missing or corrupted data}, often affected by Gaussian noise, and even \textit{adversarial noise}, such as that introduced by poisoning attacks~\cite{tian2022comprehensive}.
Despite recent promising attempts in training models under noisy conditions~\cite{daras2024consistent,daras2024ambient,daras2024much},  these are often limited by specific noise assumptions:  \cite{daras2024consistent} requires the exact variance of the added Gaussian noise, \cite{daras2024ambient} focuses solely on missing data and \cite{daras2024much} assumes knowledge of clean and noisy data. A more general method, with fewer assumptions, remains an open challenge.
\begin{figure*}[tb]
  \centering
    \begin{overpic}[width=\textwidth]{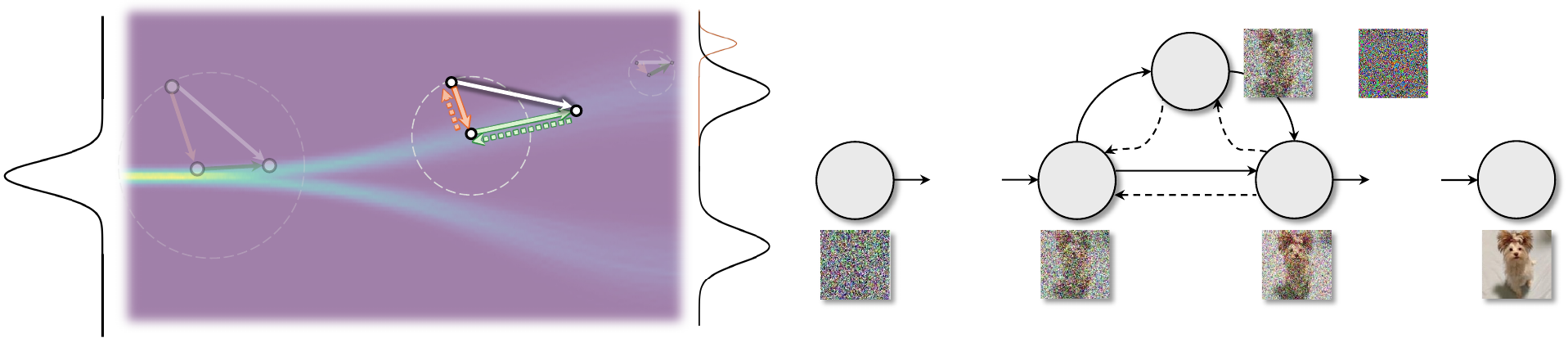} 
    \put(47.5,19){\resizebox{28pt}{!}{$p_{\text{noise}}(\bx_0)$}}
    \put(49,14){$p_{\text{data}}(\bx_0)$}
    \put(0,13){\resizebox{20pt}{!}{$q(\rvx_T)$}}
    \put(53,10){\resizebox{11pt}{!}{$\rvx_T$}}
    \put(60,10){...}
    \put(90,10){...}
    \put(84.85,17){\resizebox{5pt}{!}{$+$}}
    \put(67,10){$\rvx_t$}
    \put(80.5,10){\resizebox{16pt}{!}{$\rvx_{\scriptstyle{t-1}}$}}
    \put(95.5,10){$\rvx_{0}$}
    \put(73.85,16.85){\resizebox{16pt}{!}{$\rvx_{t}{\scriptstyle+}\pert$}}
    \put(30,11){\resizebox{8pt}{!}{$\rvx_t$}}
    \put(26,17){\resizebox{17pt}{!}{$\rvx_{t}{\scriptstyle+}\pert$}}
    \put(37.5,14.5){\resizebox{16pt}{!}{$\rvx_{t-1}$}}
    \put(32.25,11.75){\resizebox{21pt}{!}{$\bepsilon_{\theta}(\rvx_t,t)$}}
    \put(27,13){\resizebox{5pt}{!}{$\pert$}}
  \end{overpic}
  \caption{\tbf{Inducing smoothness into diffusion trajectories.} 
We train the denoising network to follow the \textcolor{mossgreen}{score function \ie, $\rvx_{t} \mapsto  \rvx_{t-1}$ using just $\bepsilon_{\theta}(\rvx_t,t)$}, \textcolor{lightred}{but we also perturb locally the data point as $\rvx_{t}{\scriptstyle+}\pert$ inside a $\ell_p$ ball centered on $\rvx_t$} and then imposing equivariance: $\rvx_{t}{\scriptstyle+}\pert \mapsto \bepsilon_{\theta}(\rvx_t,t)+\pert \triangleq \rvx_{t-1}.$ This equals  to adding an intermediate step in the Markov Chain, behaving as an additional denoising step in the training, making the model resilient to possible outliers or noise in the dataset---$p_{\text{noise}}(\bx_0)$---not proper of $p_{\text{data}}(\bx_0)$.  The local noising step can be implemented as adversarial~\cite{goodfellow2014explaining} or as random, akin to randomized smoothing~\cite{cohen2019certified}. Perturbation strength is adaptive, large in the noise phase and it shrinks in the content phase. 
  \fwdarrow{} indicates the forward process; \revarrow{} the reverse process.
  }
  
\label{fig:fig1}
\end{figure*}
Additionally, DMs are prone to memorizing training data, as demonstrated in studies on AI safety and red teaming~\cite{jagielskimeasuring,somepalli2023diffusion,carlini2023extracting}. 
While DMs surpassed GANs generative capabilities and addressed their stability problems, like mode collapse and mode coverage~\cite{zhong2019rethinking}, they have also introduced  memorization and information leakage issues, particularly in case of over-parametrized models.
Mode coverage must further deal with minor modes that do not belong to the data manifold, which should be discarded to avoid spurious data correlations.
To  address these challenges and achieve a balance between mode coverage and memorization, we propose smoothing the DM's trajectory space, as in \cref{fig:fig1}. Thereby, our contributions are as follows:
\begin{itemize}[itemsep=2pt, leftmargin=*]
\item Despite a few papers applied AT to DMs~\cite{yang2024structure,sauer2024adversarial}, none have formally defined AT in this context,  discussed its practical implications and shed light on its benefits, i.e. denoising the data distribution.
We are the first to reconnect AT to denoising, linking this technique to prior works~\cite{daras2024ambient,daras2024consistent,daras2024much}.

\item Inspired by~\cite{zhang2019theoretically} , we develop an AT algorithm tailored for DMs. We show that differently than classifiers, while they require enforcing \emph{invariance} in AT, score-based models require \emph{equivariance} to properly learn the data distribution, as formalized in our key finding in \cref{eq:ours}.

\item We empirically demonstrate our method's flexibility in handling noisy data, facing extreme variability like outliers, preventing memorization, and improving robustness.
We show experiments in low (3D) and in high-dimensional settings with known distributions. Following ~\cite{daras2024consistent,daras2024much} we use Gaussian noise to perturb the data. 
We validate our results on real datasets such as CIFAR-10~\cite{krizhevsky2009learning}, CelebA~\cite{liu2015faceattributes}, and LSUN Bedroom \cite{yu15lsun}, under heavy corruption, achieving strong performance.
\end{itemize}

\section{Adversarial training smooths the diffusion flow}\label{sec:method}
\subsection{Preliminaries}\label{sec:preliminaries}
Diffusion Models (DMs)~\cite{ho2020denoising} aim to learn a data distribution, $p_{\text{data}}(\rvx)$ by noising data with a fixed procedure, mapping them to $\gN(\vzero, \rmI)$ using a Markov Chain $q(\rvx_T,\ldots,\rvx_1|\rvx_0)=\prod_{t=1}^T q(\rvx_t|\rvx_{t-1})$, where, given a noisy input $\rvx_{t-1}$, the next state $\rvx_t$ is reached through  the following gaussian transition: 
 \begin{equation}
q(\rvx_{t}|\rvx_{t-1})=\gN\big(\rvx_t;\sqrt{1-\sigma(t)}\rvx_{t-1},\sigma(t)\rmI\big),
\end{equation}
 $\sigma(t)$ is the noise scheduler: a monotonically decreasing time-varying function chosen s.t.  $\sigma(0) = \sigma_{\text{min}}$, $\sigma(T) = \sigma_{\text{max}}$ and  $0<\sigma_{\text{min}}<\sigma_{\text{max}}<1$. 
The generation is achieved with a learnable ``decoding step'' that reverts data from noise estimation $p(\rvx_{t-1}|\rvx_t)$. 
If the noise scheduler is chosen carefully to take small noising steps, then the approximation $q(\rvx_T | \rvx_0) \approx \gN(\vzero, \rmI)$ and the following equation hold:
\begin{equation*}
q(\rvx_t | \rvx_0) = \gN\big(\rvx_t;\sqrt{\alpha_t}\rvx_{t-1},(1-\alpha_t)\rmI\big)~~\text{where}~~\alpha_t \doteq \prod_{s=1}^t 1-\sigma(t)
\end{equation*}
This means we can encode directly from $\rvx_0 \mapsto \rvx_t$ as:
 \begin{equation}
\rvx_t = \sqrt{\alpha_t}\rvx_0 + \sqrt{1-\alpha_t}~\bepsilon ~~\text{where}~~\bepsilon \sim \gN(\vzero, \rmI).
\label{eq:encode}
\end{equation}
The process is analyzed in~\cite{song2021scorebased} as denoising score matching, following the Stochastic Differential Equation (SDE)   $\rvx_t = \mathbf{f}(\rvx_t, t) t + g(t)  \mathbf{w}_t $,
where $\mathbf{w}$ is the standard Wiener process, $\mathbf{f}(\cdot, t): \mathbb{R}^d \rightarrow \mathbb{R}^d$ is the drift coefficient, and $g(\cdot): \mathbb{R}\rightarrow \mathbb{R}$  is the diffusion coefficient. 
Generation is performed by solving the probability flow ODE (PF-ODE),  from $t=T$ to $0$ and starting from $\rvx_T \sim \gN (0, \sigma^2_\text{max}I)$, whose solution is learned from the DM.
For a given $\rvx_{0}$, the simplified version of training objective $\Loss_\text{DM}$  reported in \cite{ho2020denoising} is thus defined as:
\begin{equation}
\label{eq:LDM}
\begin{aligned}
\Loss_\text{DM}&= \E_{\substack{\bepsilon \sim \gN(\vzero, \rmI)\\ t\sim \mathcal{U}(\vzero, \rmI)}}\biggl[ \norm{\bepsilon-\bepsilon_{\theta}\big(\rvx_t(\rvx_0,\bepsilon),t\big)}_2^2\biggl]
\end{aligned}
\end{equation}
whose objective is to infer the noise  $\bepsilon$ applied to the initial image, ensuring that the starting point $\rvx_0 $ is correctly reconstructed, enabling the model---the denoising network $\bepsilon_{\theta}$---to correctly generate in-distribution data during inference.
For inference we solve the SDE using $\bepsilon_{\theta}$ and the recurrency:
\begin{equation}
   \rvx_{t-1}(\theta) =  \frac{1}{\sqrt{1-\sigma(t)}}\left(\rvx_t(\theta)-\frac{\sigma(t)}{\sqrt{1-\alpha_t}} \bepsilon_\theta\big(\rvx_t(\theta), t\big)\right) + \sigma(t) \bz, ~ \bz \sim \gN(\vzero, \rmI),\forall t \in [0,\cdots, T].
   \label{eq:inference}
\end{equation}

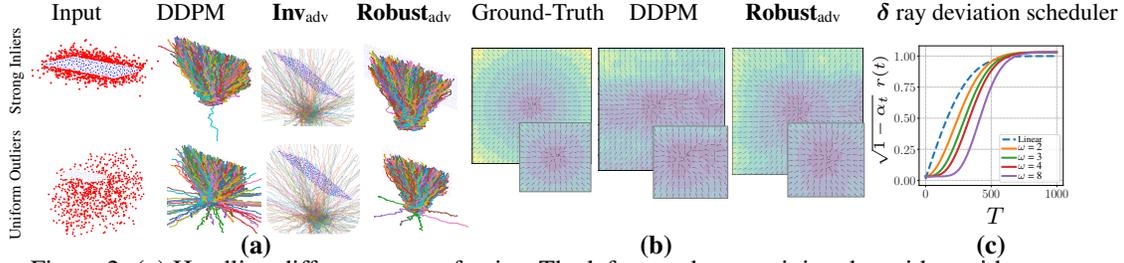
\begin{figure}[t]
    \centering
    \begin{overpic}[width=\textwidth]{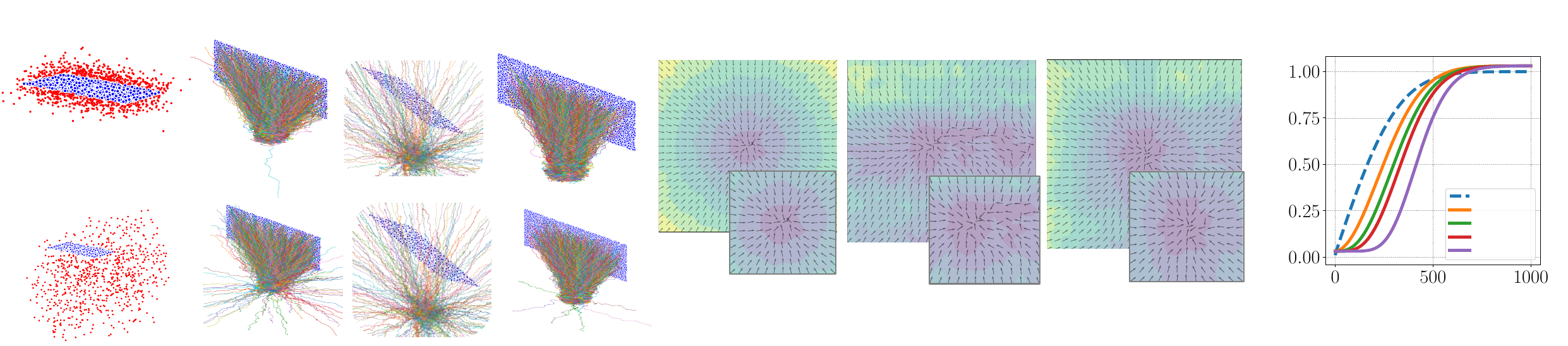}
    \put(2,21){{\small Input}}
    \put(12,21){{\small DDPM}}
     \put(23,21){{\small \textbf{Inv}$_\text{adv}$}}
    \put(31,21){{\small \robust{adv}}}
    
    \put(-2,0){\rotatebox{90}{\tiny{Uniform Outliers}}}
    \put(-2,12){\rotatebox{90}{\tiny{Strong Inliers}}}
    \put(42,21){{\small Ground-Truth}}
    \put(57,21){{\small DDPM}}
    \put(68,21){{\small \textbf{Robust}$_\text{adv}$}}
    \put(20,-1.25){\textbf{(a)}}\put(58,-1.25){\textbf{(b)}}\put(90,-1.25){\textbf{(c)}}
    \put(80.5,21){{\small$\pert$ ray deviation scheduler}}
    \put(80,7){\rotatebox{90}{{\tiny$\sqrt{1-\alpha_t}~~r(t)$}}}
    \put(91,1.5){{\footnotesize $T$}}
    \put(94,9.35){\resizebox{9pt}{!}{Linear}}
    \put(94,8.55){\resizebox{9pt}{!}{$\omega$~=~2}}
    \put(94,7.65){\resizebox{9pt}{!}{$\omega$~=~3}}
    \put(94,6.75){\resizebox{9pt}{!}{$\omega$~=~4}}
    \put(94,5.85){\resizebox{9pt}{!}{$\omega$~=~8}}
    \end{overpic}
    \caption{\textbf{(a)} Handling different types of noise. The leftmost shows training data either with strong inlier noise \emph{(top)} or uniform outliers \emph{(bottom)}. The trajectories reveal that DDPM~\cite{ho2020denoising} struggles with both, while if you train with invariance (\tbf{Inv}$_\text{adv}$) the process diverges. Instead ours (\robust{adv}) is more robust, avoiding diverging trajectories and better reaching the data centroid. 
    \tbf{(b)} Score vector fields: versors represent the score field, colormap shows magnitude, \textcolor{parula_1}{\rule{0.3cm}{0.15cm}} less \textcolor{parula_10}{\rule{0.3cm}{0.15cm}} more intense. \emph{(left)} Ground-truth \emph{(middle)} DDPM~\cite{ho2020denoising};  \emph{(right)} Our \robust{adv}. AT yields smoother, more consistent scores, better matching the data shape, while shrinking variability and increasing field intensity. \tbf{(c)} Adversarial perturbation ray. The curves vary with $\omega$, controlling the slope of $\sqrt{1-\alpha_t}~~r(t)$ to shorten the content phase and reduce the curve's steepness in DDPM.
    }
    \label{fig:fig2}
\end{figure}

\subsection{Motivation, ``in vitro'' experiments, and noise types}
\minisection{Motivation and overview} Adversarial samples for classifiers are the key idea towards robust models~\cite{goodfellow2014explaining,madry2017towards}. 
To ensure classifiers' invariant response to future adversarial perturbations, AT is designed to maintain consistency of the model's output.
Unlike classifiers,  DMs resolve a regression task, so the AT problem must be formulated differently. 
In our work, we propose the first principled approach to adversarially train DMs. 
As shown in \cref{fig:fig2}~(a) our methods can estimate the underlying distribution even in presence of strong inliers noise or uniform outliers, while invariance leads the process to diverge; moreover, given \cref{fig:fig2}~(b), it can smooth  DMs' scores leading to more stable fields.

\minisection{``In vitro'' experimental setup} We experiment on synthetic 3D datasets where we have the possibility to go from ``linear'' and unimodal datasets to more complex multi-modal. 
\texttt{oblique-plane} assumes $p_{\text{data}}$ lives approximately on a 2D subspace with equation $x+y+z=30$.
\texttt{3-gaussians}, is a multi-modal 3D Mixture of Gaussians distributions s.t.
$\frac{1}{3}\mathcal{N}([10,10,10],\sigma)+\frac{1}{3}\mathcal{N}([20,20,20],\sigma)+\frac{1}{3}\mathcal{N}([10,30,30],\sigma)$ and $\sigma=0.25$. 
Regarding high-dimensional data $\rvx \in \mathbb{R}^{32\times32\times3}$ we use the simple ``Smithsonian Butterflies'' dataset, consisting of aligned \texttt{butterflies} images resized to $32\times 32$ and then flattened, yet still in controlled setting. 
We linearize the data using principal components, fitting a $25$ dimensional subspace, embedded in a $3072$-D space, retaining 70\% of the variance as $\bx^{\prime} = \bsf{\mu} + \sum_i \lambda_i\bsf{\alpha}_i \mbf{U}_{i}$. 
Sampling stochasticity comes from $\bsf{\alpha}\sim \mathcal{N}(0;\sigma)$, while $\bsf{\mu}\in \mathbb{R}^{3072}$,   $\mbf{U}\in \mathbb{R}^{25\times 3072}$ and $\lambda_i$ are the mean, dataset's principal components and their singular values. 
Linearized data are visually similar to real ones; thus we discard the real data and train the DMs to fit $\{\bx^{\prime}\}_{i=1}^N$. 
This allows us to \textit{perfectly measure} distance between the generated samples and the linearized distribution, not relying on proxy metrics, such as FID, but instead measuring the closed-form reconstruction error between the subspace and the generations as $\rho=\norm{\rvx_0(\theta)- \mbf{U}\mbf{U}^\top \rvx_0(\theta)}$, where $\rvx_0(\theta)$ is the generation using \cref{eq:inference}. We also measure the Peak Signal-to-Noise Ratio (PSNR).

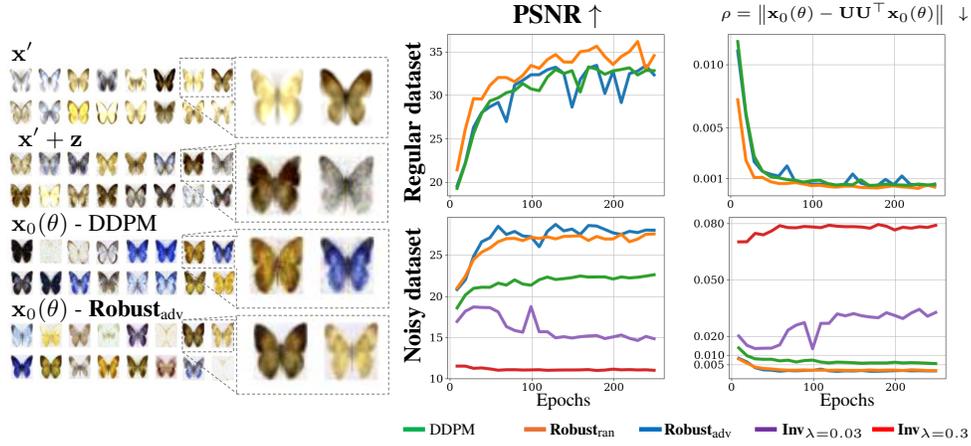
\begin{figure*}[tbh]
    \centering
    \begin{overpic}[trim=140 100 150 20, clip,width=0.9\textwidth]{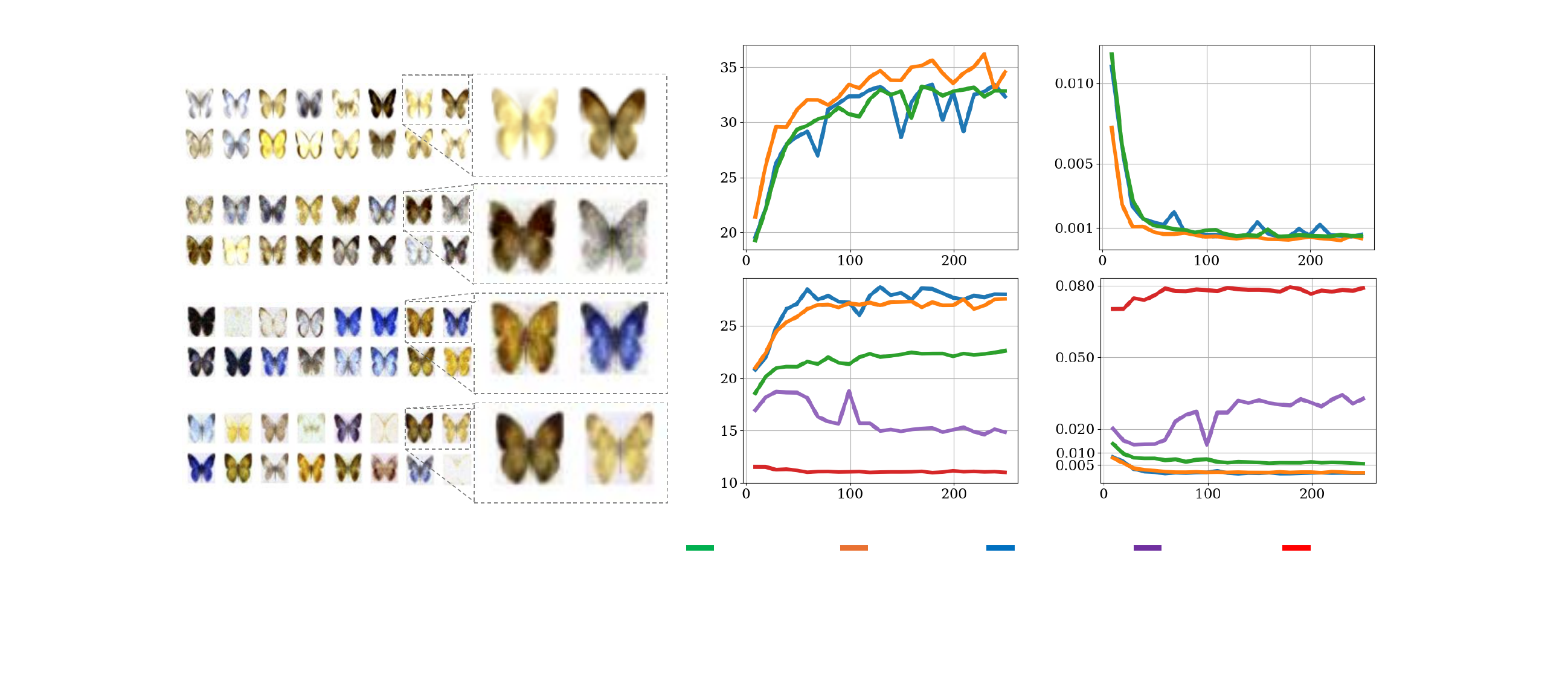}
    \put(54,43.5){{\textbf{PSNR $\uparrow$}}}
    \put(76,44){{\tiny\textbf{$\rho=\norm{\rvx_0(\theta)- \mbf{U}\mbf{U}^\top \rvx_0(\theta)}~~\downarrow$}}}
    \put(56.5,3){{\scriptsize Epochs}}
    \put(86.5,3){{\scriptsize Epochs}}
    \put(45.2,0){{\tiny DDPM}}
    \put(57.9,0){{\tiny \textbf{Robust}$_\text{ran}$}}
    \put(70,0){{\tiny \textbf{Robust}$_\text{adv}$}}
    \put(81.8,0){{ \tiny {\textbf{Inv}}$_{\lambda = 0.03}$}}
    \put(94.7,0){{\tiny {\textbf{Inv}}$_{\lambda = 0.3}$}}
    \put(42.5,24.5){\rotatebox{90}{{\small \tbf{Regular dataset}  }}}
    \put(42.5,7.5){\rotatebox{90}{{\small \tbf{Noisy dataset} }}}
    \put(1,39.5){{\small$\bx^{\prime}$}}
    \put(1,30.5){ {\small$\bx^{\prime} + \bz$}}
    \put(1,21.5){{\footnotesize $\rvx_0(\theta)$ - DDPM}}
    \put(1,12.5){{\footnotesize $\rvx_0(\theta)$ - \textbf{Robust}$_\text{adv}$} }
    \end{overpic}
    \caption{
    (\textit{left}) On the linearized \texttt{butterflies} dataset, we measure closed-form reconstruction error. From top to bottom: training data, corrupted data, DDPM generations, and \robust{adv} results.
    (\textit{right}) Metric plots: first column shows PSNR, second column the closed-form reconstruction error. 
    First row: clean data; second row: 90\% of data corrupted with Gaussian noise ($\sigma = 0.1$). 
    We also include an ablation on \textit{invariance regularization} with $\lambda=\{0.3$, $0.03\}$. Strong invariance prevents learning the distribution; reducing $\lambda$ helps but is still far below other methods.
    }
    \label{fig:plane-robust}
\end{figure*}

\minisection{Noise model tested} We experiment with different noise models: for 3D data we add inlier noise by increasing either $\sigma$ of Gaussians or $\bsf{\alpha}$ as $\bsf{\alpha}\sim \mathcal{N}(0;\sigma)$ in case of subspace $\bsf{\mu} + \sum_i \lambda_i\bsf{\alpha}_i \mbf{U}_{i}$. 
We then include outliers by adding strong noise in the ambient space: for 3D data embedding the original point cloud with dense, grid-like, uniform noise; for \texttt{butterflies} simply adding Gaussian noise once linearized as $\bx^{\prime}+\bz$ where $\bz \sim \gN(\vzero, \sigma\rmI)$. 
\cref{fig:fig2}~(a) and \cref{fig:plane-robust} show noise types and datasets.

\subsection{Injecting adversarial noise in the diffusion flow}
\minisection{Injecting additional noise in the trajectory space}
\label{minisec:diffusion-pert} 
Given that DM encoding process already perturbs the data with Gaussian noise, it is not trivial to add another perturbation. 
After extensive tests and research, we found out that it is necessary to craft the $\pert$ perturbation following these requirements: 
i) the ray of the perturbation, i.e. $r(t)=\norm{\pert(t)}_p$, has to be time-dependent following the noise scheduler $\sigma(t)$; 
ii) recalling the diffusion phases as defined in~\cite{choi2022perception}, the ray cannot be large in the content phase, otherwise the approach may merge two different modes of the data distribution yielding over smoothing; 
iii) when $t\to T$, thus we are close to pure noise, $r(t)$ can have a high value with a maximum of 1 to maintain the assumption mentioned in \cref{sec:preliminaries}; 
iv) finally, for $t\to 0$, $r(t)$ also have to tend to zero yet keeping a constant bias $\gamma$ at the end; if we do not do so, the approach may under smooth the data and the denoising will not occur. 
We then modify \cref{eq:encode} as:
\begin{equation}
\rvx_t = \sqrt{\alpha_t}\rvx_0 + \sqrt{1-\alpha_t}~(\bepsilon+\pert)~~\text{where}~~\pert \sim \mathcal{U}\big(-r_\beta(t), r_\beta(t)\big),~~r_\beta(t)\doteq \frac{(\sqrt{1-\alpha_t})^\omega+\gamma\cdot \beta}{\sqrt{1-\alpha_t}}
\label{eq:delta-encode}
\end{equation}
where $~\bepsilon \sim \gN(\vzero, \rmI)$ and $\omega \geq 1$ is an exponent to shrink the ray in the content phase and $\gamma$ is the bias to keep the ray at a minimum but not zero. The denominator in $r_\beta(t)$ is needed so that it simplifies with $\sqrt{1-\alpha_t}$ of \cref{eq:delta-encode}. $\beta$ is a scalar to simply increase the bias randomly, set to 1 by default. \cref{fig:fig2}~(c) shows the adversarial perturbation ray in function of time compared to the normal scheduler.

\minisection{Two types of perturbations: random vs adversarial} The additional noise to enforce smoothness can be of two types: random $\pert_{\text{ran}}$, akin randomized smoothing~\cite{cohen2019certified}, or adversarial $\pert_{\text{adv}}$, similar to AT~\cite{goodfellow2014explaining}.

\textit{{\small Random:}} The random noise applied to enforce smoothness is defined as $r_\beta(t)$ in \cref{eq:delta-encode} yet sampling $\beta \sim \mathcal{U}[0.5,2]$, a stochastic parameter included to randomize the ray so that the process is resilient to variability in the ray. $\pert_{\text{ran}}$ would then be a uniform variable whose standard deviation is 
$r_\beta(t) / \sqrt{3}$.

\textit{{\small Adversarial:}} In the adversarial setting, we employ the Fast Gradient Sign Method (FGSM) with a random start~\cite{kurakin2017atscale}. The perturbation is first initialized like $\pert_{\text{ran}}$, followed by a single FGSM step. The resulting perturbation is then projected back onto the $\ell_\infty$ ball of radius $r_\beta(t)$ to ensure $\|\pert_{\text{adv}}\|_{\infty} \leq r_\beta(t)$. The optimization $\pert_{\text{adv}}$ considers the following cost function:
\begin{equation}
 \mathcal{J_\theta}(\rvx_t, \pert,t ) = \norm{\bepsilon_{\theta}\big(\rvx_t+\pert,t\big)-\bepsilon_{\theta}\big(\rvx_t,t\big)}_2^2
\label{eq:attack}
\end{equation}
The final adversarial perturbation $\pert_{\text{adv}}$ is then computed by taking a step in the direction of the gradient of $\mathcal{J_\theta}$, followed by projection:
\begin{equation}
\pert_{\text{adv}} = \mathbb{P}_{r_\beta(t)}\bigg[\pert_{\text{ran}} 
 + \frac{r_\beta(t)}{\sqrt{3}} \mathcal{S}\big(\nabla_{\rvx_t} \mathcal{J_\theta}(\rvx_t, \pert_\text{ran},t) \big)\bigg].
\label{eq:delta-adv}
\end{equation}

where $\mathbb{P}_{r_\beta(t)}$ projects the adversarial perturbation onto the surface of $\rvx_t$'s neighbor $\ell_\infty$-ball , $\mathcal{S}$ is the sign operator and 
$r_\beta(t) / \sqrt{3}$ 
is the standard deviation of the attack.

\subsection{Adversarial training for diffusion models}

\minisection{Na\"{i}ve invariance does not work} Classic AT derived from classifiers enforces output invariance to perturbations. This objective, does not simply transfer to DMs since they solve a regressive task,  which must take into account input variations.
 \cref{eq:inv}, which applies classical AT invariance  concept, causes DMs to learn a different distribution than $p_{\text{data}}(\bx)$,
  misleading the generation to producing noisy data.
As an evidence, we provide qualitative
\begin{wrapfigure}{r}{0.45\linewidth}
\vspace{-12pt}
\begin{minipage}{\linewidth}
\footnotesize 
 \begin{equation}
    \mathcal{L} = \arg\min_{\theta}\norm{\bepsilon_{\theta}\big(\rvx_t+\pert,t\big)-\bepsilon_{\theta}\big(\rvx_t,t\big)}_2^2
     \label{eq:inv}
 \end{equation} 
\end{minipage}
\vspace{-12pt}
\end{wrapfigure}
 results of this behavior for \texttt{oblique-plane} and quantitative ones for \texttt{butterflies}.
\cref{fig:fig2}~(a) shows that, on \texttt{oblique-plane} dataset, invariance causes the model to learn a different distribution w.r.t. the real one---see \tbf{Inv}$_\text{adv}$ setting. Moreover, \cref{fig:plane-robust} (\textit{right, bottom row}) shows the PSNR and  reconstruction error when
\begin{wrapfigure}{r}{0.47\textwidth}
\vspace{-1.5em}
    \begin{minipage}{0.47\textwidth}
        \begin{algorithm}[H]
            \caption{AT for Diffusion Models}
            \label{alg:adv-training}
            \begin{algorithmic}
                \STATE {\bfseries Input:} dataset $\mathcal{D}$, model $\theta$, max timestep $T$, scheduler $\alpha_t$, strength $\lambda$, ray scheduler $r_\beta(t)$
                \REPEAT
                \STATE Sample $\rvx_0 \sim \mathcal{D}$, $\bepsilon \sim \mathcal{N}(\mathbf{0}, \mbf{I})$,\\
                $t \sim \mathcal{U}(\{0,\dots,T\})$, $\beta \sim \mathcal{U}[0.5,2]$,\\
                $\pert \sim \mathcal{U}[-r_\beta(t), r_\beta(t)]$
                \STATE $\rvx_t = \sqrt{\bar{\alpha}_t}\rvx_0 + \sqrt{1-\bar{\alpha}_t}\bepsilon$
                \STATE Compute $\pert_\text{adv}$ using~\cref{eq:delta-adv}
                \STATE $\rvx_t^\text{adv} = \sqrt{\bar{\alpha}_t}\rvx_0 + \sqrt{1-\bar{\alpha}_t}(\pert_\text{adv} + \bepsilon)$
                \STATE $\theta \leftarrow \theta - \eta \nabla_\theta \Loss_\text{AT}(\rvx_t, \rvx_t^\text{adv}, t, \bepsilon)$
                \UNTIL{convergence}
            \end{algorithmic}
        \end{algorithm}
    \end{minipage}
\vspace{-1.3em}
\end{wrapfigure}
 applying the invariance to \texttt{butterflies} noisy data ($90\%$ corrupted samples, $\sigma=0.1$):
the model is unable to properly reconstruct the data distribution. This claim is even more enforced when observing varying $\lambda$, as its value decreases the model is able to recover it back.
When moving on higher-dimensional data, invariance resulted in worse FID, being the model unable to generate points onto the data manifold, achieving FID is 356.9 on 50K generated samples from CIFAR-10 dataset.

\minisection{Key change is equivariance} Being the diffusion a regression task, we formulated again the AT taking into account the need for input sensitivity of the model by enforcing \emph{equivariance}. 
The intuition is depicted in the introductory~\cref{fig:fig1} and a theoretical discussion is given in \cref{sec:theory}. Since the forward process still needs to ``land'' in the data distribution, despite the additional perturbations $\pert$, the network must learn to denoise from $\rvx_t + \pert$ in such a way that the denoised point always matches the previous one in the chain $\rvx_{t-1}$. This objective is reached by taking into account $\pert$ in the AT loss as $\arg\min_{\theta}\norm{\bepsilon_{\theta}\big(\rvx_t+\pert,t\big)-[\bepsilon+\pert]}_2^2.$ While this equation enforces equivariance, it does not yet enforce smoothness, since two outputs of the network do not interact together.

\minisection{Our Training} Building on top of the previous setting, we propose our AT formulation, detailed in \cref{alg:adv-training} . The noisy sample $\rvx_t$ is defined as in \cref{eq:encode}, while the perturbed noisy sample $\rvx_t + \pert$ is defined as in \cref{eq:delta-encode}, considering either $\pert_{\text{ran}}$ or $\pert_\text{adv}$ perturbating elements. We have the regular term---$\mathcal{L_\text{DM}}$ as in \cref{{eq:LDM}}---where the method teaches the network to flow towards the data distribution using $\bepsilon$. Yet, we added our novel term $\mathcal{L_\text{reg}}$ that \emph{enforces equivariance and smoothness around the regular trajectory of the DM.}
Our final formulation is given in \cref{eq:ours}, where  $\lambda_t = \frac{\lambda \cdot \sqrt{3}}{\beta\cdot r(t)}$ is a time-dependent hyper-parameter defining regularization's strength. $\lambda_t$ depends on a constant $\lambda$ and is rescaled according to the norm of the perturbation through its standard deviation. \cref{eq:ours} is intended to be applied directly to the DDPM framework which is based on a $\bepsilon$-predicting network. 
\begin{equation}
\mathcal{L}_{\text{AT}}(\rvx_t,\rvx^\text{adv}_t, t,\bepsilon)=\arg\min_{\theta}\underbrace{\norm{\bepsilon_{\theta}\big(\rvx_t,t\big)-\bepsilon}_2^2}_{\mathcal{L_\text{DM}} \text{ to fit data distr.}}+ \underbrace{\lambda_t\norm{\bepsilon_{\theta}\big(\rvx_t^\text{adv},t\big)-[\bepsilon_{\theta}\big(\rvx_t,t\big)+\pert]}_2^2}_{\mathcal{L_\text{reg}} \text{ to enforce smoothness}}
    \label{eq:ours}    
\end{equation}

\section{Experimental results}\label{sec:expt}
\begin{figure}
    \centering

    \begin{minipage}[b]{\textwidth}
        \centering
    \begin{minipage}[b]{0.5\textwidth}
        \begin{overpic}[trim= 26.5 24.5 26.5 0, clip,width=\textwidth]{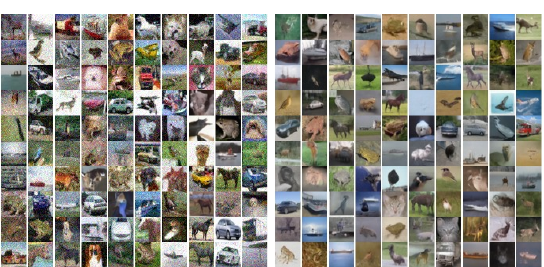}
            \put(4,48){DDPM~\cite{ho2020denoising}, $\sigma=0.2$}
            \put(58,48){\robust{adv}, $\sigma=0.2$}
             \end{overpic}
    \end{minipage}
    \hfill
    \begin{minipage}[b]{0.45\textwidth}
        \centering
        \begin{overpic}[width=0.48\textwidth]
        {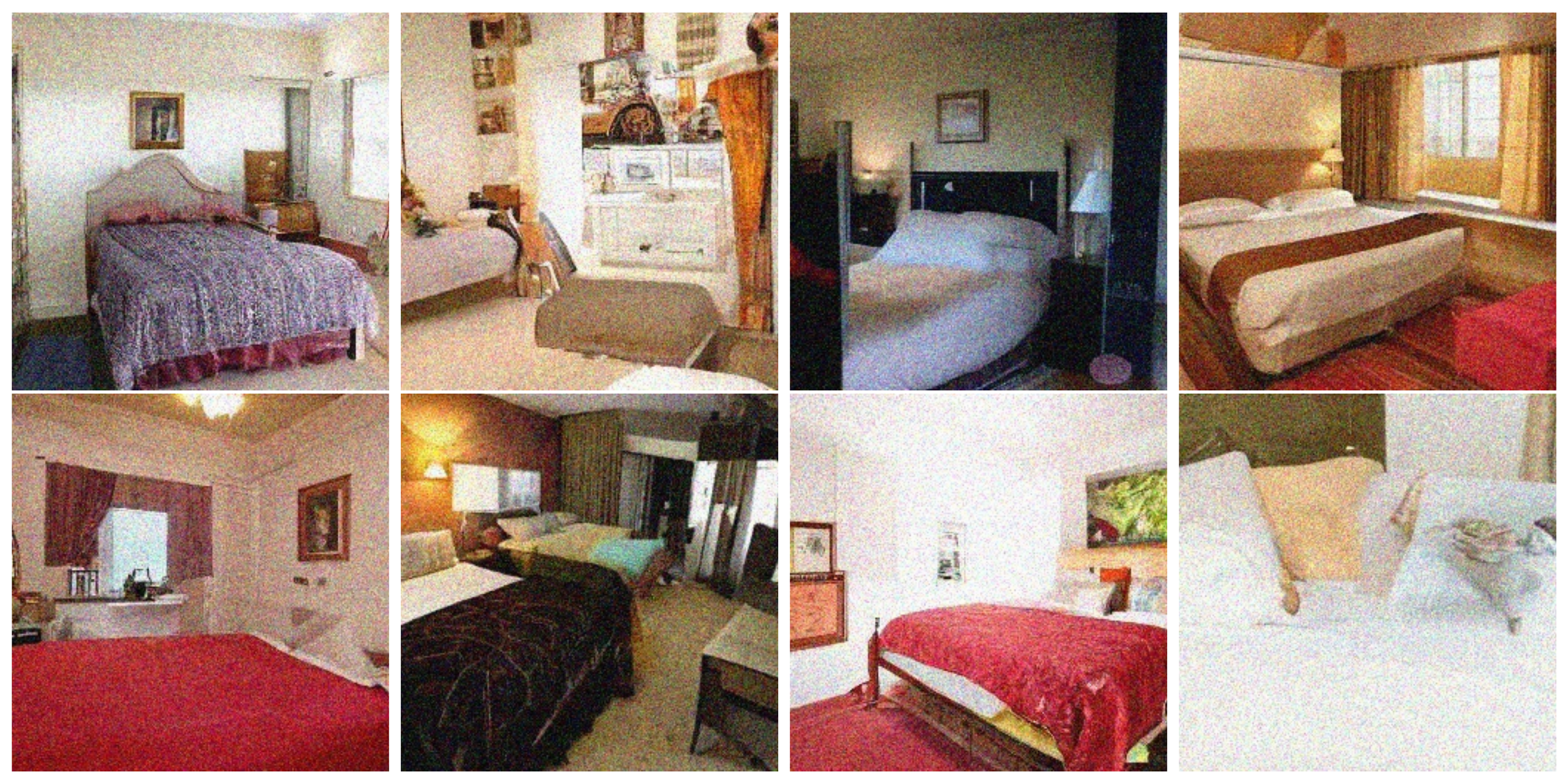}
            \put(24,53){DDPM~\cite{ho2020denoising}}
            \put(-11,7.5){\rotatebox{90}{{$\sigma=0.1$}}}
        \end{overpic}
        \hfill
        \begin{overpic}[width=0.48\textwidth]
        {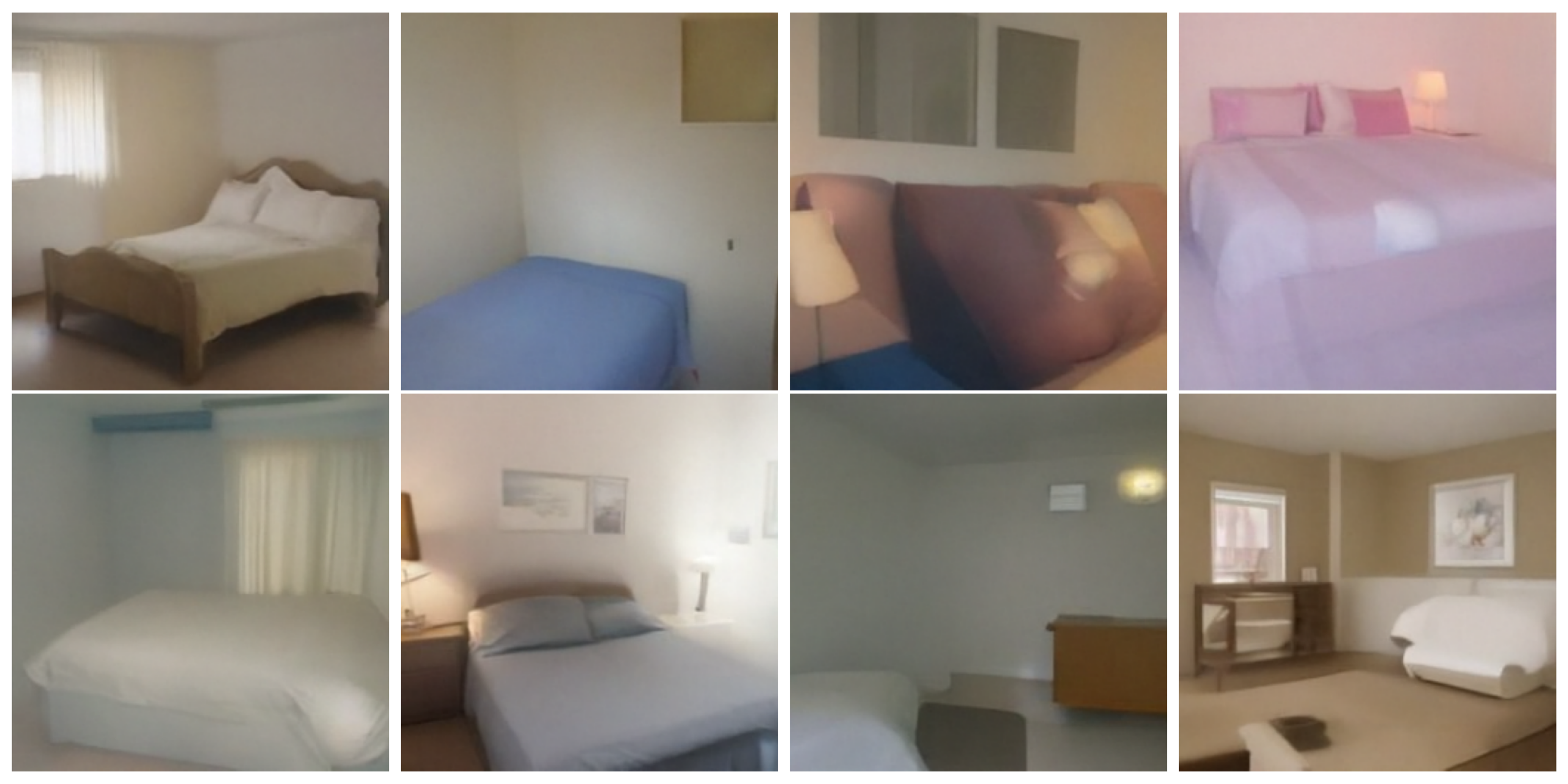}
            \put(22,53){\robust{adv}}
        \end{overpic}
        
        \vspace{0.15cm}
        
        \begin{overpic}[width=0.48\textwidth]
        {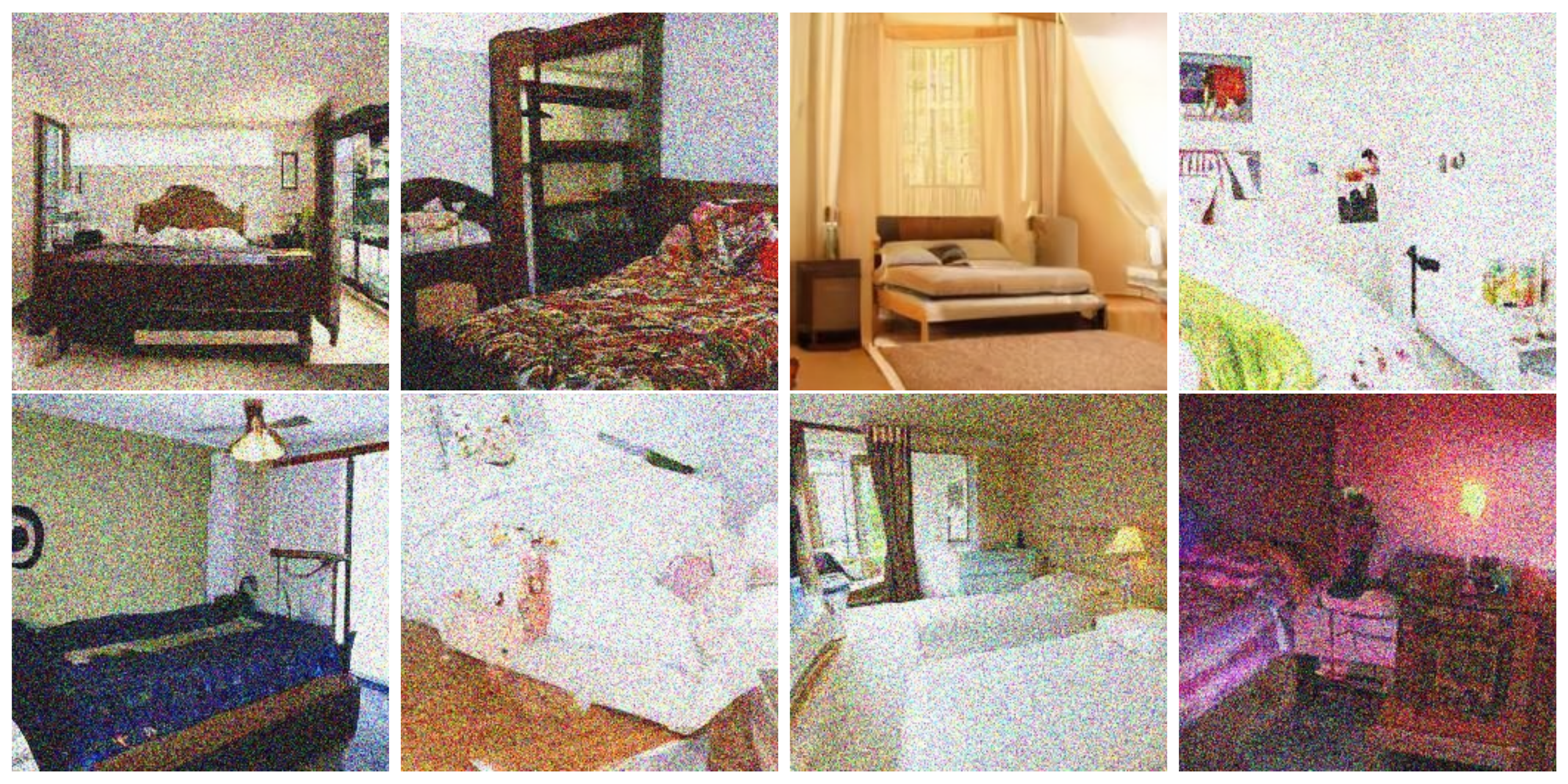}
            \put(-11,7.5){\rotatebox{90}{{$\sigma=0.2$}}}
        \end{overpic}
        \hfill
        \begin{overpic}[width=0.48\textwidth]
        {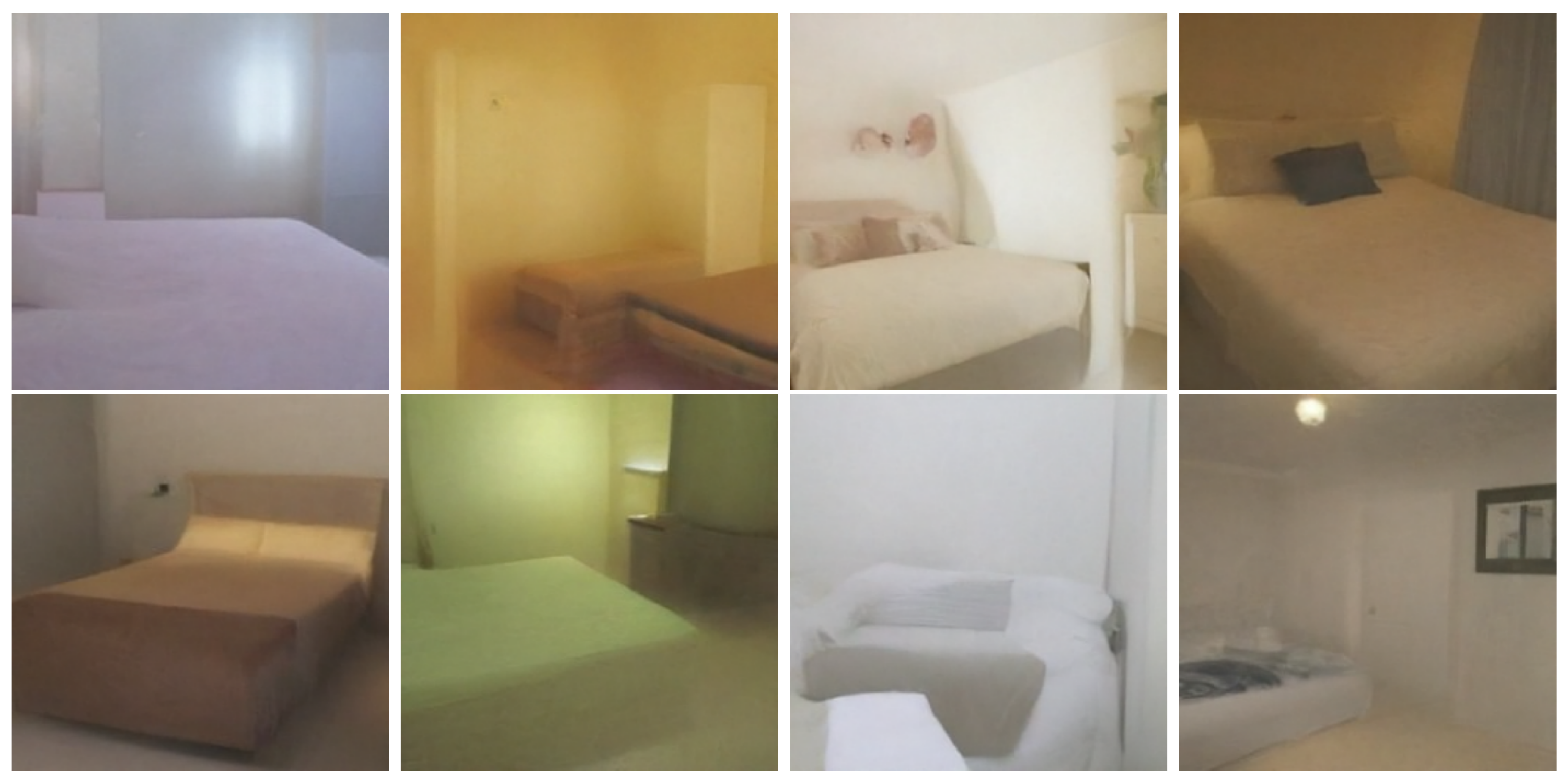}
        \end{overpic}
               
    \end{minipage}
\end{minipage}
    
    \vspace{1.3em}
    
    \begin{minipage}[b]{\textwidth}
        \centering

        \begin{minipage}[b]{0.24\textwidth}
            \centering
            \begin{overpic}[trim= 350 570 0 0, clip,width=\textwidth]{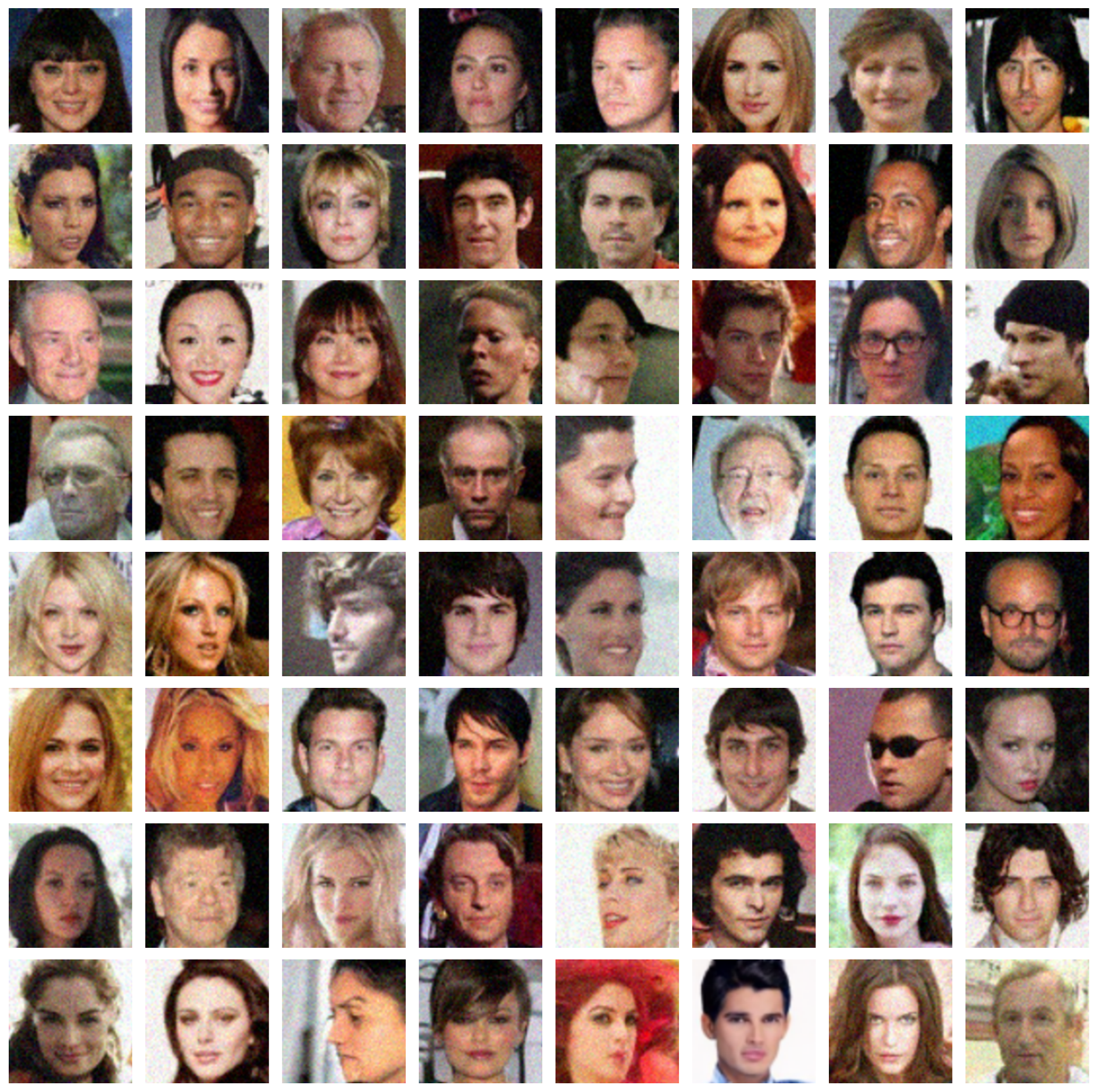}
                \put(16, 63){\small DDPM, $\sigma=0.1$}
            \end{overpic}
            
        \end{minipage}
        \hfill
        \begin{minipage}[b]{0.24\textwidth}
            \centering
            \begin{overpic}[trim= 350 570 0 0, clip,width=\textwidth]{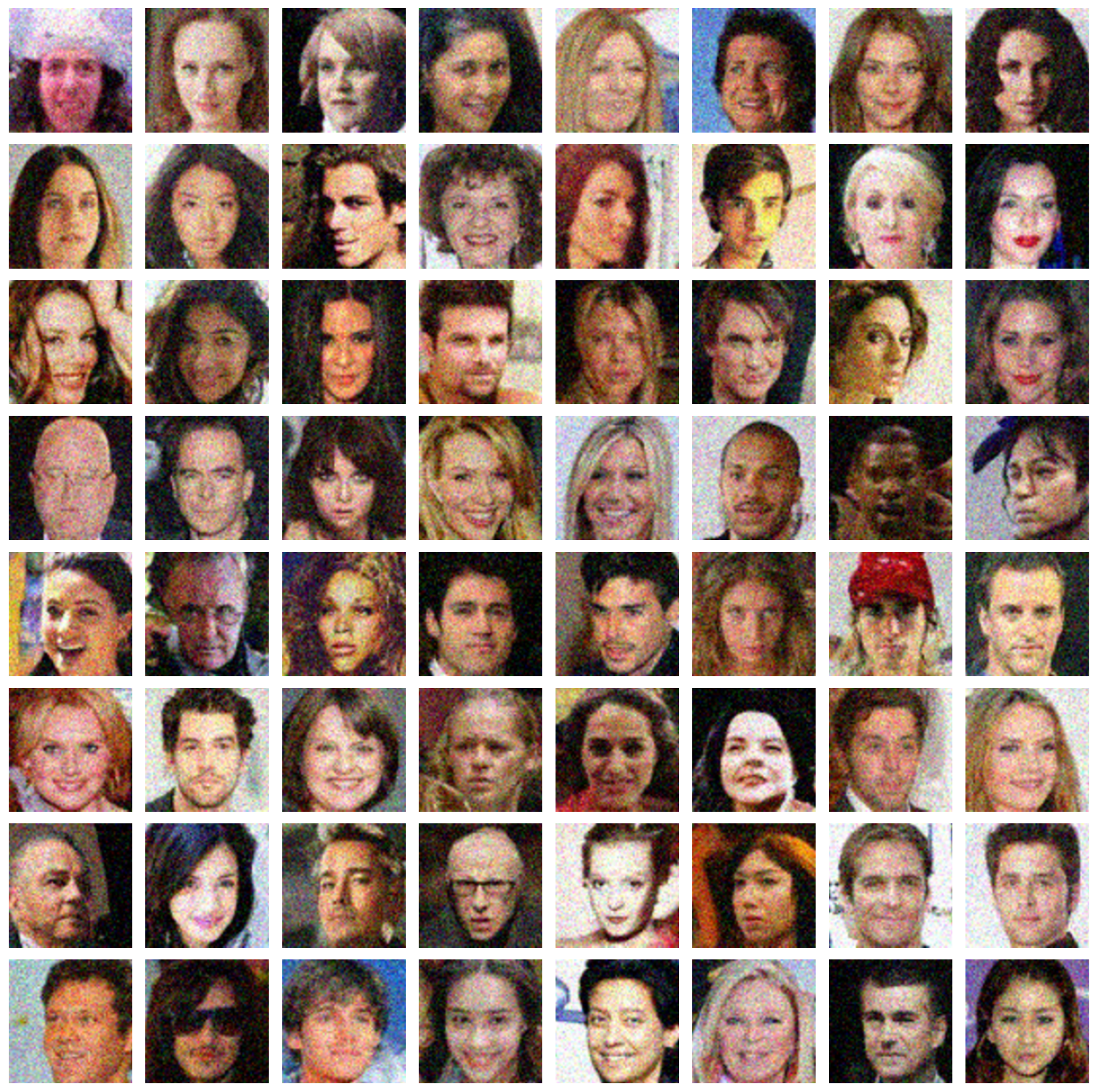}
                \put(16, 63){\small DDPM, $\sigma=0.2$}
            \end{overpic}
        \end{minipage}
        \hfill
        \begin{minipage}[b]{0.24\textwidth}
            \centering
            \begin{overpic}[trim= 350 570 0 0, clip,width=\textwidth]{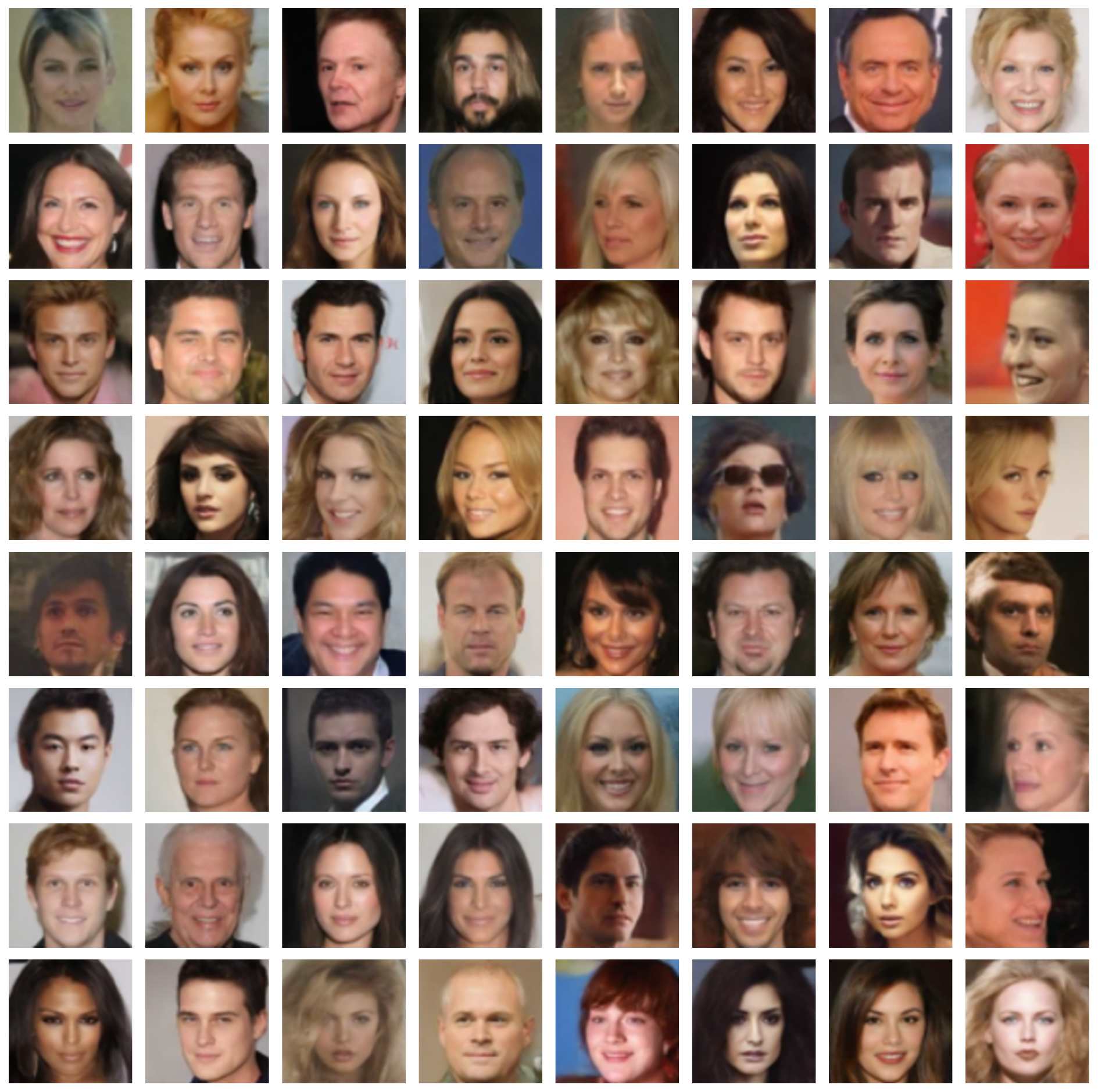}
               \put(10, 63){\small \robust{adv}, $\sigma=0.1$}
            \end{overpic}
        \end{minipage}
        \hfill
        \begin{minipage}[b]{0.24\textwidth}
            \centering
            \begin{overpic}[trim= 350 570 0 0, clip,width=\textwidth]{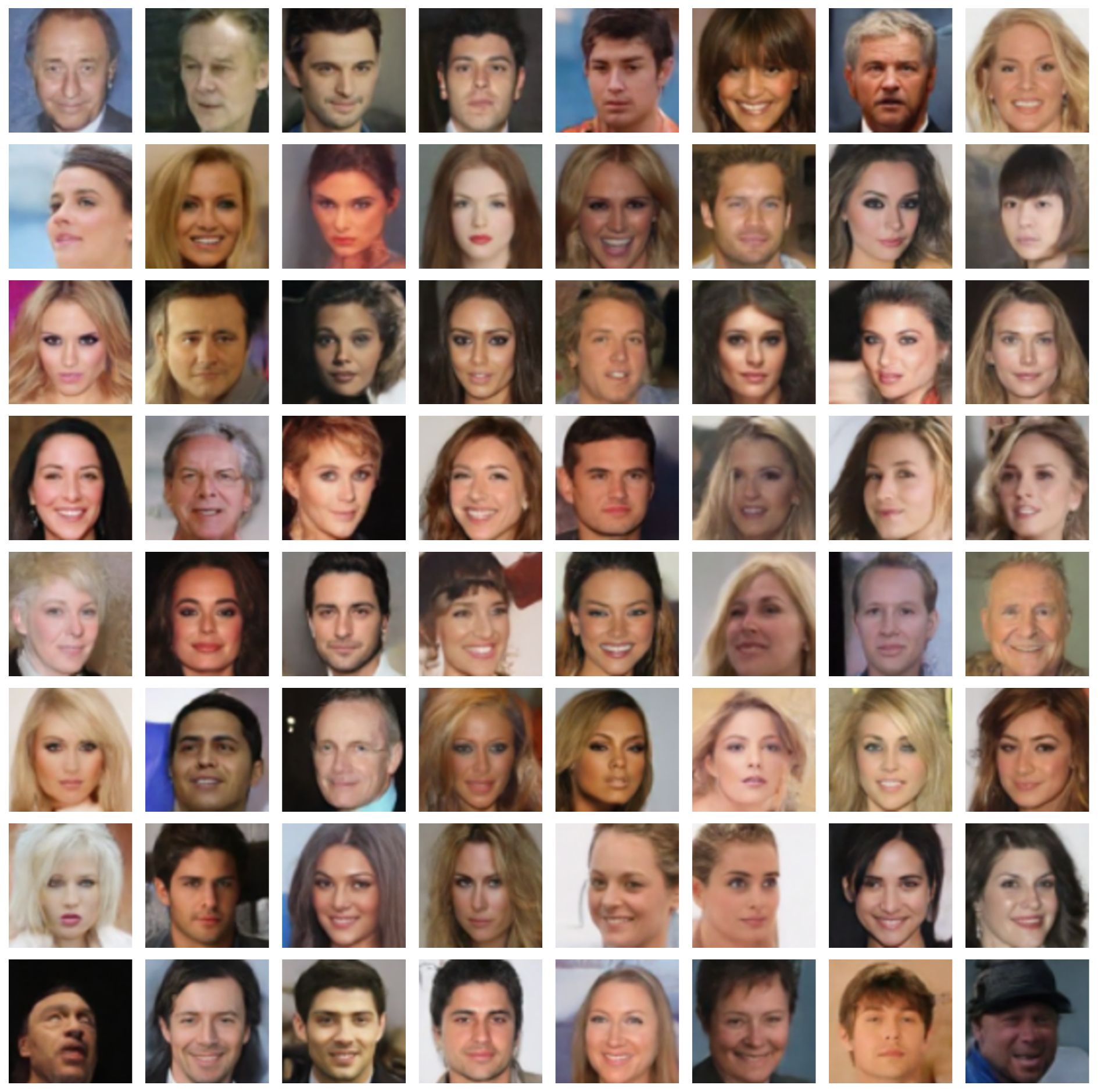}
            \put(10, 63){\small \robust{adv}, $\sigma=0.2$}
            \end{overpic}
        \end{minipage}
    \end{minipage}
    \caption{(\textit{top-left}) Despite 90\% of training data being corrupted with Gaussian noise, \robust{adv} generates smooth objects without artifacts, while DDPM retains noise. $\sigma=0.2$ equals adding $40\%$ of CIFAR-10 variability ($\sigma_{\text{data}}=0.5$). (\textit{top-right}) DDPM generates bedrooms that are irregular and unrealistic propagating the noise whereas \robust{adv} bedrooms are smooth and neat.
    (\textit{bottom}) Results on CelebA. DDPM replicates noise, while ours discards it and produces cleaner faces.}
    \label{fig:qualitatives}
\end{figure}
We use a progression of datasets from extremely controlled to real data: from synthetic 3D datasets where we have the possibility to simulate ``linear'' and unimodal distributions to more complex, multi-modal. We present results ``in vitro'' to guide our analysis on low-dimensional and also high-dimensional datasets taking perfect measure of errors. 
We further offer results on real datasets such as CIFAR-10~\cite{krizhevsky2009learning} (50K images, $32\times32$ pixels), CelebA~\cite{liu2015faceattributes} (202K images, $64\times64$ pixels) and LSUN Bedroom~\cite{yu15lsun} (303K images, $256\times256$ pixels), quantitatively assessing quality, using established metrics such as IS~\cite{salimans2016improved} and FID~\cite{heusel2017gans} computed on 50K images. Following prior art~\cite{daras2024consistent,daras2024much}, we experiment with Gaussian noise as corruption $p_{\text{noise}}(\bx)$ and only work in challenging settings, testing a percentage $p$ of corrupted data of $p=90\%$ with two levels of $\sigma=\{0.1,0.2\}$.
When computing FID, we always test on the \emph{clean dataset} despite training with noisy datasets. Unlike prior work, our method neither distinguishes between clean and noisy samples nor requires knowledge of the applied $\sigma$ for corruption. Our methods are indicated by \robust{adv} when using adversarial perturbation and  \robust{ran} if random. Finally, we show additional experiments that support our claims on less memorization, faster sampling, and robustness to attacks.
We set $\omega=2$ and $\gamma$ to $8/255$. Regularization strength  $\lambda$ is set to $0.3$: we have observed that when raising it  to $0.5$ we get an over-smoothing effect while low values prevent too much denoising.

\subsection{Evaluation using DDPM and DDIM}
\minisection{Controlled Experiments} \cref{fig:plane-robust} (\textit{right}) shows the results when training on high dim. data living on a subspace. When training on the clean, regular dataset, the baseline and our Robust DMs perform similarly though \robust{ran} has slightly better PSNR. When we train on the noisy dataset, $\{\bx^{\prime} + \bz\}_{i=1}^N$, then both Robust DMs offer superior performance (orange and blue curves) with wide gaps compared to the baseline (green curve) in both PSNR and reconstruction error. Specifically \robust{adv} appears to be better at noise unlearning. DDPM generations often consist in samples with saturated colors that are unlikely to be found in the training set while our method has better fidelity---see \cref{fig:plane-robust}(\textit{left}).
\begin{wrapfigure}{r}{0.5\textwidth}
    \centering
    \begin{overpic}[width=0.5\textwidth]{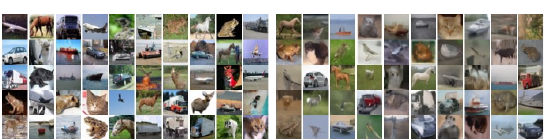}
            \put(18,25){{\small DDPM~\cite{ho2020denoising}}}
            \put(68,25){\robust{adv}}
        \end{overpic}
    \caption{Despite the FID increasing once trained on clean data, images by \robust{adv} appear smoother and background clutter is removed. }
    \label{fig:qualitative-cifar}
    \vspace{-1.1em}
\end{wrapfigure}
\minisectionq{Random or adversarial} We can also reply this question by ablating the use of $\pert_{\text{adv}}$ and $\pert_{\text{ran}}$. \cref{tab:model_comparison_rand_adv}(\textit{top}) shows that the adversarial perturbation can guarantee much stronger denoising effect than random yet being more expensive for training. The
impact of our \cref{eq:ours} is remarkable even in the case of random perturbation with an FID far below the baselines.

\minisection{Resistant to noise by design} \cref{tab:model_comparison_DDPM} compares our approach with the baseline DDPM and DDIM on CIFAR-10, CelebA and LSUN Bedroom yet corrupted with Gaussian noise. We start by showing that, despite our method has an inductive bias towards denoising the data distribution, if we apply it to the original dataset with no noise ($p=0\%$), we only get a slight increase in the FID. However, if we visually inspect the results we discover that ours is actually smoothing background features and the shape but still outlines of the objects are visible, as shown in \cref{fig:qualitative-cifar}.
When we switch to noisy settings, we have a large improvement over the baseline for both DDPM and DDIM.  
We highlight that while the baseline FIDs skyrocket to very high values for $p=90\%,\sigma=0.2$, the robust DMs are able to keep it in a reasonable range, generating images unaffected by the noise in the data.
 Quantitative evaluations are provided in \cref{tab:model_comparison_DDPM}, showing major improvement of the regularized training over standard training.
\cref{fig:qualitatives} illustrates our method's benefits on the proposed datasets under noisy data conditions.  More results and images are available in the supplementary material.

\begin{figure}[t]
\centering
\begin{minipage}{0.37\textwidth}
\centering
\captionof{table}{{Top:} Random vs adv. noise. {Bottom:} \robust{adv} allows fewer steps for better FID. Results on CIFAR-10.}
\resizebox{\textwidth}{!}{
\begin{tabular}{lcccc}
\toprule
\rowcolor{headerbg} $\sigma~\rightarrow$  & \multicolumn{2}{c}{0.1} & \multicolumn{2}{c}{0.2} \\
 \rowcolor{headerbg}metrics $\rightarrow$ & FID & IS & FID & IS \\
\midrule
\robust{ran} & 79.21 & 5.21 & 68.04 & 4.34 \\
\robust{adv} & \textbf{24.70} & \textbf{7.21} & \textbf{24.81} & \textbf{7.07} \\
\bottomrule
\label{tab:model_comparison_rand_adv}
\end{tabular}
}
\vspace{0.6em}
\resizebox{\textwidth}{!}{
\begin{tabular}{lcccc}
\toprule
\rowcolor{headerbg}steps $\rightarrow$ & \multicolumn{2}{c}{300} & \multicolumn{2}{c}{500} \\
\rowcolor{headerbg} metrics $\rightarrow$ & FID & IS & FID & IS \\
\midrule
DDPM~\cite{ho2020denoising} & 224.38 & 3.33 & 28.07 & \textbf{8.46} \\
\robust{adv} & \textbf{37.89} & \textbf{6.39} & \textbf{24.34} & 7.53 \\
\bottomrule
\end{tabular}
}
\end{minipage}
\hfill
\hspace{0.2em}
\begin{minipage}{0.56\textwidth}
\centering
\captionof{table}{Performance under different noise levels on different real datasets.  Values indicate FID $\downarrow$~/ IS $\uparrow$.}
\resizebox{0.93\textwidth}{!}{

\begin{tabular}{llcc|cc}
\toprule
\rowcolor{headerbg}
\textbf{p \%} & \textbf{\( \sigma \)} & \textbf{DDPM} & \textbf{\robust{adv}} & \textbf{DDIM} & \textbf{\robust{adv}} \\
\midrule
\multicolumn{6}{c}{\cellcolor{headerbg}\textbf{CIFAR-10}} \\
\midrule
\multirow[t]{2}{*}{\centering 0} & 0 & \textbf{7.2 / 8.95} & 28.68 / 7.04 & \textbf{11.62 / 8.36} & 31.20 / 6.38 \\
\multirow[t]{2}{*}{\centering 0.9} & 0.1 & 58.05 / 6.93 & \textbf{24.70 / 7.21} & 59.28 / 6.89 & \textbf{25.48 / 6.85} \\
\multirow[t]{2}{*}{\centering 0.9}& 0.2 & 102.68 / 4.19 & \textbf{24.81 / 7.07} & 105.43 / 4.09 & \textbf{24.93 / 6.69} \\
\midrule
\multicolumn{6}{c}{\cellcolor{headerbg}\textbf{CelebA}} \\
\midrule
\multirow[t]{2}{*}{\centering 0} & 0 & \textbf{3.49} / \tbf{2.61} & 19.83 / 2.13 & \textbf{6.19} / \tbf{2.61} & 17.59 / 2.18\\
\multirow[t]{2}{*}{\centering 0.9} & 0.1 & 54.90 / \tbf{2.40} &\textbf{14.54} / 2.09 &41.29 / \tbf{2.48}& \textbf{17.98} / 2.22\\
\multirow[t]{2}{*}{\centering 0.9}& 0.2& 96.03 /\tbf{2.65} & \textbf{16.53} / 2.11 &89.28 / \tbf{2.62}& \textbf{20.24} / 2.20\\
\midrule
\multicolumn{6}{c}{\cellcolor{headerbg}\textbf{LSUN Bedroom}} \\
\midrule
\multirow[t]{2}{*}{\centering 0} & 0 & \tbf{9.90} / 2.31
&57.13 / \tbf{2.34}&\tbf{27.00} / \tbf{3.15}& 48.80 / 2.39\\
\multirow[t]{2}{*}{\centering 0.9} & 0.1 & 53.81 / \tbf{3.33}&\tbf{44.07} / 2.35&50.53 / 3.19 & \tbf{48.90} / \tbf{3.96}\\
\multirow[t]{2}{*}{\centering 0.9}& 0.2 & 95.85  / \tbf{4.08}&\tbf{44.27} / 2.50&82.20 / \tbf{4.39}& \tbf{61.98} / 3.66\\
\bottomrule
\end{tabular}
}
\label{tab:model_comparison_DDPM}
\end{minipage}
\end{figure}

\minisection{Time complexity} Training with AT strongly impacts training time due to overhead of computations needed. DDPM training operations comprehend a single forward pass to get model prediction and a backward pass for weights update.
Our regularization adds a backward pass to obtain adversarial loss gradients over the perturbation and double the same DDPM operations. The estimate slow down is $\times 2.5$ for \robust{adv} whereas \robust{ran} is less time-consuming since it does not have to backpropagate for the adversarial perturbation.
Despite the training time being higher than the baseline, remarkably the inference time is the same as other methods and we can attain faster sampling---see \cref{sec:sampling}.

\subsection{Robust diffusion models memorize less}

\begin{figure}[h]
    \centering
    \begin{minipage}{0.45\linewidth}
        \centering
        \includegraphics[width=\linewidth]{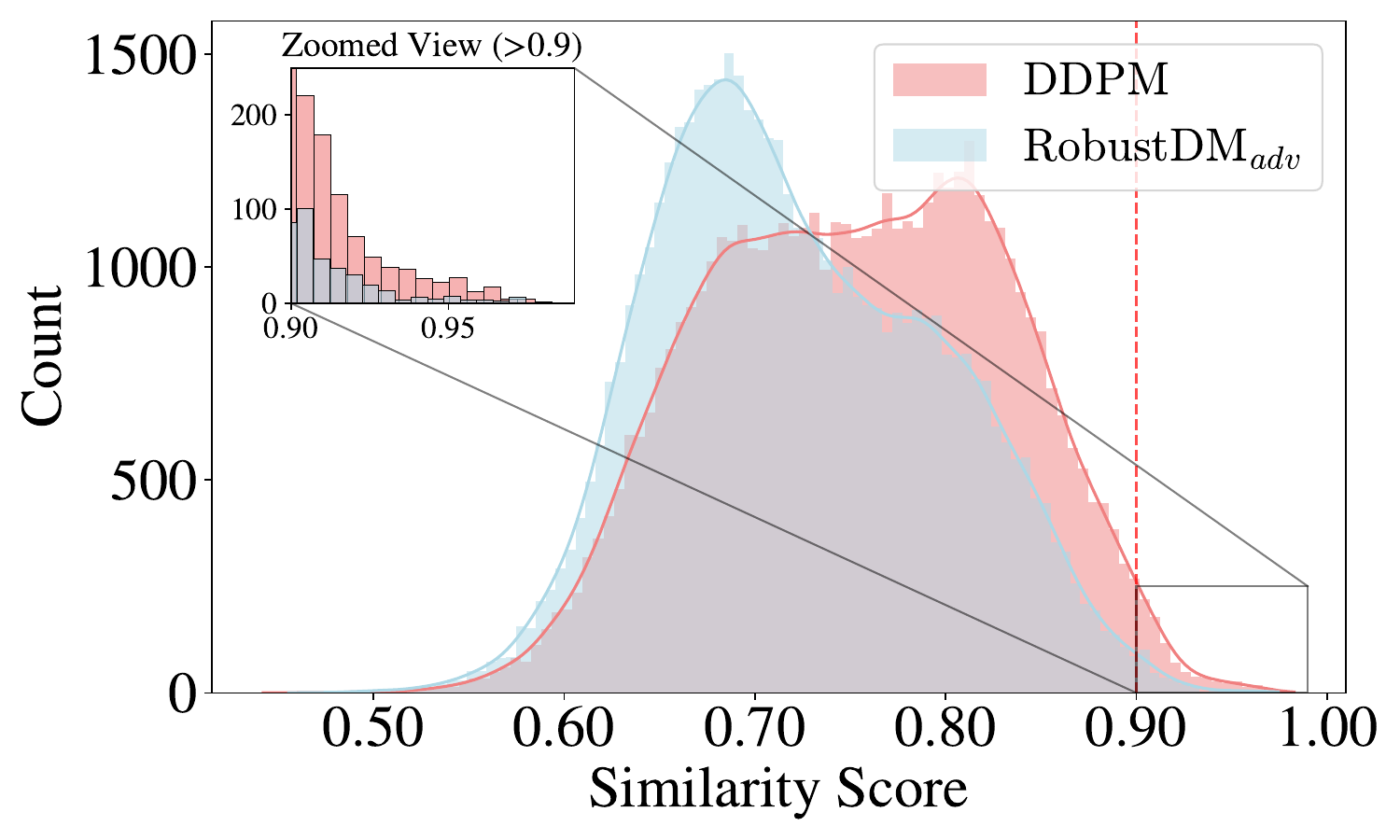}
    \end{minipage}
    \hfill
    \begin{minipage}{0.53\linewidth}
        \centering
        \begin{overpic}[width=\linewidth]{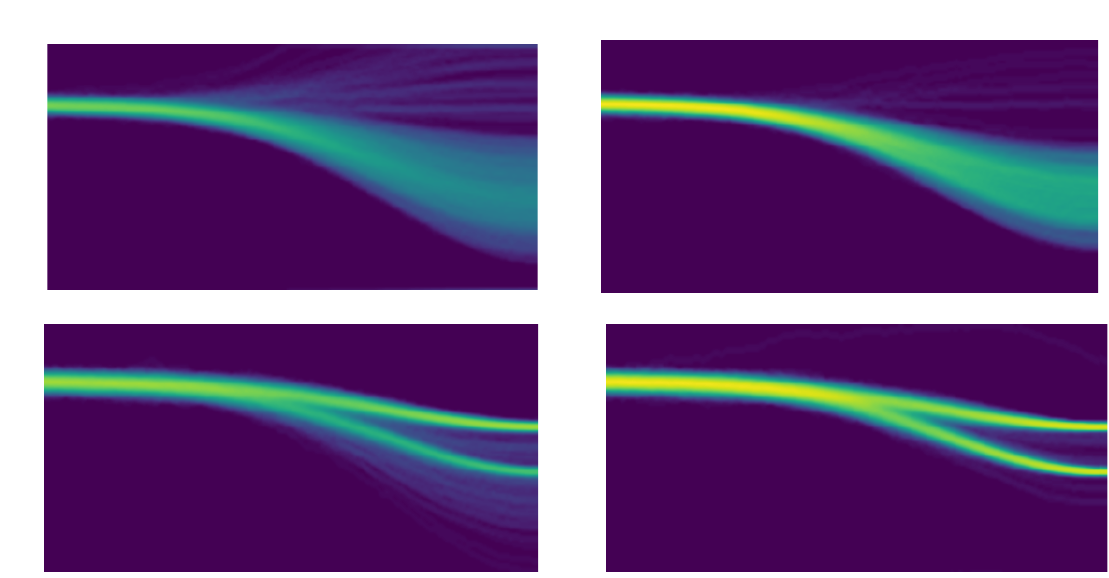}
            \put(68,50){\robust{adv}}
            \put(15,50){{\small DDPM ~\cite{ho2020denoising}}}
            \put(0,26){\rotatebox{90}{{\tiny\texttt{oblique-plane}}}}
            \put(0,3.5){\rotatebox{90}{{\tiny\texttt{3-gaussians}}}}
        \end{overpic}
    \end{minipage}
    \caption{(\textit{left}) The histogram shows similarities between generated samples and CIFAR-10, with values above 0.9 indicating near-duplicates. DDPM memorizes more, while \robust{adv} reduces near-replicas. (\textit{right}) Regular training tends to have diverging trajectories w.r.t. the data distribution, while \robust{adv} trades off variability for resilience with trajectories more clustered, sharp, and less faded.}
    \label{fig:mem-flow}
\end{figure}

Following~\cite{daras2024ambient} we show that Robust DMs are naturally less prone to memorize the training data.
We perform an experiment following~\cite{somepalli2023diffusion}: using DDPM and our \robust{adv} trained on clean CIFAR-10, we synthesize 50K images from each of them and measure the similarities of those images with the one in the training set, embedding the images with DINO-v2~\cite{oquabdinov2}.
In~\cite{daras2024ambient} a similar experiment was done yet using DeepFloyd IF instead of U-Net DDPM. Although U-Net has much less parameters than DeepFloyd IF---millions vs billions---one could assume that U-Net will overfit less. \cref{fig:mem-flow} (\textit{left}) shows that still a decent amount of generated samples have similarity higher than $0.90$. 
Similarity $\geq 0.9$ roughly corresponds to the same CIFAR image. Robust models have a histogram that is drastically shifted on the left and the curve of the histogram in the right part decays more rapidly, having less samples in the region  $\geq 0.9$.

\subsection{Smooth diffusion flow enables faster sampling}
\label{sec:sampling}
\minisection{Smooth diffusion flow}
\cref{fig:mem-flow} (\textit{right}) shows the diffusion flow from standard normal distribution to the data distribution.
To do so, we use DDPM \cite{ho2020denoising} framework and low-dimensional 3D data, projected to 2D for clarity. In \texttt{oblique-plane}, we can see how \robust{adv} captures less variability,  filtering out noise, while DDPM heatmap is more faded. Moreover, DDPM, misled by the noise, introduces a very subtle additional mode, whereas ours keeps the generation unimodal.
The same remarks hold for a multi-modal dataset: in \texttt{3-gaussians}  DDPM's trajectories are distorted by noise, while ours remain straight, preserving the multi-modal structure (only two modes are visible due to projection).

\minisection{Faster sampling} \cref{fig:mem-flow} (\textit{right}) shows that the diffusion flow of \robust{adv} is more compact and sharp, less faded. This could imply that the inference process may still recover the right path in case the regressed score vector is corrupted or is noisy or in case we deliberately use fewer steps in~\cref{eq:inference} for faster sampling. We tested this hypothesis and the trade-off table of FID in function of the number of steps taken is shown in \cref{tab:model_comparison_rand_adv} (\textit{bottom}). Even more, if we cross compare \cref{tab:model_comparison_rand_adv} (\textit{bottom}) with \cref{tab:model_comparison_DDPM}, on clean data \robust{adv} scores a better FID with 500 steps (24.34) vs 1000 steps (28.68).
This experiment supports our claim showing that \robust{adv} is still able to generate samples with good fidelity even if using fewer inference steps. The degradation using less steps is widely more graceful than DDPM especially when we take only $300$ steps over $1000$.

\begin{wrapfigure}{r}{0.5\textwidth}
    \begin{minipage}{0.5\textwidth}
    \vspace{-2.2em}
        \begin{algorithm}[H]
            \caption{DM Trajectory Attack}
            \label{alg:attack}
            \begin{algorithmic}
                \STATE {\bfseries Input:} attack ratio $p$, max timesteps $T$, model $\bepsilon_\theta$, scheduler $\alpha_t$, $\sigma(t)$, attack strength $\phi$;
                \STATE Sample $\rvx_T \sim \mathcal{N}(0,I)$
                \FOR{$t = T$ to $0$}
                    \STATE $\rvx_{t-1} \leftarrow \bepsilon_\theta(\rvx_t, t)$
                    \STATE $\hat{\rvx}_0 \leftarrow \frac{\rvx_t - \sqrt{1-\bar{\alpha}_t} \bepsilon_\theta(\rvx_t,t)}{\sqrt{\bar{\alpha}_t}}$
                    \STATE $\tilde{\boldsymbol{\mu}}_t(\rvx_t, \hat{\rvx}_0) \leftarrow\frac{\sqrt{\bar{\alpha}_{t-1}} \sigma(t)}{1 - \bar{\alpha}_t} \mathbf{\hat{\rvx}}_0 + \frac{\sqrt{\alpha_t} (1 - \bar{\alpha}_{t-1})}{1 - \bar{\alpha}_t} \mathbf{x}_t$
                    \STATE $\rvx_t' = \rvx_t + \pert$, $\pert \sim \mathcal{N}(0, \phi^2  \sigma(t)^2 I) $
                    \STATE $\rvx'_{t-1} \leftarrow \bepsilon_\theta(\rvx'_t, t)$
                    \STATE $\hat{\rvx}'_0 \leftarrow \frac{\rvx'_t - \sqrt{1 - \bar{\alpha}_t} \bepsilon_\theta(\rvx'_t,t)}{\sqrt{\bar{\alpha}_t}}$
                    \STATE $L \leftarrow \norm{\tilde{\boldsymbol{\mu}}_t(\rvx_t, \hat{\rvx}_0) - \tilde{\boldsymbol{\mu}}_t(\rvx'_t, \hat{\rvx}'_0)}_2^2$
                    \STATE $\rvx_t^\text{adv} = \rvx_t + \sigma(t) \cdot \text{sign}(\nabla_{\rvx_t} L)$
                    \STATE $\rvx_{t-1} \leftarrow \bepsilon_\theta(\rvx_t^\text{adv}, t)$
                \ENDFOR
            \end{algorithmic}
        \end{algorithm}
    \end{minipage}
    \vspace{-1.35em}
\end{wrapfigure}

\subsection{Robustness to adversarial attacks}
\begin{figure}[htb]
    \centering
    \begin{minipage}{0.5\textwidth}
        \centering
        \begin{overpic}[trim=0 0 0 300, clip, width=\linewidth]{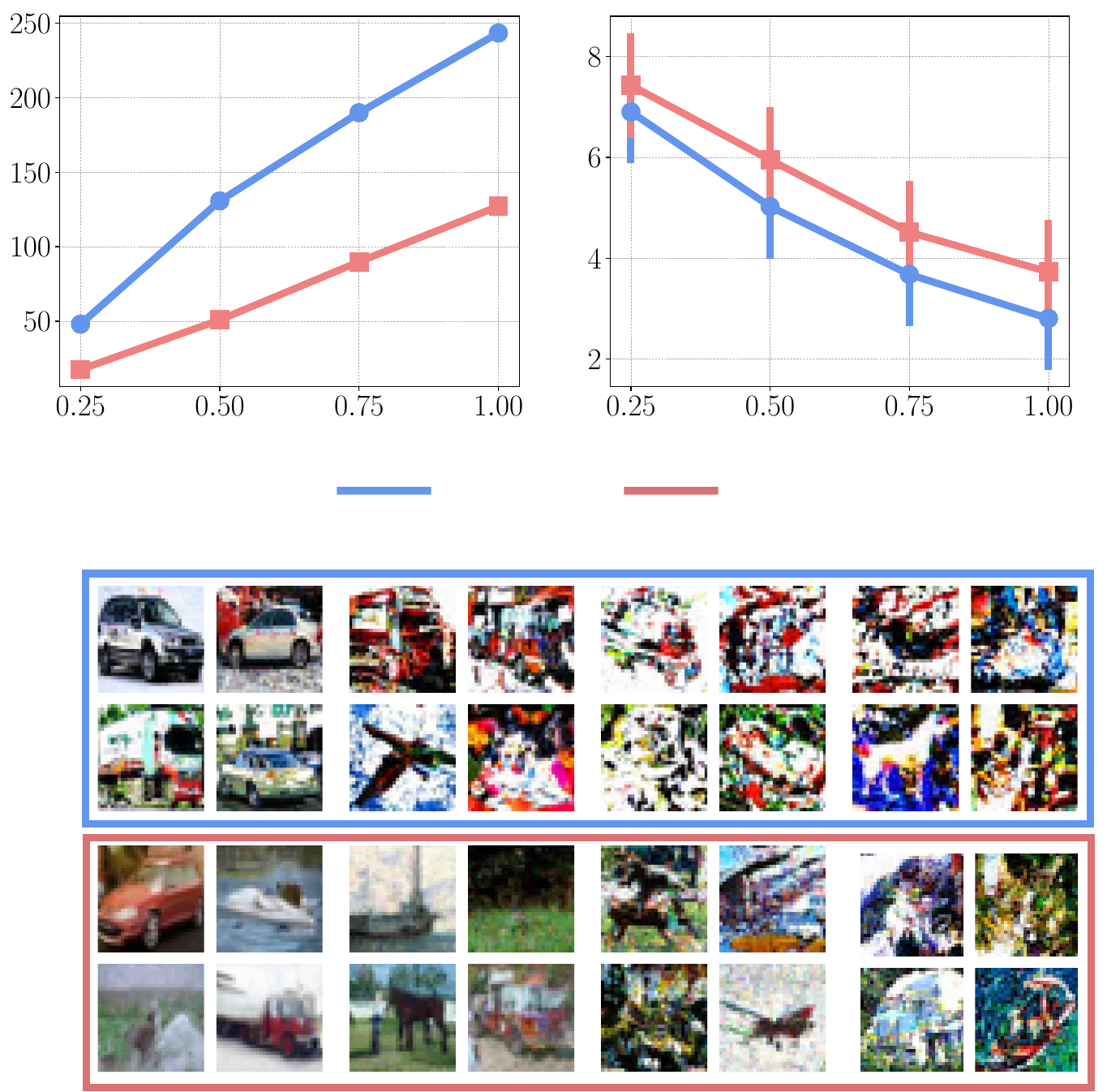}
            \put(15,47.75){{\footnotesize$25\%$}}
            \put(38,47.75){{\footnotesize$50\%$}}
            \put(63,47.75){{\footnotesize$75\%$}}
            \put(84,47.75){{\footnotesize$100\%$}}
            \put(3.,28){\rotatebox{90}{{\footnotesize DDPM}}}
            \put(3.,2.5){\rotatebox{90}{{\footnotesize {\robust{adv}}}}}
        \end{overpic}
        \caption*{\textbf{(a)}}
    \end{minipage}%
    \hfill
    \begin{minipage}{0.45\textwidth}
    \centering
    \resizebox{0.9\textwidth}{!}{
    \begin{tabular}{lcccc}
        \toprule
        \rowcolor{headerbg}\multicolumn{5}{l}{\textbf{FID $\downarrow$ w/ FGSM} --- \cref{alg:attack} and \cref{app:attack}} \\
        \rowcolor{headerbg}steps attacked $\rightarrow$ & 25 & 50 & 75 & 100 \\
        \midrule
        DDPM & 49.8 & 131.7 & 190.4 & 243.4 \\
        \robust{adv}  & \textbf{19.29} & \textbf{52.0} & \textbf{90.7} & \textbf{127.7} \\
        \bottomrule
    \end{tabular}
    }

    \vspace{0.5cm} 

    \resizebox{0.9\textwidth}{!}{
    \begin{tabular}{lcccc}
        \toprule
        \rowcolor{headerbg}\multicolumn{5}{l}{\textbf{FID $\downarrow$ w/ PGD} --- \cref{app:PGD}} \\
        \rowcolor{headerbg}steps attacked $\rightarrow$ & 25 & 50 & 75 & 100 \\
        \midrule
        DDPM & 55.7 & 134.5 & 200.3 & 248.1 \\
        \robust{adv}  & \textbf{22.7} & \textbf{55.8} & \textbf{98.1} & \textbf{128.6} \\
        \bottomrule
    \end{tabular}
    }
    \caption*{\textbf{(b)}}

    \label{tab:fid_attacks}
\end{minipage}
\caption{\textbf{(a)}  Robustness to Adversarial Attacks. While the baseline DDPM is susceptible to adversarial attacks, Robust DMs better resist them, yielding superior FID and IS for different percentages of time steps attacked (e.g., 25\% means 250 out of 1000 DDPM steps are attacked). \textbf{(b)} FID scores under FGSM and PGD attacks, varying the percentage of attacked time steps.}
\label{fig:attacks-composit}
\end{figure}

Our method is naturally resistant to attacks. Like classifiers, AT enforces robustness to adversarial perturbations in the diffusion flow. We propose an attack primarily as an analytical tool to better understand the fundamental sensitivity of the generative process to perturbations. The attack takes into account the stochastic nature of DM inference and the fundamental hypothesis of gaussianity for each diffusion stage. We propose attacking a DM in a white-box setting defining a sequence of adversarial perturbations that could maximally disrupt the trajectory at \emph{some} of the intermediate inference steps, defined as described in \cref{alg:attack}. We also propose a procedure to determine the range of values of the perturbation in order to maintain the assumption of the diffusion process; more information can be found in \cref{app:attack}. \cref{fig:attacks-composit} shows that our method is much more robust to attacks in diffusion flow: \robust{adv} can tolerate up to $50\%$ of time step attacked and still generate samples with decent fidelity. 
Only at $75\%$ time steps attacked, the generation fails for both.
The attack illustrated in \cref{alg:attack} is a single iteration attack, in \cref{app:PGD} we extend the pool of considered attacks to the iterative ones and provide the model performance in that setting.

\section{Related work}\label{sec:related}

\minisection{Diffusion models}
Score-based generative models ~\cite{song2019generative} introduce a maximum likelihood framework to learn the score function to generate samples that are likely to be part of the real data distribution. These models express the inference process through a Stochastic Differential Equations (SDE) ~\cite{dhariwal2021diffusion}.  Denoising Diffusion Probabilistic Models~(DDPMs)~\cite{ho2020denoising} first introduce diffusion process as  score-based generative framework , becoming \textit{de-facto} the standard algorithm in generative modeling on high-dimensional data, overcoming the previous adversarial min-max game between a generator and a discriminator (GANs)~\cite{goodfellow2020generative}. 
DMs not only achieve higher fidelity~\cite{dhariwal2021diffusion} but also provide a more stable training.
DMs have been extensively improved: working on the logarithmic likelihood estimate~\cite{nichol2021improved}, faster sampling~\cite{nichol2021improved,song2021denoising} and performing the diffusion process in the latent space~\cite{rombach2022high} instead of the pixel space.
In~\cite{karras2022elucidating,Karras2024edm2}, the authors provide insightful clarifications on several DMs design choices. Furthermore, they introduce an improved U-Net architecture featuring redesigned network layers that ensure consistent activation, weight, and update magnitude, achieving state-of-the-art FID on CIFAR and other benchmarks.
Lastly, \cite{song2023consistency} proposed \emph{consistency} models, a distillation method for one-step inference by directly mapping noise to data. The name \emph{consistency} arises from the fact that they enforce different noisy versions in the same trajectory to map to the same data. Unlike them, we enforce local \emph{smoothness}  of trajectories so its score field remains locally consistent.

\minisection{Denoising and inverse problems with DMs} The attention to apply DMs in scenarios where the data is not assumed to be curated and clean but is instead degraded or corrupted, has increased in recent years~\cite{aali2023solving,xiang2023ddm,daras2024ambient, daras2024much}. 
Given the specific challenges related to training with noisy data, this problem is closely related to inverse problems~\cite{tachella2024unsure,kawarg2024sure}.
Recently, a line of research focused on applying Stein’s Unbiased Risk Estimator (SURE)~\cite{metzler2018unsupervised} and its subsequent improvements, including UNSURE~\cite{tachella2024unsure}, GSURE~\cite{kawarg2024sure}, Soft Diffusion~\cite{darassoft}, and methods leveraging optimal transport for training with noise~\cite{daohigh}.

\minisection{Adversarial robustness} Adversarial robustness is loosely linked to denoising since AT can be seen as a way to remove spurious correlations~\cite{ye2024spurious} with improved out-of-domain generalization when transferring to a new domain~\cite{ilyas2019adversarial} or related to causal learning~\cite{zhang2020causal,zhangadversarial}. Surprisingly, only a few papers show that AT actually increases spurious correlations in some cases~\cite{moayeri2022explicit,moayeri2022comprehensive}, although robustness tools are still used to assess if a neural model relies on spurious associations between the input and output class~\cite{singlasalient,neuhaus2023spurious}. 
AT variants have been used to improve domain shift~\cite{salman2020adversarially} and out of distribution~\cite{wang2022improving}.
While it is reasonable to say that AT has been extensively studied on classifiers, its application to DMs remains unexplored, except for~\cite{sauer2024adversarial} where it is applied for fast sampling and \cite{yang2024structure} which investigates the batch samples interconnection.

\minisection{Adversarial defenses with denoising or randomized smoothing} There exist several adversarial defense techniques leveraging denoising~\cite{salman2020denoised, carlini2023certified} and randomized smoothing~\cite{cohen2019certified}, mainly in the context of classifiers. Regarding DMs \cite{song2024mimicdiffusion, liang2023advDM,liang2023mist} have shown adversarial perturbations, if applied at inference time, can significantly disrupt their generative capabilities, leading to deviations from clean data distributions.
Further works introduce the concept of robustness when fine-tuning DMs to make them robust in the context of adversarial purification~\cite{song2018pixeldefend,nie2022diffusion,lin2024adversarial}. 
While these methods differ in adversarial samples definition~\cite{li2025adbm,liu2025towards}, they share similar underlying objectives. 
Our work introduces AT in Diffusion Models to enforce local smoothness in the score field, which may help counteract such deviations. This suggests also a potential avenue for applying robust diffusion models in adversarial purification, being the optimization based on the same objective function.

\section{Conclusions and future work}\label{sec:conc}
We presented the first attempt to incorporate AT into DM training showing that AT for generative modeling implies smoothing the data distribution and can be effectively used for denoising the data. We also show that we need to reinterpret it as \emph{equivariant} property and not \emph{invariance}.
Our method has been proved to be highly robust even under 90\% of corrupted data with strong Gaussian noise.

\minisection{Broader impact} We first introduce the concept of robust trajectories in DMs, laying the foundation of broader studies of its possible applications.
There are many potential consequences of our work: we believe the robustness of DMs is not well studied and
can have highly beneficial societal impact. A direct application would be avoiding to learn outliers or other minor modes that are not proper of latent factors of our data.
Our research can lead generative AI   to be more robust as less susceptible to adversarial perturbations and memorizing less the data having models that better respect privacy, possibly improving Stable Diffusion~\cite{podellsdxl}.

\minisection{Limitations and future work} In the encoding steps of the denoising we believe we can improve the encoding functions by letting the network learn an $\omega$ that is input dependent. We also have to extend our method to work in fully corrupted settings ($p=100\%$) and port our approach to Elucidating DM framework (EDM)~\cite{karras2022elucidating,Karras2024edm2} in order to scale to larger datasets.

\newpage
\appendix
\onecolumn
\section{Appendix}
\subsection{Theoretical considerations on adversarial training for diffusion models}\label{sec:theory}
To craft an appropriate adversarial loss, at first forward and reverse processes are redefined in light of this further intermediate state. The main aim of performing adversarial training on a diffusion model is to enhance the robustness capability against adversarial attacks in its reverse process by providing the algorithm with some data that have previously been corrupted. We model this corruption process as an additional chain state and in this section we provide theoretical discussion for this assumption.

\subsubsection{The forward process}
The theoretical definition of the DDPM forward process is the following:
$$
q\left(\rvx_{1: T} \mid \rvx_0\right):=\prod_{t=1}^T q\left(\rvx_t \mid \rvx_{t-1}\right), \quad q(\rvx_{t}|\rvx_{t-1})=\gN\big(\rvx_t;\sqrt{1-\sigma(t)}\rvx_{t-1},\sigma(t)\rmI\big).
$$
where  $q(\mbf{x}_t \mid \mbf{x}_{t-1})$ represents the transition probability of the process to move from the state $\mbf{x}_{t-1}$ at the timestep $t-1$ to the state $\mbf{x}_t$ at the timestep $t$.
To achieve the aim of integrating the perturbation in the framework, the forward chain can be redefined considering a different dynamic of the adversarial forward process. A sample at the time step $t$ is first derived as defined above and then to it is added an adversarial perturbation $\pert_{\theta, t}$ that depends on the model actual state and on the value of $\rvx_t$. 
The overall attacked procedure to the model intermediate steps can be represented as a concatenation of two transitions.
The primary step is the ordinary DDPM transition from $\rvx_{t-1}$ to $\rvx_t$, which is modeled as $ q(\rvx_t \mid \rvx_{t-1})$. The attack transition can be modeled as the step that goes from $\rvx_t$ to $\rvx_t +\pert_t$ in the $t$-th timestep, being defined similarly as above $q'(\rvx_t + \pert \mid \rvx_t)$
The two transitions are designed to happen in the same time step $t$ of the chain and, being the two of them independent on each other, it is possible to model their interaction as a sub-sequence of steps of a Markov Chain.
This is possible since the transition $\rvx_{t-1} \rightarrow \rvx_t$ is already modeled like this and  $\rvx_{t} \rightarrow \rvx_t +\pert_t$ depends only on the weights of the model (which are constant when crafting the attack, so considerable as constant within the same evaluation) and the value of $\rvx_t$ conceived as ``previous state''. 
The resulting transition probability \( q''(\rvx_t + \pert_t \mid \rvx_{t-1}) \) is:
$$
q''(\rvx_t + \pert_t \mid \rvx_{t-1}) = q'( \rvx_t + \pert_t \mid \rvx_t) \cdot q(\rvx_t \mid \rvx_{t-1}).
$$

The overall chain can be written as:
\begin{equation*}
    q''\left(\mbf{x}_{1: T} +\pert_{1:T} \mid \mbf{x}_0\right)=\prod_{t=1}^T q'(\mbf{x}_t + \pert_t \mid \mbf{x}_t)\cdot q  \left(\mbf{x}_t \mid \mbf{x}_{t-1}\right).
\end{equation*}
with $ q(\rvx_{t}|\rvx_{t-1})=\gN\big(\rvx_t;\sqrt{1-\sigma(t)}\rvx_{t-1},\sigma(t)\rmI\big)$.
Being $q(\cdot)$ a gaussian transition and being the perturbation addition still modeled as a gaussian transition, the DM hypothesis of having only intermediate gaussian transitions still holds.

\subsubsection{Reverse process}
The reverse process in the diffusion models has the aim to define an algorithm that approximates the forward function and makes it possible to reconstruct the input.
In DDPM formulation the backward process is defined as:
$$
p_\theta\left(\mbf{x}_{0: T}\right):=p\left(\mbf{x}_T\right) \prod_{t=1}^T p_\theta\left(\mbf{x}_{t-1} \mid \mbf{x}_t\right), \quad p_\theta\left(\mbf{x}_{t-1} \mid \mbf{x}_t\right):=\mathcal{N}\left(\mbf{x}_{t-1} ; \boldsymbol{\mu}_\theta\left(\mbf{x}_t, t\right), \mbf{\Sigma}_\theta\left(\mbf{x}_t, t\right)\right).
$$
Following the previous substitutions, the desired equivalence when applying perturbations in the forward process would be : 
\[
p(\mbf{x}_t + \pert_t \mid \mbf{x}_{t-1}) = q''(\mbf{x}_t + \pert_t \mid \mbf{x}_{t-1}).
\]
that, if considering its approximation, reduces to:
\[
p(\mbf{x}_{t-1} \mid \mbf{x}_t + \pert_t) \propto p(\mbf{x}_t + \pert_t \mid \mbf{x}_{t-1})\cdot p(\mbf{x}_{t-1}).
\]
This consideration holds also in this case, so if we substitute the objective distributions $p(\cdot)$ with the desired ones we get:
$$ p(\mbf{x}_{t-1} \mid \mbf{x}_t + \pert_t) \propto p(\mbf{x}_{t-1}) \cdot q''(\mbf{x}_t + \pert_t \mid \mbf{x}_{t-1}) = 
p(\mbf{x}_{t-1}) \cdot q'(\mbf{x}_t + \pert_t \mid \mbf{x}_t) \cdot q(\mbf{x}_t \mid \mbf{x}_{t-1}).
$$
The above equations hold in case the reverse process is defined in closed form, while in our case the reverse function is learned function by $p_\theta(\cdot)$, which is designed and learned to properly converge to $p_{\text{data}}(\bx)$ at a specific timestep $0$ of the chain. To properly learn this objective the network is trained to learn to regress the amount of noise added in the forward process by minimizing the following simplified objective:
\begin{equation} 
    \Loss(\bx_t;\net)=\norm{\bsf{\epsilon}-\epsilon_{\net}(\bx_t,t)}_2^2 \quad \text{where} \quad \bsf{\epsilon} \sim \mathcal{N}(0,\mbf{I}) \quad \text{given } t \in [0,
    \ldots, T].
    \label{eq:DM_loss}
\end{equation}
where $q'(\mbf{x}_t + \pert_t \mid \mbf{x}_t)$ represents the transition probability of going from the state $\mbf{x}_t$ to the state $\mbf{x}_t ' = \mbf{x}_t + \pert_t$ in the same timestep $t$, the transition from an uncorrupted state to a corrupted one through $\pert_t$.
In this case there is no modeling available as the distribution depends on the kind of attack being performed during the training process but also depends on the state of the model, as the attack is crafted in white box mode: 
$$\pert_{\theta, t} = \underset{\|\pert\|\leq \varepsilon} \arg\max\norm{\bepsilon_{\theta}\big(\rvx_t+\pert,t\big)-\bepsilon_{\theta}\big(\rvx_t,t\big)}_2^2.$$

Given the proposed setting the aim is to define a cost function that allows to model the correct $\mbf{x}_{t-1}$ when considering the inverted process.
The probability distribution that the reverse process needs to learn is:
$$p_\theta(\mbf{x}_{t-1}\mid \mbf{x}_t + \pert_{\theta, t}) \propto p_\theta(\mbf{x}_{t-1}\mid \mbf{x}_t) p'_\theta(\mbf{x}_t \mid \mbf{x}_t + \pert_{\theta , t}).$$

\subsubsection{Variational lower bound in case of perturbation}
The Diffusion Models loss function is derived from an optimization regarding the variational lower bound. The ELBO is defined canonically as:
\begin{equation}
    \mathbb{E}[-\log \;p_\theta(\mbf{x}_0)] \leq \mathbb{E}_q \bigg[-\log \frac{p_\theta(\mbf{x}_{0:T})}{q(\mbf{x}_{1:T}\mid \mbf{x}_0)}\bigg]
    \\
    = \mathbb{E}_q \bigg[-\log(p_{\mbf{x}_t}) - \sum_{t \geq 1} \log\frac{p_\theta(\mbf{x}_{t-1} \mid \mbf{x}_t)}{q(\mbf{x}_t \mid \mbf{x}_{t-1})}\bigg] := L
\end{equation}
and the Diffusion Model loss derivation is the following:
$$
\begin{aligned}
L & =\mathbb{E}_q\left[-\log \frac{p_\theta\left(\mbf{x}_{0: T}\right)}{q\left(\mbf{x}_{1: T} \mid \mbf{x}_0\right)}\right] \\
& =\mathbb{E}_q\left[-\log p\left(\mbf{x}_T\right)-\sum_{t \geq 1} \log \frac{p_\theta\left(\mbf{x}_{t-1} \mid \mbf{x}_t\right)}{q\left(\mbf{x}_t \mid \mbf{x}_{t-1}\right)}\right] \\
& =\mathbb{E}_q\left[-\log p\left(\mbf{x}_T\right)-\sum_{t>1} \log \frac{p_\theta\left(\mbf{x}_{t-1} \mid \mbf{x}_t\right)}{q\left(\mbf{x}_t \mid \mbf{x}_{t-1}\right)}-\log \frac{p_\theta\left(\mbf{x}_0 \mid \mbf{x}_1\right)}{q\left(\mbf{x}_1 \mid \mbf{x}_0\right)}\right] \\
& =\mathbb{E}_q\left[-\log p\left(\mbf{x}_T\right)-\sum_{t>1} \log \frac{p_\theta\left(\mbf{x}_{t-1} \mid \mbf{x}_t\right)}{q\left(\mbf{x}_{t-1} \mid \mbf{x}_t, \mbf{x}_0\right)} \cdot \frac{q\left(\mbf{x}_{t-1} \mid \mbf{x}_0\right)}{q\left(\mbf{x}_t \mid \mbf{x}_0\right)}-\log \frac{p_\theta\left(\mbf{x}_0 \mid \mbf{x}_1\right)}{q\left(\mbf{x}_1 \mid \mbf{x}_0\right)}\right] \\
& =\mathbb{E}_q\left[-\log \frac{p\left(\mbf{x}_T\right)}{q\left(\mbf{x}_T \mid \mbf{x}_0\right)}-\sum_{t>1} \log \frac{p_\theta\left(\mbf{x}_{t-1} \mid \mbf{x}_t\right)}{q\left(\mbf{x}_{t-1} \mid \mbf{x}_t, \mbf{x}_0\right)}-\log p_\theta\left(\mbf{x}_0 \mid \mbf{x}_1\right)\right]
\end{aligned}
$$

$$
=\mathbb{E}_q\left[D_{\mathrm{KL}}\left(q\left(\mbf{x}_T \mid \mbf{x}_0\right) \| p\left(\mbf{x}_T\right)\right)+\sum_{t>1} D_{\mathrm{KL}}\left(q\left(\mbf{x}_{t-1} \mid \mbf{x}_t, \mbf{x}_0\right) \| p_\theta\left(\mbf{x}_{t-1} \mid \mbf{x}_t\right)\right)-\log p_\theta\left(\mbf{x}_0 \mid \mbf{x}_1\right)\right].
$$
In light of the previous considerations of the forward and backward process, it is possible to reconsider ELBO derivation as follows:

\begin{align}
\hspace{-60pt}
L &= \mathbb{E}_q \left[ 
-\log \frac{p_\theta(\mathbf{x}_{0:T})}{q''(\mathbf{x}_{1:T} \mid \mathbf{x}_0)} 
\right] \nonumber \\
&= \mathbb{E}_q \left[ 
-\log \frac{p_\theta(\mathbf{x}_T) \prod_{t=1}^T p_\theta(\mathbf{x}_{t-1} \mid \mathbf{x}_t + \delta_t)}{\prod_{t=1}^T q(\mathbf{x}_t \mid \mathbf{x}_{t-1}) \, q'(\mathbf{x}_t + \delta_t \mid \mathbf{x}_t)} 
\right] \nonumber \\
&= \mathbb{E}_q \left[ 
-\log p(\mathbf{x}_T) - \sum_{t \geq 1} \log \frac{p_\theta(\mathbf{x}_{t-1} \mid \mathbf{x}_t + \delta_t)}{q(\mathbf{x}_t \mid \mathbf{x}_{t-1}) \, q'(\mathbf{x}_t + \delta_t \mid \mathbf{x}_t)} 
\right] \nonumber \\
&= \mathbb{E}_q \left[ 
-\log p(\mathbf{x}_T) - \sum_{t \geq 1} \log \frac{p_\theta(\mathbf{x}_{t-1} \mid \mathbf{x}_t)}{q(\mathbf{x}_t \mid \mathbf{x}_{t-1})} - \sum_{t \geq 1} \log \frac{p'_\theta(\mathbf{x}_t \mid \mathbf{x}_t + \delta_t)}{q'(\mathbf{x}_t + \delta_t \mid \mathbf{x}_t)} 
\right] \nonumber \\
&= \mathbb{E}_q \left[ 
-\log p(\mathbf{x}_T) - \sum_{t > 1} \log \frac{p_\theta(\mathbf{x}_{t-1} \mid \mathbf{x}_t)}{q(\mathbf{x}_{t-1} \mid \mathbf{x}_t, \mathbf{x}_0)} \frac{q(\mathbf{x}_{t-1} \mid \mathbf{x}_0)}{q(\mathbf{x}_t \mid \mathbf{x}_0)} 
\right. \nonumber \\
&\quad \left. - \sum_{t > 1} \log \frac{p'_\theta(\mathbf{x}_t \mid \mathbf{x}_t + \delta_t)}{q'(\mathbf{x}_t \mid \mathbf{x}_t + \delta_t, \mathbf{x}_0)} \frac{q'(\mathbf{x}_t \mid \mathbf{x}_0)}{q'(\mathbf{x}_t + \delta_t \mid \mathbf{x}_0)} \right] \nonumber \\
&= \mathbb{E}_q \left[ 
-\log \frac{p(\mathbf{x}_T)}{q''(\mathbf{x}_T \mid \mathbf{x}_0)} - \sum_{t > 1} \log \frac{p_\theta(\mathbf{x}_{t-1} \mid \mathbf{x}_t)}{q(\mathbf{x}_{t-1} \mid \mathbf{x}_t, \mathbf{x}_0)} 
\right. \nonumber \\
&\quad \left. - \sum_{t > 1} \log \frac{p'_\theta(\mathbf{x}_t \mid \mathbf{x}_t + \delta_t)}{q'(\mathbf{x}_t \mid \mathbf{x}_t + \delta_t, \mathbf{x}_0)} \frac{q'(\mathbf{x}_t \mid \mathbf{x}_0)}{q'(\mathbf{x}_t + \delta_t \mid \mathbf{x}_0)} \right] \nonumber \\
&\quad - \mathbb{E}_q \left[ \log p_\theta(\mathbf{x}_0 \mid \mathbf{x}_1) - \log p'_\theta(\mathbf{x}_0 \mid \mathbf{x}_1 + \delta_1) \right] \nonumber \\
&= \mathbb{E}_q \left[ 
-\log \frac{p(\mathbf{x}_T)}{q''(\mathbf{x}_T \mid \mathbf{x}_0)} 
- \sum_{t > 1} \log \frac{p_\theta(\mathbf{x}_{t-1} \mid \mathbf{x}_t)}{q(\mathbf{x}_{t-1} \mid \mathbf{x}_t, \mathbf{x}_0)} 
\right. \nonumber \\
&\quad \left. - \sum_{t > 1} \log \frac{p'_\theta(\mathbf{x}_t \mid \mathbf{x}_t + \delta_t)}{q'(\mathbf{x}_t \mid \mathbf{x}_t + \delta_t, \mathbf{x}_0)} 
- \log p_\theta(\mathbf{x}_0 \mid \mathbf{x}_1) 
- \log p'_\theta(\mathbf{x}_0 \mid \mathbf{x}_1 + \delta_1) 
\right] .
\end{align}

The components to be optimized can be seen as two KL-divergences, recalling the formal definition of DDPM optimization. 
To lower the loss functions the two resulting KL divergences have to be reduced by optimizing both the measure of divergence between the forward $\mbf{x}_t$ and the approximated one, by correctly estimating the $\bepsilon$ and the measure of the $\pert_t$ noise  is added to $\mbf{x}_t$ at the timestep $t$. This distance measure is represented by the second KL divergence.
To transition from the notation $q(\mbf{x}_{t} \mid \mbf{x}_{t-1})$ to $q(\mbf{x}_{t-1} \mid \mbf{x}_t, \mbf{x}_0)$ it is first necessary to apply Bayes theorem and the chain rule of probability---the exact same reasoning can be used for the second sum.
\begin{enumerate}
    \item Start with the conditional probability distribution $q'(\mbf{x}_{t} \mid \mbf{x}_{t-1})$.
    \item Apply Bayes' theorem to express $q'(\mbf{x}_{t} \mid \mbf{x}_{t-1})$ in terms of $q'(\mbf{x}_{t-1} \mid \mbf{x}_t)$:
    \begin{align*}
    q(\mbf{x}_{t} \mid \mbf{x}_{t-1}) &= \frac{q(\mbf{x}_{t-1} \mid \mbf{x}_t) \cdot q(\mbf{x}_t)}{q(\mbf{x}_{t-1})}.
    \end{align*}
    \item Now, consider conditioning on an additional variable $\mbf{x}_0$. According to the chain rule of probability, we have:
    \begin{align*}
    q(\mbf{x}_{t-1}, \mbf{x}_t) &= q(\mbf{x}_{t-1} \mid \mbf{x}_t) \cdot q(\mbf{x}_t).
    \end{align*}
    \item We want to express $q(\mbf{x}_{t-1} \mid \mbf{x}_t)$ in terms of $\mbf{x}_0$ as well. So, we can rewrite the joint distribution $q(\mbf{x}_{t-1}, \mbf{x}_t)$ as $q(\mbf{x}_{t-1} \mid \mbf{x}_t, \mbf{x}_0) \cdot q(\mbf{x}_t, \mbf{x}_0)$.
    \item Use the chain rule again to break down $q(\mbf{x}_t, \mbf{x}_0)$:
    \begin{align*}
    q(\mbf{x}_t, \mbf{x}_0) &= q(\mbf{x}_t \mid \mbf{x}_0) \cdot q(\mbf{x}_0).
    \end{align*}
    \item Substituting these expressions back into our Bayes' theorem-derived expression, we get:
    \begin{align*}
    q(\mbf{x}_{t} \mid \mbf{x}_{t-1}) &= \frac{q(\mbf{x}_{t-1} \mid \mbf{x}_t, \mbf{x}_0) \cdot q(\mbf{x}_t \mid \mbf{x}_0)}{q(\mbf{x}_{t-1} \mid \mbf{x}_0)}.
    \end{align*}
    \item Rearrange terms to isolate $q(\mbf{x}_{t-1} \mid \mbf{x}_t, \mbf{x}_0)$, yielding the desired expression:
    \begin{align*}
    q(\mbf{x}_{t-1} \mid \mbf{x}_t, \mbf{x}_0) &= \frac{q(\mbf{x}_{t} \mid \mbf{x}_{t-1}) \cdot q(\mbf{x}_{t-1} \mid \mbf{x}_0)}{q(\mbf{x}_t \mid \mbf{x}_0)}.
    \end{align*}
\end{enumerate}

\subsection{Attack formulation}\label{app:attack}In inference mode, it is possible to represent the inverse Markov Chain as the sequence of intermediate realizations of gaussian distributions with fixed parameters regarding mean scaling and variance scaling.
From the paper~\cite{ho2020denoising} in Eqs. 6 and 7 the $t$-th step of the inference can be written as the sampling from the posterior distribution $q(\rvx_{t-1}|\rvx_tt,\rvx_0) = \mathcal{N}(\rvx_{t-1}; \Tilde{\boldsymbol{\mu}}_t(\rvx_t,\rvx_0), \Tilde{\boldsymbol{\beta}_t}\boldsymbol{I} )$, where: 
$$
\tilde{\boldsymbol{\mu}}_t(\mbf{x}_t, \mbf{x}_0) := \frac{\sqrt{{\alpha}_{t-1}} \sigma(t)}{1 - {\alpha}_t} \mbf{x}_0 + \frac{\sqrt{\alpha_t} (1 - {\alpha}_{t-1})}{1 - {\alpha}_t} \mbf{x}_t, \quad 
\tilde{\boldsymbol{\beta}}_t := \frac{1 - {\alpha}_{t-1}}{1 - {\alpha}_t} \sigma(t).
$$
This implies that, at each time step, the expected variance and mean of the distribution are defined in a specific manner. During inference, the value of $\rvx_0$ corresponds to the output obtained after the network's prediction. In the context of the DDPM framework, $\rvx_0$ is replaced by the estimated value, which depends on the epsilon-predicting network:
\[
\hat{\rvx}_0 = \frac{\rvx_t - \sqrt{1 - {\alpha}_t} \bepsilon_\theta(\rvx_t)}{\sqrt{{\alpha}_t}},
\]

To properly craft the attack, and still consider it legitimate, it is essential to scale it to the correct standard deviation to align with the diffusion process. Failing to do so would result in the network's inference being affected not by the perturbation itself but by the incorrect range of the perturbation, causing errors due to the inability to maintain the process within its gaussian assumptions.

In this context, the attack procedure follows the FGSM approach with random start. However, the perturbation is then scaled to match the appropriate variance at timestep $t$ to maintain consistency with the diffusion process.
The FGSM attack generates an adversarial example by perturbing the noisy sample $\rvx_t$ in the direction of the gradient of a cost function $ \Loss$ with respect to $\rvx_t$. Specifically, the adversarial perturbation is given by:
\[
\rvx_t' = \rvx_t + \phi \cdot \text{sign}\big(\nabla_{\rvx_t}  \Loss(\rvx_t)\big),
\]
where $\phi$ controls the magnitude of the perturbation, $\text{sign}(\cdot)$ represents the element-wise sign function.

The adversarial attack in this approach is integrated into the diffusion process by leveraging the predictive functions including a variance-handling mechanisms defined in the model in order to guarantee to concretely adapt to the gaussian hypothesis of the reverse Markov Chain. 
The adversarial attack begins with perturbing the input $\rvx_t$ defining its $\rvx_t'$ as:

$$\rvx_t' = \rvx_t + \pert, \qquad \pert \triangleq \mathcal{N}(0, \phi^2\cdot \sigma(t)^2).$$

The cost function for the adversarial attack is theoretically  defined based on the mean prediction:
$$
\mathcal{L}_{FGSM}= \norm{\tilde{\boldsymbol{\mu}}_t(\mbf{\rvx}_t, \mbf{\rvx}_0)- \tilde{\boldsymbol{\mu}}_t(\mbf{\rvx'}_t, \mbf{\rvx}_0)}_2^2
$$
where $\tilde{\boldsymbol{\mu}}_t$ represents the predicted mean of the diffusion process at time step $t$, which depends on both the input, respectively the clean sample $\rvx_t$ and the adversarial one $\rvx_t'$, and the original sample $\rvx_0$. The optimization goal is to maximize the discrepancy between the predicted means of the clean and adversarial inputs, ensuring that the perturbation effectively disrupts the reverse diffusion process. This cost function, if considered in light of the model's prediction in the $\epsilon$-prediction setting, can be formulated as:
$$
\mathcal{J_\theta}(\rvx_t, \pert,t ) = \norm{\bepsilon_{\theta}\big(\rvx_t+\pert,t\big)-\bepsilon_{\theta}\big(\rvx_t,t\big)}_2^2
$$
To compute the adversarial perturbation $\pert$, the gradient of the loss $\mathcal{J_\theta}$ with respect to $\rvx_t'$ is used:
$$
\pert = \sigma(t)\cdot \text{sign}\left(\nabla_{\rvx_t} \mathcal{J_\theta}(\rvx_t, \pert,t )\right),
$$
where $\sigma(t)$ scales the perturbation to ensure it adheres to the variance of the gaussian noise in the reverse diffusion process at the $t$-th step. This step aligns the adversarial attack with the stochastic nature of the model, ensuring the perturbation remains consistent with the gaussian hypothesis.

The final adversarial example is then obtained as:
\[
\rvx_t^\text{adv} = \rvx_t + \pert.
\]

The adversarially perturbed sample $\rvx_t^\text{adv}$ is fed back into the reverse diffusion process, following the recurrence of the inference.
\subsection{Iterative attack}
\label{app:PGD}
In \cref{alg:attack} we described the attack version that applies a single step attack procedure applied to each and every inference timestep.
In this section we propose a multi-step attack approach based on the PGD iterative attack that, similarly to what described in the previous algorithm, aims to attack model generation at timestep level. We again highlight that this attack is not intended as a practical attack proposed in this paper, the main aim of showing this attack approaches is to provide a procedure to assess the abilities of the DM to be resilient against minor perturbations applied to every sampling iteration.
In \cref{alg:multi-attack} we propose the multi-step approach, implemented by applying at every iteration the PGD-20 attack.
\begin{algorithm}[htb]
    \caption{Adversarial Attack on a Diffusion Model.}
    \label{alg:multi-attack}
    \begin{algorithmic}
    \STATE {\bfseries Input:} percentage of attacked timesteps $p$, total timesteps $T$, model $\bepsilon_\theta$, scheduler values $\alpha_t$ and $\sigma(t)$, perturbation strength $\phi$, iterations $\mathbf{N}$, the projection operator $\mathbb{P}$
    \STATE $\rvx_T \sim \mathcal{N}(0,I)$
    \FOR{t = $T$ to 0}
        \STATE $\rvx_{t-1} \leftarrow \bepsilon_\theta(\rvx_t, t)$
        \STATE $\hat{\rvx}_0 \leftarrow\frac{\rvx_t - \sqrt{1 - \bar{\alpha}_t} \bepsilon_\theta(\rvx_t,t)}{\sqrt{\bar{\alpha}_t}}$
        \STATE $\tilde{\boldsymbol{\mu}}_t(\rvx_t, \hat{\rvx}_0) \leftarrow \frac{\sqrt{\bar{\alpha}_{t-1}} \sigma(t)}{1 - \bar{\alpha}_t} \mathbf{\hat{\rvx}}_0 + \frac{\sqrt{\alpha_t} (1 - \bar{\alpha}_{t-1})}{1 - \bar{\alpha}_t} \mathbf{x}_t$
        \STATE $\pert_0 \sim \mathcal{N}(0, \phi^2  \sigma^2(t) I)$ 
        \FOR {n = 0 to $N-1$}
        \STATE $\rvx_{t,n}' = \rvx_t + \pert_n$; 
        \STATE $\rvx'_{t-1, n} \leftarrow \bepsilon_\theta(\rvx'_{t,n}, t)$
        \STATE $\hat{\rvx}'_{0,n} \leftarrow\frac{\rvx'_{t,n} - \sqrt{1 - \bar{\alpha}_t} \bepsilon_\theta(\rvx'_{t,n},t)}{\sqrt{\bar{\alpha}_t}}$
        \STATE $L = \norm{\tilde{\boldsymbol{\mu}}_t(\rvx_t, \hat{\rvx}_0) - \tilde{\boldsymbol{\mu}}_t(\rvx'_{t,n}, \hat{\rvx}'_{0,n})}_2^2$
        \STATE $\pert_{n+1}={\sigma(t)/N} \cdot \text{sign}(\nabla_{\rvx_t} L)$
        \ENDFOR
        \STATE $\pert= \mathbb{P}(\pert, -\sigma(t), \sigma(t))$
        \STATE $\rvx_{t}^\text{adv} = \rvx_{t} + \pert$
        \STATE $\rvx_{t-1} \leftarrow \bepsilon_\theta(\rvx_t^\text{adv}, t)$
    \ENDFOR
    \end{algorithmic}
\end{algorithm}
In this case, being the attack iterative, it is necessary to project at the end the perturbation in order to keep its values within the range $[-\sigma(t),\sigma(t)]$. This values has been chosen following the gaussianity hypothesis of the intermediate MC states. Diffusion models model intermediate data through intermediate gaussian distributions where the possible values would have standard deviation $\sigma(t)$. In order not to diverge too much from data distribution and be in a suitable range of possible values we decided to impose as ray of the projection interval the same standard deviation, making it also adaptive to the considered timestep.
The table \cref{tab:fid_attacks} shows model performance under this PGD like version of a diffusion model attack.

\maketitle

\section{Supplementary Material}
This supplementary material is intended to complement the main paper by providing further motivation for our assumptions and design choices, as well as additional ablation studies on the proposed datasets to demonstrate the effectiveness of our method. It is organized into the following sections.

\cref{sec:motivation} discusses the main differences among the considered approaches, offering a deeper analysis that includes both geometrical and empirical motivations behind the adopted design choices. It also clarifies the distinction between invariance and equivariance, and presents statistics on the adversarial perturbation $\pert$;
\cref{sec:qual-diff} presents a more detailed analysis of the diffusion flow dynamics by examining the trajectories obtained from low-dimensional datasets under different conditions. 
\cref{sec:qual-cifar} provides an extensive qualitative ablation across the real-world datasets introduced in the paper, showcasing a wide variety of samples and comparisons;
\cref{sec:analysis} offers additional observations and insights into the proposed approach. \textbf{We encourage readers to zoom in and compare the results for a better understanding of their quality}.

\subsection{Observations and motivations on our adversarial training framework}
\label{sec:motivation}

\subsubsection{Equivariant and invariant functions for adversarial training}
Adversarial training in classification has been widely studied over years  \cite{goodfellow2014explaining,madry2017towards,zhang2019theoretically,wang2019improving,shafahi2019adversarial,wong2020fast,sriramanan2021towards,sri2020guided, wang2023better,zhu2021towards,mirza2024shedding} in different settings, threat models and under different perspectives.
These methods share the objective to enforce invariance in the neural network $f_\theta$, since the final objective is to enforce the output of the network not to vary in the presence of minor changes in the network input.
However, in generative modeling, particularly diffusion models (DMs), enforcing invariance hinders learning the correct distribution, making the model unable to take into account input changes in its prediction.
Ignoring the adversarial perturbations applied during a perturbed training leads to deviations in trajectories, resulting in an inaccurate learned distribution. Conversely, training the model to incorporate the negative of the perturbation helps it recognize and manage potential deviations, enabling it to handle noise with broader standard deviations more effectively.
In \cite{Nguyen2019_ANSR}, the authors extend the concept of adversarial attacks to regression tasks, even though considering regression tasks on tabular datasets.
Their proposed method addresses these attacks by introducing an adversarial training loss based on numerical stability, improving performance under adversarial conditions. 
Even though the latter bridges the concepts of regression and AT, an analysis of implications in the case of randomized and adversarial training applied to the generative model is still a topic to cover, particularly with reference to generative models.
In this spirit, we propose a new training framework inspired by AT with the aim of shedding light on the concept of adversarial training for
DMs, exploiting knowledge from both functional analysis and classification neural networks.

\begin{figure}[h]
    \centering
    \includegraphics[width=0.55\linewidth]{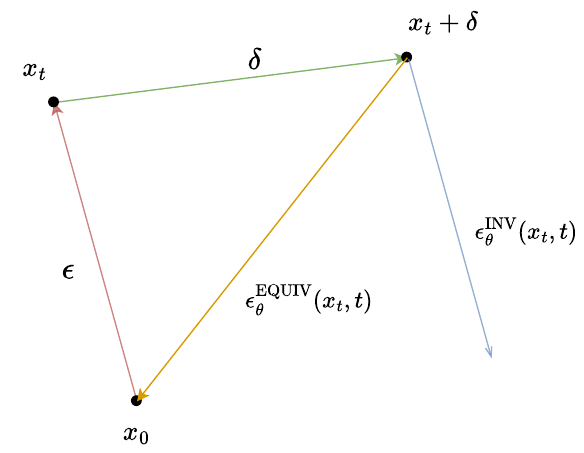}
    \caption{Not applying equivariance $\epsilon_{\theta}^{\text{EQUIV}}(\mbf{x}_t,t)$, the model drifts and ends up in a different point of the space than the desired one, learning then the perturbation that we added as in $\epsilon_{\theta}^{\text{INV}.}(\mbf{x}_t,t)$}
    \label{fig:equivariance-appendix}
\end{figure}

Formally defining the two properties, we can define both invariance and equivariance.
Given a function $f:X\rightarrow Y$, as well as a specified group actions $A$, $f$ is said to be \textit{equivariant} with respect to a transform $a\in A$ if and only if     \begin{equation}
        f(a\circ x)= a \circ f(x), \,\,\,\,\, x \in X
    \end{equation}
Given a function $f:X\rightarrow Y$, as well as a specified group action $A$, $f$ is said to be \textit{invariant} with respect to an $a\in A$ transform if and only if 
\begin{equation}
     f(a\circ x)= f(x), \,\,\,\,\, x \in X 
\end{equation}
In \cref{fig:equivariance-appendix} we extend Fig.1 of the paper and depict what happens at the trajectory level if we enforce invariance instead of equivariance.
The vector $\epsilon$ represents the noise that is added by the diffusion process, $\delta$ represents the added noise by the adversarial training. Finally, we will have two different versions of the noisy point, namely $\rvx_t$ and $\rvx_t +\delta$. The model, if unattacked, would like to regress a portion of noise equivalent to $-\epsilon$ so that it is able to correctly go back to $\rvx_0$. When applying $\delta$, the network's objective still has to be the same. The figure shows that if, given the noisy staring point $\rvx_t + \delta$, the model is enforced to learn again $-\epsilon$, so if the invariance is applied, the ending point would be some other point in the space different wrt. $\rvx_0$. On the contrary, if equivariance is applied, the network is forced to regress $-(\epsilon  + \delta)$, making the model able to correctly regress $\rvx_0$.

\subsubsection{Invariance regularization does not work}
As empirical evidence of the inconsistency of invariance training in AT for DMs, we prove it on low-dimensional data.
As a proof-of-concept, we consider the \texttt{oblique-plane} 3D dataset as data to train on, and then we impose adversarial training, following the same setting as in Algorithm 1,  enforcing instead invariance by minimizing the  loss function:
\begin{equation}
\mathcal{L}_{\text{AT}}(\rvx_t,\rvx^\text{adv}_t, t,\bepsilon)=\arg\min_{\theta}\underbrace{\norm{\bepsilon_{\theta}\big(\rvx_t,t\big)-\bepsilon}_2^2}_{\mathcal{L_\text{DM}} \text{ to fit data distr.}}+ \underbrace{\lambda_t\norm{\bepsilon_{\theta}\big(\rvx_t^\text{adv},t\big)-[\bepsilon_{\theta}\big(\rvx_t,t\big)]}_2^2}_{\mathcal{L_\text{reg}} \text{ to enforce invariance}}   
\end{equation}
We decided to implement this example on 3D data in order to have the possibility of observing the behavior of 3D trajectories. The plot shows it displaying side-to-side DDPM \cite{ho2020denoising} and invariance in the same data settings as the one displayed in Fig. 2. 
The model, by enforcing invariance, loses the ability to correctly reconstruct the data manifold, not being able to generate points in the data distribution, whereas the model trained through standard DDPM learns the data distribution but still suffers from learning the noise in case of noisy data.
The same behavior can be observed when looking at the trajectories. This analysis clarifies even more what is the generation dynamics. The model creates sparse trajectories that do not tend to be clustered, neither at the beginning of the generation nor at the end, thereby causing generated samples to be completely off the data subspace.

\begin{figure}[h!]
    \centering
    \begin{overpic}[trim=1cm 4cm 4cm 1.3cm, clip, width=\linewidth]{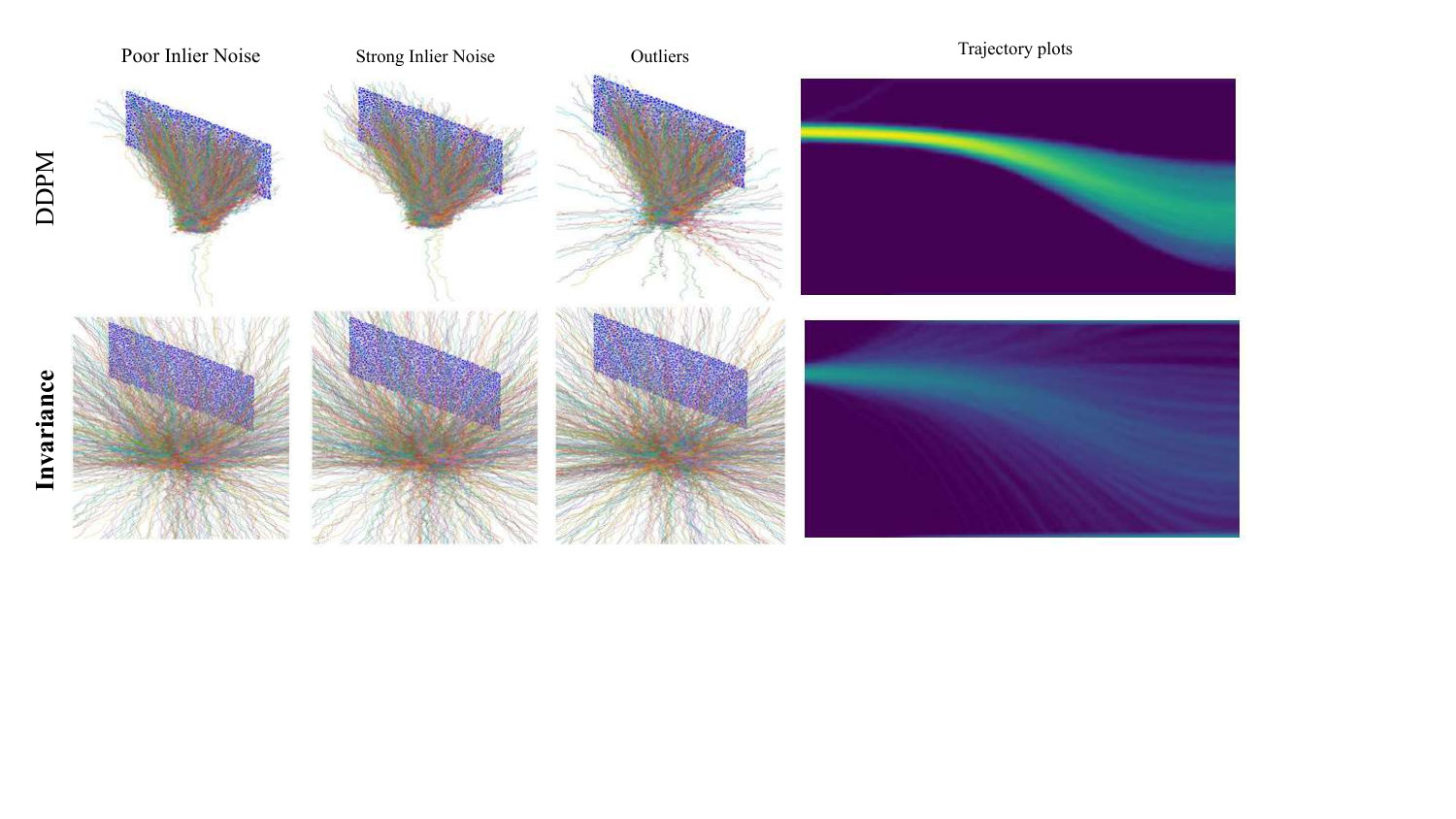}
        \put(6, 45){\tiny{\textbf{Poor inlier noise}}}
        \put(26, 45){\tiny{\textbf{Strong inlier noise}}}
        \put(50, 45){\tiny{\textbf{Outliers}}}
        \put(75, 45){\tiny{\textbf{Model trajectories}}}
        \put(-2, 29 ){\rotatebox{90}{DDPM \cite{ho2020denoising}}}
        \put(-2 , 10){\rotatebox{90}{\textbf{Inv}$_{\text{adv}}$}}

    \end{overpic}
    \caption{The application of invariance on 3D data highlights the incorrect behavior of the training procedure: the learnt data distribution is completely different from the reference one.}
    \label{fig:inv-3d-plots}
\end{figure}
A comparison between the plots in \cref{fig:inv-3d-plots} and those in \cref{fig:heatmaps} further emphasizes the benefits of adversarial training with equivariance. The contrast shown in the compared trajectories clearly illustrates how our approach consistently produces trajectories that are more clustered, sharper, and better aligned with the underlying data manifold, thereby reinforcing the inadequacy of conventional adversarial training methods.

\subsubsection{Notes on definition of the adversarial perturbation} 
One of the main points of our work is defining a suitable perturbation $\pert$ for unconditional diffusion models that aims at disrupting generation trajectories without relying on acting on the model's inputs.
In order to craft this kind of attack, we focused on exploiting generation dynamics in order to correctly perturb it at each of its steps.
Inspired by adversarial attacks with random start (such as R-FGSM \cite{wong2020fast}),  $\pert$ is first initialized by randomly sampling from a uniform distribution, whose bounds are $[-r_\beta(\cdot),  r_\beta(\cdot)]$.
The initialization distribution is chosen to be a uniform distribution with varying bounds but always centered at zero. This choice ensures that the perturbation has zero mean, which is essential when applied within the diffusion process. A non-zero mean would not only bias the estimation of the noise but also violate the Gaussian transition assumption, which requires the noise to be zero-centered.
The parameter $\beta$ is sampled from uniform distribution as follows  $\beta \sim \mathcal{U}[0.5,2]$. Its aim is to enhance the model's robustness to trajectory deviations by randomly varying the perturbation's bounds. 
Once the perturbation bounds are defined, it is straightforward to calculate the standard deviation of the initialization distribution. The mean value of $\pert$ is given by:
$$\mathbb{E}[\pert] = \frac{[(-r_\beta(\cdot)) + (r_\beta(\cdot))]}{2} = 0.$$
The variance of the distribution is defined as:
$$\mathbb{VAR}[\pert] = \frac{[(-r_\beta(\cdot)) + (r_\beta(\cdot))]^2}{12} = \frac{ (2r_\beta(\cdot))^2}{12}  = \frac{ r_\beta(\cdot)^2}{3} $$
The variance is consequently defined as a $\beta$-dependent quantity as it is rescaled batch-wise by this parameter, assuring a random dynamic change of the perturbation bounds.

\newpage
\subsection{Comprehensive analysis of the diffusion flow dynamics}\label{sec:qual-diff}
\begin{figure}[p]
    \centering
    \begin{overpic}[width=\columnwidth]{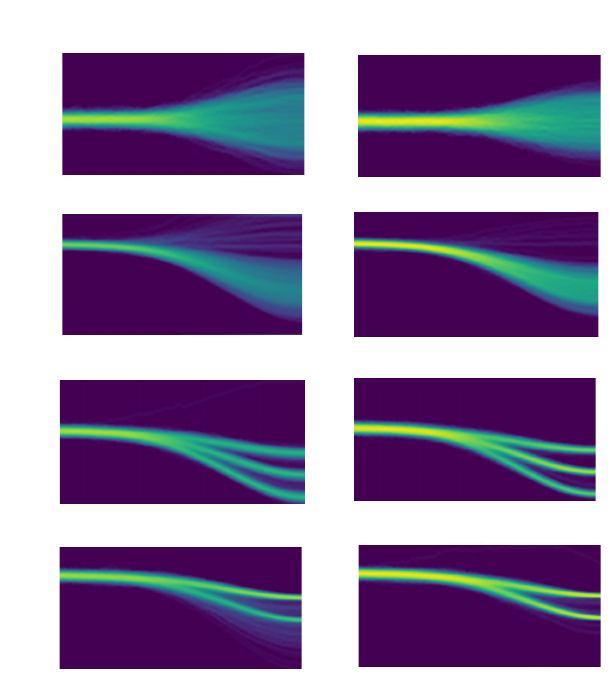}
    \put(66,95){\robust{adv}}
    \put(22,95){{DDPM ~\cite{ho2020denoising}}}
    \put(5.5,80){\rotatebox{90}{{\small Inlier noise}}}
    \put(5.5,32.5){\rotatebox{90}{{\small Inlier noise}}}
    \put(5.5,56){\rotatebox{90}{{\small Outlier noise}}}
    \put(5.5,8){\rotatebox{90}{{\small Outlier noise}}}
    \put(0,66){\rotatebox{90}{\texttt{oblique-plane}}}
    \put(0,20){\rotatebox{90}{\texttt{3-gaussians}}}
    \end{overpic}
    \caption{Diffusion flow: DMs vs \robust{adv}. Left column shows the results by~\citet{ho2020denoising} under two different types of noise. Regular training tends to incorporate the noise inside the diffusion flow, making it more prone to generate undesirable and unexpected results; Right column is \robust{adv} that trades off variability for resilience. Indeed, heatmaps on the right are more concentrated, clear, and less faded.}
    \label{fig:heatmaps}
\end{figure}

In order to better understand data behavior during the generation procedure, we report in this section further trajectory plots. The plots can only be visualized if the data taken into account is low-dimensional in order to properly track points' behavior in the generation. We exploit the low-dimensional datasets proposed in the paper to further investigate trajectory behavior. Additional qualitative samples of the Diffusion Flow are shown in \cref{fig:heatmaps}, supplementing Fig. 4 and Fig. 6, which can be found in the main paper.
Unlike this one, here we have the chance to show also the difference between models' behavior when data distribution is affected by strong inlier noise and outliers on both unimodal distribution \texttt{oblique-plane} and \texttt{3-gaussians}.

The plot shows that even though the DDPM model reaches the distribution of the final part of the trajectories, those are sparse and, even in the case of inlier noise, they appear not to be densely clustered, with some completely diverging from the data distribution. When applying the regularization, particularly in this case, the model is trained with adversarial noise, the density increases in the trajectories, defining more sharp and clustered paths, strongly discouraging significant deviations from their central modes.
This feature is especially useful when the initial data distribution is noisy, as it helps the model avoid learning erroneous points that stray from the true data distribution, preventing it from capturing the noise present in the starting data. In particular, when outlier noise is present, regularization minimizes its influence, resulting in denser and sharper trajectories that better align with the true data distribution clusters.

\subsection{Additional qualitative samples under multiple settings} \label{sec:qual-cifar}
\subsubsection{Trained on clean CIFAR-10 with $p=0\%$}

\begin{figure}[p]
    \centering
    \begin{overpic}[width=\textwidth]{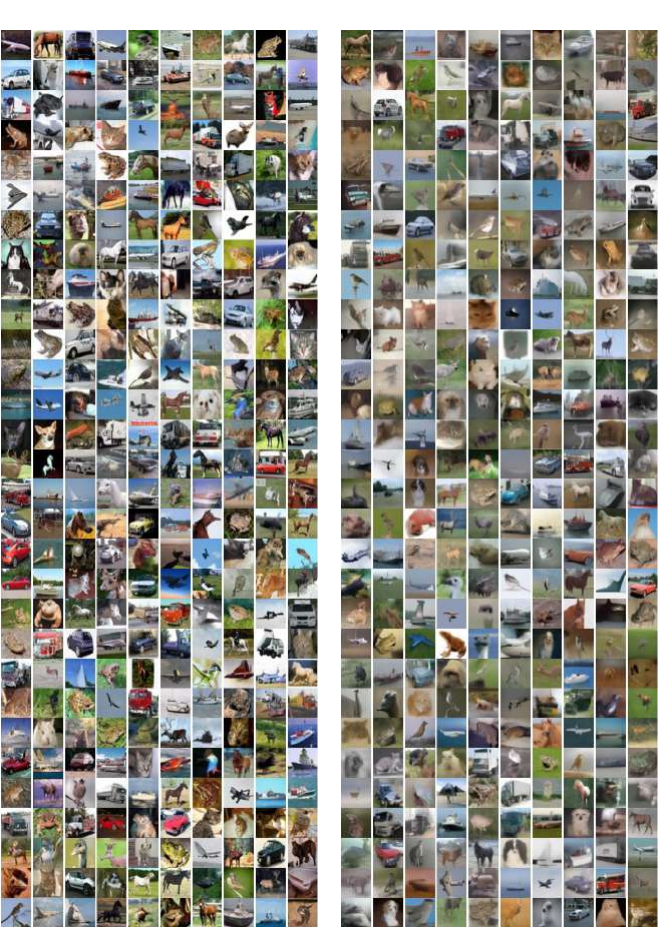}
     \put(25,99.5){{\large clean CIFAR-10 with $p=0\%$}}
     \put(10,97.25){{ DDPM~\cite{ho2020denoising}}}
     \put(52,97.25){\robust{adv}}
    \end{overpic}
    \caption{{Trained on clean CIFAR-10 with $p=0\%$}. Despite the FID decreases once trained on clean data, generated images by \robust{adv} look smooth, and the clutter in the background has been canceled.}
    \label{fig:qual-clean}
\end{figure}

\minisection{DDPM vs \robust{adv}} In \cref{fig:qual-clean} of this supplementary material, we extend Fig. 4 in the paper and show 300 samples from DDPM vs 300 samples from \robust{adv}, both trained on the original dataset. 
Although our method has not been designed to work directly with uncorrupted data, the images that ours generates result in smooth images, the clutter in the background has been canceled, yet the objects and animals are still clearly recognizable, and part of the noise in the background of CIFAR-10 has been removed.
We think that it is reasonable to justify the drop we have in the FID with our method denoising action, which is, for example, the removal of part of the characteristic background noise proper of CIFAR-10. This effect can be the reason for the evaluation penalizing us.

\begin{figure}[p]
    \centering
    \begin{overpic}[width=\textwidth]{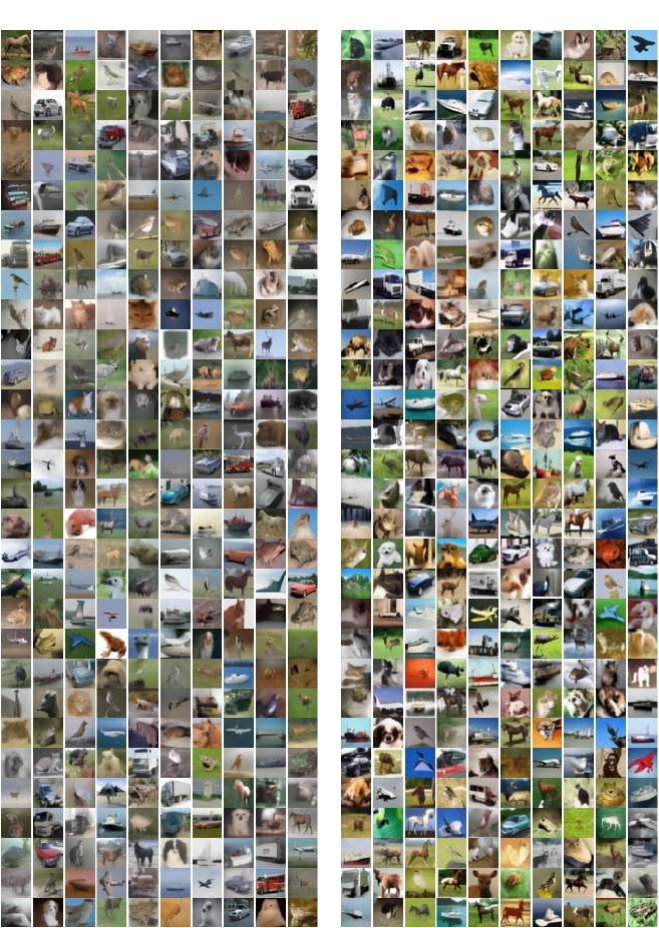}
     \put(25,99.5){{\large clean CIFAR-10 with $p=0\%$}}
     \put(6,97.25){\robust{adv} w/ 1000 steps (28.68 FID $\downarrow$)}
     \put(42,97.25){\robust{adv} w/ 500 steps (\tbf{24.34} FID $\downarrow$)}
    \end{overpic}
    \caption{{Trained on clean CIFAR-10 with $p=0\%$ but comparing less steps ($500$) vs the default DDPM scheduler used for training ($1000$)}. Although we run \robust{adv} with a scheduler with fewer steps ($500$) and do not use it in training, the images on the right with $500$ steps have better FID than with the original scheduler on the left.}
    \label{fig:qual_clean_500_1000_steps}
\end{figure}

\minisection{500 vs 1000 steps} We expand the current section by including some samples that focus on enriching the paper's discussion about faster sampling. In \cref{fig:qual_clean_500_1000_steps} we offer on the left the results by \robust{adv} with $1000$ inference steps trained on uncorrupted data. On the right instead, we show the qualitative samples still with \robust{adv} yet using a scheduler with $500$ inference steps, thereby cutting $50\% $ of the inference time.
\emph{Surprisingly, the faster sampling yields better FID. We get 28.68 FID with $1000$ steps and $24.34$ with $500$ steps.} In terms of differences, taking more steps generates images with warmer and natural colors, whereas taking fewer steps seems to improve the details of the objects, and the colors look brighter and saturated, probably being closer to the actual CIFAR-10 images.

\subsubsection{Trained on noisy CIFAR-10 with $p=90\%,~\sigma=0.1$}
\begin{figure}[p]
    \centering
    \begin{overpic}[width=\textwidth]{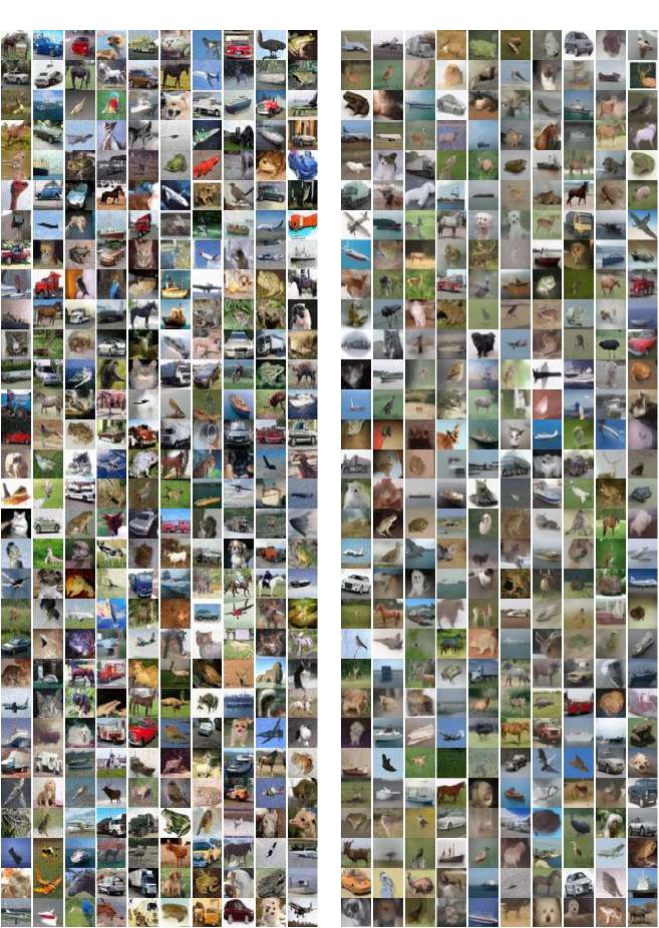}
    \put(18,99.5){{\large noisy CIFAR-10 with $p=90\%,~\sigma=0.1$}}
     \put(10,97.25){{ DDPM~\cite{ho2020denoising}}}
     \put(52,97.25){\robust{adv}}
    \end{overpic}
    \caption{{Trained on noisy CIFAR-10 with $p=90\%,~\sigma=0.1$}. Despite added noise, \robust{adv} images look smooth, and the clutter in the background has been canceled along with the Gaussian noise added. Instea,d DDPM on the left propagates the noise back in the output.}
    \label{fig:qual-p09s01}
\end{figure}

In \cref{fig:qual-p09s01} of this supplementary material, we provide additional figures not present in the main paper. The figures show $300$ samples from DDPM vs \robust{adv} both trained on noisy CIFAR-10 with $p=90\%,~\sigma=0.1$.
The images that ours generates (right) are smooth, similar to the one in~\cref{fig:qual-clean}, inheriting the smoothing effect of the previous setting. In this case, the smoothing action helps absorb the Gaussian noise present in the dataset. This results in improved performance: unlike DDPM (left), ours is able to unlearn the noise and keep images still with natural colors.

\subsubsection{Trained on noisy CIFAR-10 with $p=90\%,~\sigma=0.2$}
\begin{figure}[p]
    \centering
    \begin{overpic}[width=\textwidth]{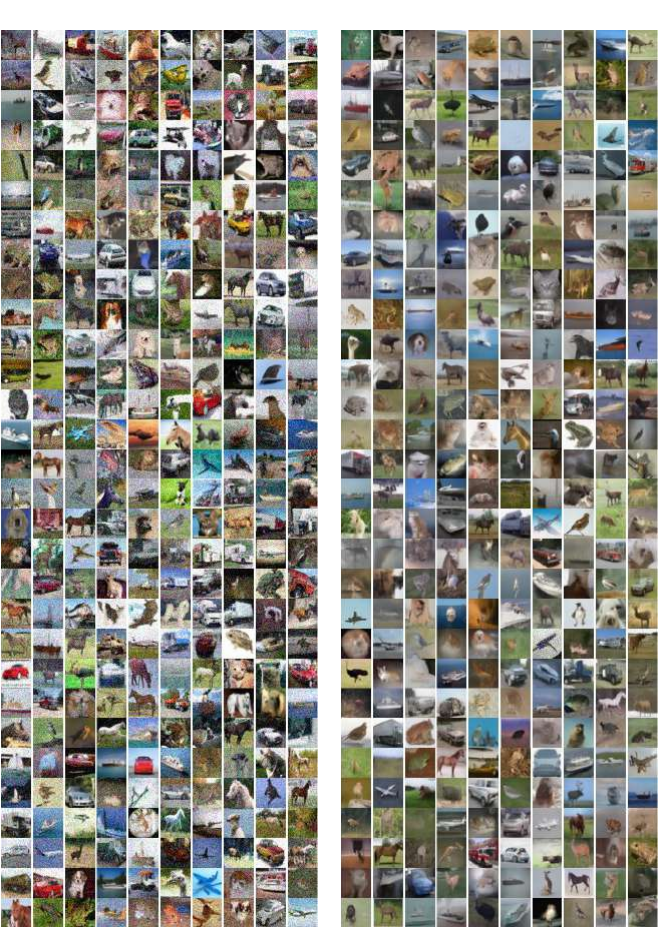}
    \put(18,99.5){{\large noisy CIFAR-10 with $p=90\%,~\sigma=0.2$}}
     \put(10,97.25){{ DDPM~\cite{ho2020denoising}}}
     \put(52,97.25){\robust{adv}}
    \end{overpic}
    \caption{{Trained on noisy CIFAR-10 with $p=90\%,~\sigma=0.2$}. Despite added noise, \robust{adv} images look smooth and the clutter in the background has been canceled along with the Gaussian noise added. Instead DDPM on the left propagates the noise back in the output.}
    \label{fig:qual-p09s02}
\end{figure}

In \cref{fig:qual-p09s02} of this supplementary material we extend Fig.~8 of the paper, enriching it with $300$ more samples per method yet trained on noisy CIFAR-10 with $p=90\%,~\sigma=0.2$. Looking at the standard deviation of the added noise, in this case $\sigma=0.2$ represents a very strong one: it means we are adding $40\%$ of the variability that is naturally present in CIFAR-10, being $\sigma_{\text{data}}=0.5$.
Despite the strong ambient noise, the images that ours generates (\textit{right}) are smooth similar to the one in~\cref{fig:qual-clean} and presence of the strong Gaussian is very rare. Unlike DDPM (\textit{left}), ours is able to unlearn the noise and keep images still with natural colors. 


\subsubsection{Trained on noisy Celeb-A with $p=90\%,~\sigma=0.1$}
\begin{figure}[p]
    \centering
    \begin{overpic}[width=\textwidth]{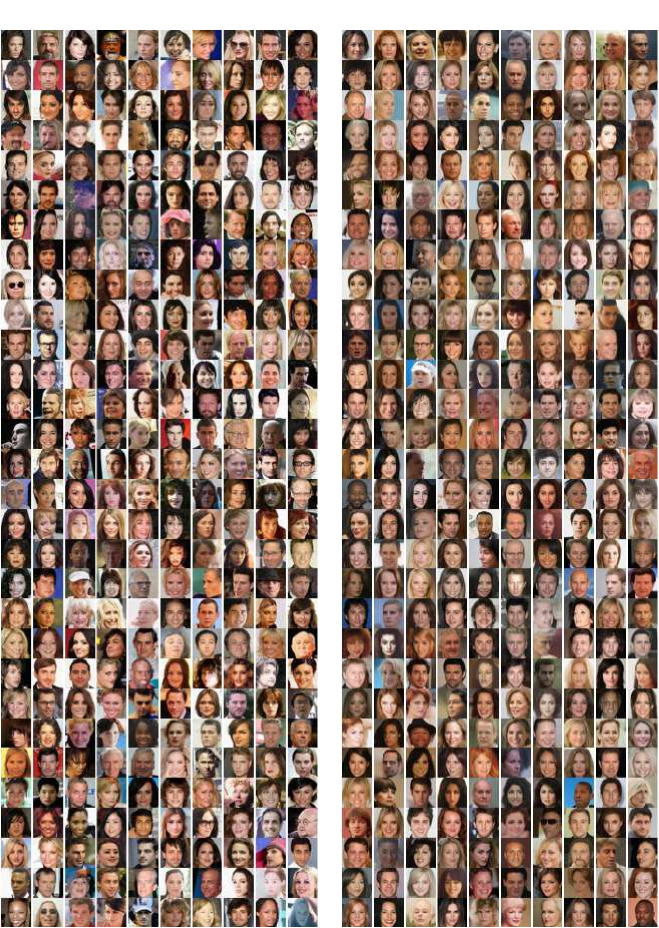}
    \put(18,99.5){{\large noisy Celeb-A with $p=90\%,~\sigma=0.1$}}
     \put(10,97.25){{ DDPM~\cite{ho2020denoising}}}
     \put(52,97.25){\robust{adv}}
    \end{overpic}
    \caption{{Trained on noisy Celeb-A with $p=90\%,~\sigma=0.1$}. Despite added noise, \robust{adv} faces look smooth and the clutter in the background has been canceled along with the Gaussian noise added. Instead, DDPM on the left propagates the noise back in the output.}
    \label{fig:qual-celeb-p09s01}
\end{figure}

We provide a more extensive qualitative analysis on the dataset CelebA~\cite{li2019celeb} in \cref{fig:qual-celeb-p09s01} of this supplementary material. To further motivate the denoising effect, we here show same $300$ samples per method yet trained on noisy Celeb-A with $p=90\%,~\sigma=0.1$. The faces that ours generates ($right$) are smooth, but now instead of absorbing the Gaussian noise present in the dataset, unlike DDPM ($left$), ours is able to unlearn the noise and keep images still with natural colors.

\subsubsection{Trained on noisy Celeb-A with $p=90\%,~\sigma=0.2$}
\begin{figure}[p]
    \centering
    \begin{overpic}[width=\textwidth]{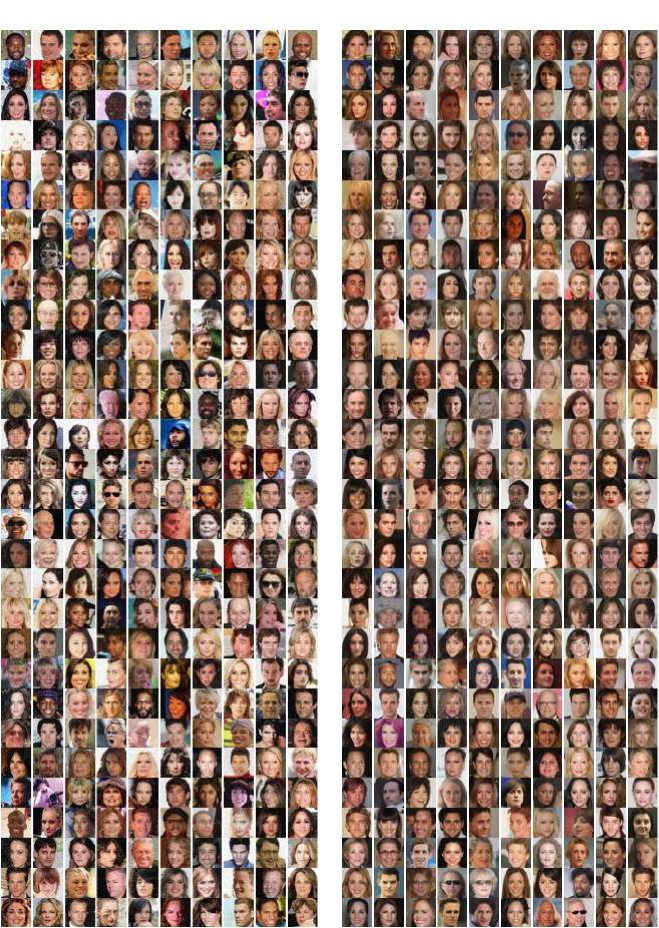}
    \put(18,99.5){{\large noisy CelebA with $p=90\%,~\sigma=0.2$}}
     \put(10,97.25){{ DDPM~\cite{ho2020denoising}}}
     \put(52,97.25){\robust{adv}}
    \end{overpic}
    \caption{{Trained on noisy Celeb-A with $p=90\%,~\sigma=0.2$}. Despite added noise, \robust{adv} faces look smooth and the clutter in the background has been canceled along with the Gaussian noise added. Instead DDPM on the left propagates the noise back in the output.}
    \label{fig:qual-celeb-p09s02}
\end{figure}

We provide a more extensive qualitative analysis on the dataset CelebA~\cite{li2019celeb} in \cref{fig:qual-celeb-p09s02} of this supplementary material. To further motivate the denoising effect, we here show same $300$ samples per method yet trained on noisy Celeb-A with $p=90\%,~\sigma=0.2$. The faces that ours generates ($right$) are smooth but now instead of absorbing the Gaussian noise present in the dataset, unlike DDPM ($left$), ours is able to unlearn the noise and keep images still with natural colors.

\subsubsection{Trained on noisy LSUN Bedroom with $p=90\%,~\sigma=0.1$}
We provide a more extensive qualitative analysis on the dataset LSUN Bedroom~\cite{yu15lsun} in \cref{fig:qual-lsu-early-01-noise} of this supplementary material. To further motivate the denoising effect, we here show same $150$ samples per method yet trained on noisy LSUN dataset with $p=90\%,~\sigma=0.1$. 
The generated images by \robust{adv} (\textit{right}) result to be smoother wrt. to the datasets ones and the DDPM generated ones (\textit{left}) ones, but the smoothing effect allows absorbing the Gaussian noise present in the dataset: unlike DDPM, ours is able to unlearn the noise and keep images still with natural colors.
\subsubsection{Trained on noisy LSUN Bedroom with $p=90\%,~\sigma=0.2$}
We further enrich the qualitative ablation on the dataset LSUN Bedroom~\cite{yu15lsun} in \cref{fig:qual-lsu-early-02-noise} of this supplementary material. To further motivate the denoising effect, we here show same $150$ samples per method yet trained on noisy LSUN dataset with $p=90\%,~\sigma=0.2$. 
The generated images by \robust{adv} (\textit{right}) result to be smoother wrt. to the datasets ones and the DDPM generated ones (\textit{left}) ones, but the smoothing effect allows absorbing the Gaussian noise present in the dataset: unlike DDPM, ours is able to unlearn the noise and keep images still with natural colors.

\begin{figure}[p]
    \centering
    \begin{overpic}[width=.8\textwidth]{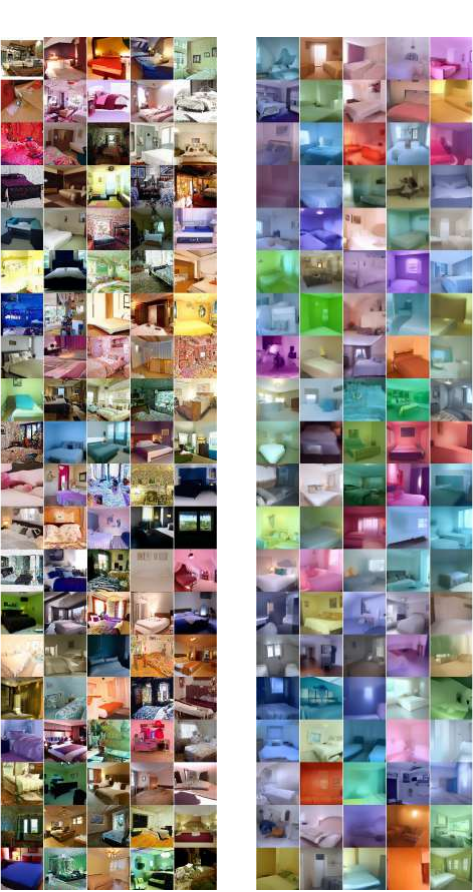}
    \put(15,100.5){{\large LSUN Bedroom early stage training}}
     \put(10,97.25){{ DDPM~\cite{ho2020denoising}}}
     \put(37,97.25){\robust{adv}}
    \end{overpic}
    \caption{{Trained on clean LSUN Bedroom.} Despite the added noise, \robust{adv} produces images that appear smooth and exhibit fewer intricate details. When noise is absent from the training data, this smoothing effect results in the removal of fine-grained information from the learned distribution, ultimately reducing data variability. }
    \label{fig:qual-lsu-early-no-noise}
\end{figure}

\begin{figure}[p]
    \centering
    \begin{overpic}[width=.8\textwidth]{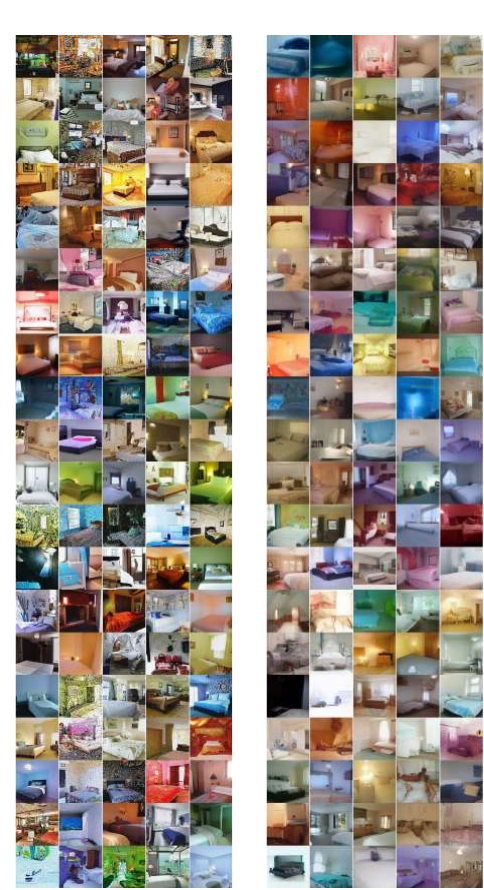}
    \put(3,100.5){{\large LSUN Bedroom with $p=90\%,~\sigma=0.1$, early stage training}}
     \put(10,97.25){{ DDPM~\cite{ho2020denoising}}}
     \put(37,97.25){\robust{adv}}
    \end{overpic}
    \caption{Trained on noisy LSUN Bedroom with $p=90\%,~\sigma=0.1$. Despite added noise, \robust{adv} images look smooth and with fewer intricate details that have been canceled along with the Gaussian noise added. Instead, DDPM on the left propagates the noise back into the output.}
    \label{fig:qual-lsu-early-01-noise}
\end{figure}

\begin{figure}[p]
    \centering
    \begin{overpic}[width=.8\textwidth]{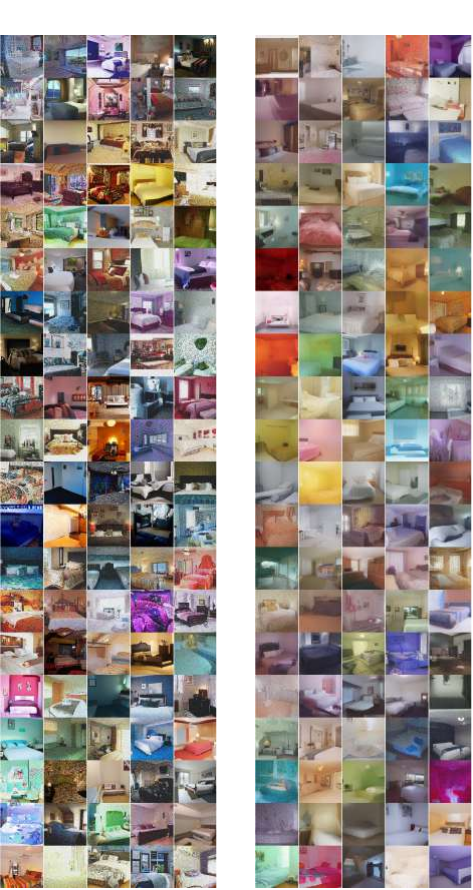}
    \put(3,100.5){{\large LSUN Bedroom with $p=90\%,~\sigma=0.2$, early stage training}}
     \put(10,97.25){{ DDPM~\cite{ho2020denoising}}}
     \put(37,97.25){\robust{adv}}
    \end{overpic}
    \caption{{Trained on noisy LSUN Bedroom with $p=90\%,~\sigma=0.2$}. Despite added noise, \robust{adv} images look smooth and with less intricate details that have been canceled along with the Gaussian noise added. Instead, DDPM on the left propagates the noise back into the output.}
    \label{fig:qual-lsu-early-02-noise}
\end{figure}

\begin{figure}[p]
    \centering
    \begin{overpic}[width=.8\textwidth]{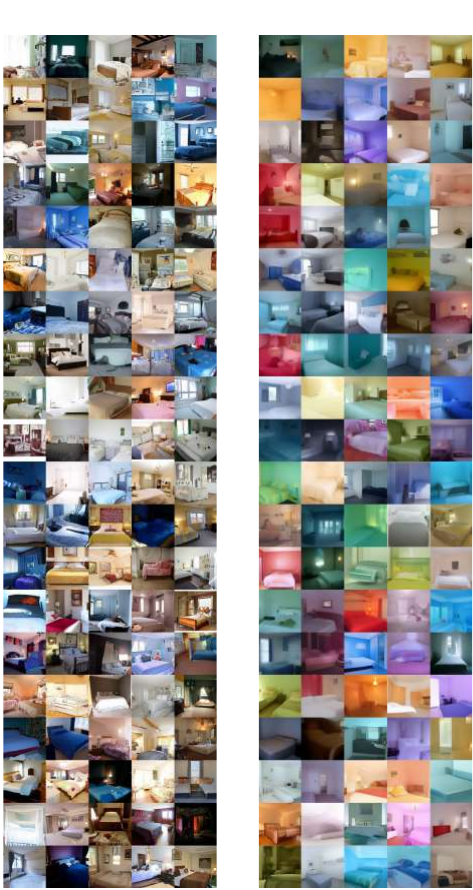}
    \put(15,100.5){{\large LSUN Bedroom late training}}
     \put(10,97.25){{ DDPM~\cite{ho2020denoising}}}
     \put(37,97.25){\robust{adv}}
    \end{overpic}
    \caption{{Trained on clean LSUN Bedroom}. Despite added noise, \robust{adv} images look smooth and with less intricate details, even though more detailed than in earlier stages. }
    \label{fig:qual-lsu-late-no-noise}
\end{figure}

\begin{figure}[p]
    \centering
    \begin{overpic}[width=.8\textwidth]{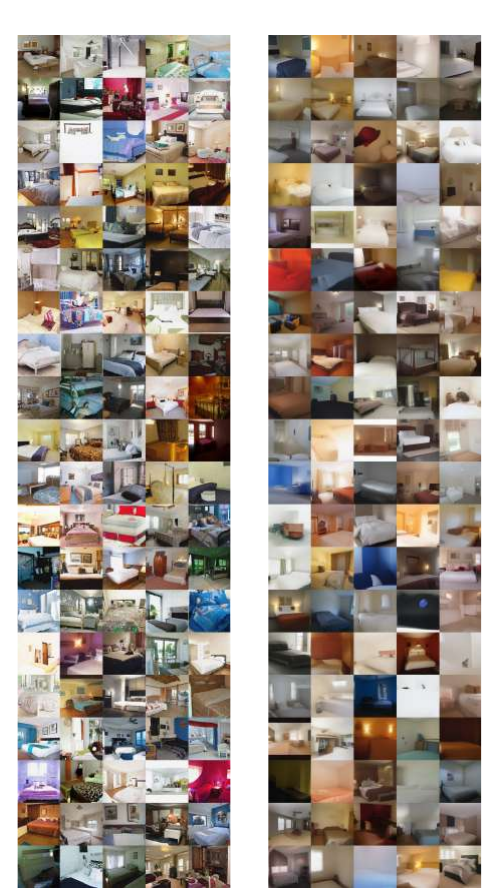}
    \put(3,100.5){{\large LSUN Bedroom with $p=90\%,~\sigma=0.1$, late stage training}}
     \put(10,97.25){{ DDPM~\cite{ho2020denoising}}}
     \put(37,97.25){\robust{adv}}
    \end{overpic}
    \caption{{Trained on noisy LSUN Bedroom with $p=90\%,~\sigma=0.1$.} With extended training, \robust{adv} not only effectively removes the noise introduced into the dataset---in contrast to DDPM---but also restores fine details, resulting in multi-view images with natural colors and enhanced realism.
}
    \label{fig:qual-lsu-late-01-noise}
\end{figure}

\begin{figure}[p]
    \centering
    \begin{overpic}[width=.8\textwidth]{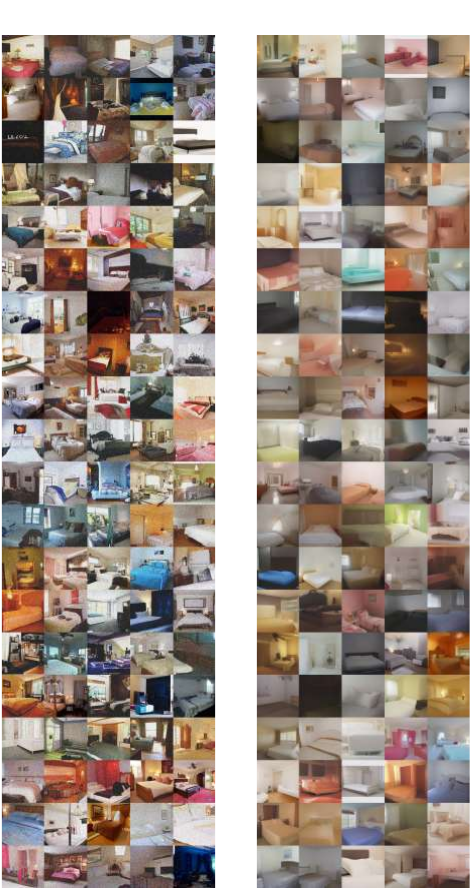}
    \put(2,100.5){{\large LSUN Bedroom with $p=90\%,~\sigma=0.2$, late stage training}}
     \put(10,97.25){{ DDPM~\cite{ho2020denoising}}}
     \put(37,97.25){\robust{adv}}
    \end{overpic}
    \caption{{Trained on noisy LSUN Bedroom with $p=90\%,~\sigma=0.2$}. With extended training, \robust{adv} effectively removes the noise introduced into the dataset, in contrast to DDPM. As a result, it produces cleaner images, albeit with less intricate details.}
    \label{fig:qual-lsu-late-02-noise}
\end{figure}

\clearpage
\subsection{Adversarial training analysis}
\label{sec:analysis}
This method aims at proposing an AT approach to the diffusion model's training whose design choices have been motivated extensively in previous sections as well as in the main paper. In this section, we want to highlight some interesting points we observed during the framework formulation.

\subsubsection{Training dynamics}
\begin{figure}[ht]
\vspace{10pt}
\centering
\setlength{\tabcolsep}{5pt} 
\renewcommand{\arraystretch}{1} 

\begin{tabular}{ccc}
  \begin{overpic}[width=0.3\linewidth]{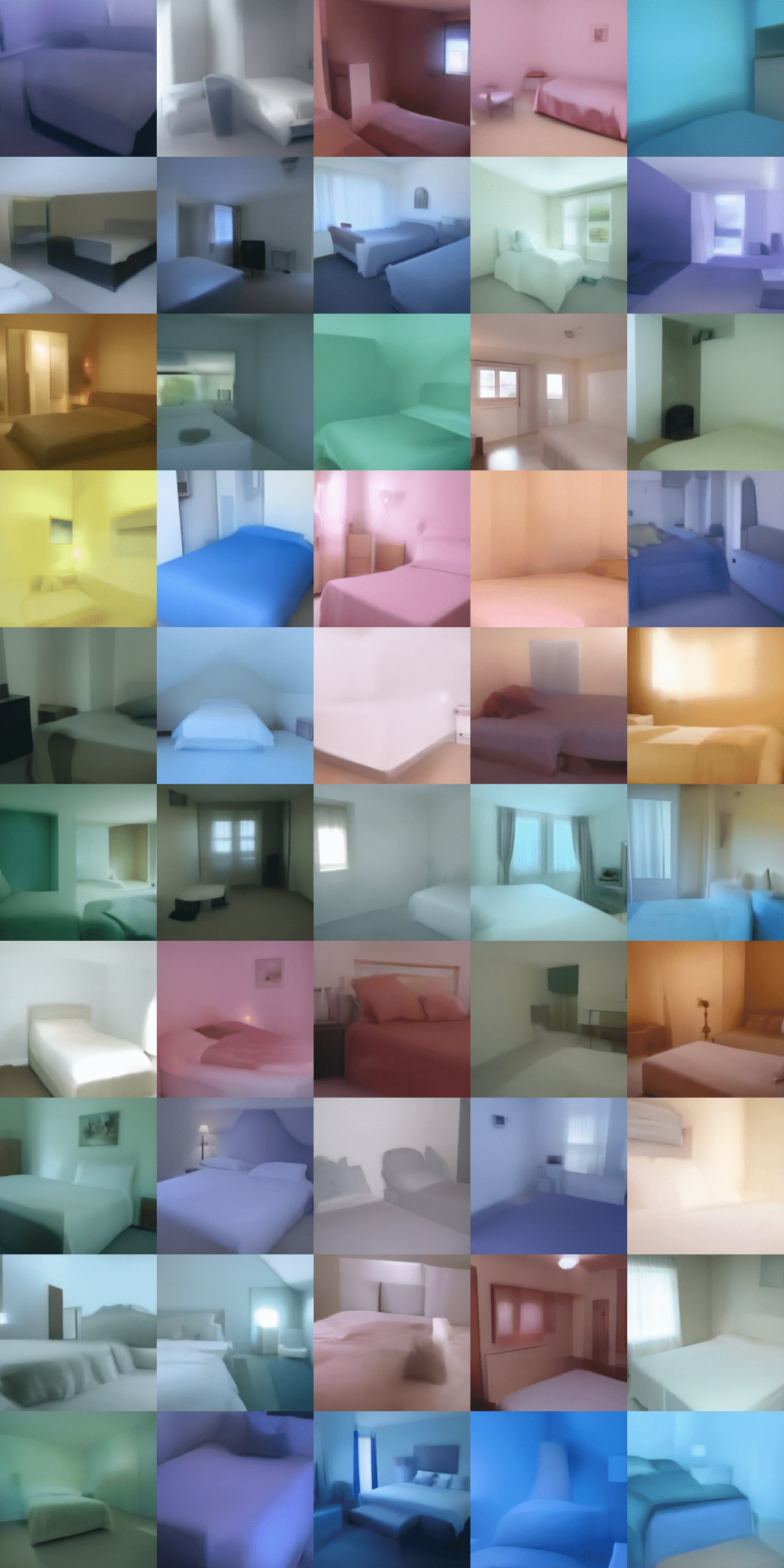}
    \put(13,102){\color{black}\small  LSUN Bedroom}
    \put(-5,35){\rotatebox{90}{\color{black}\small \texttt{early stage training}}}
  \end{overpic} &
  \begin{overpic}[width=0.3\linewidth]{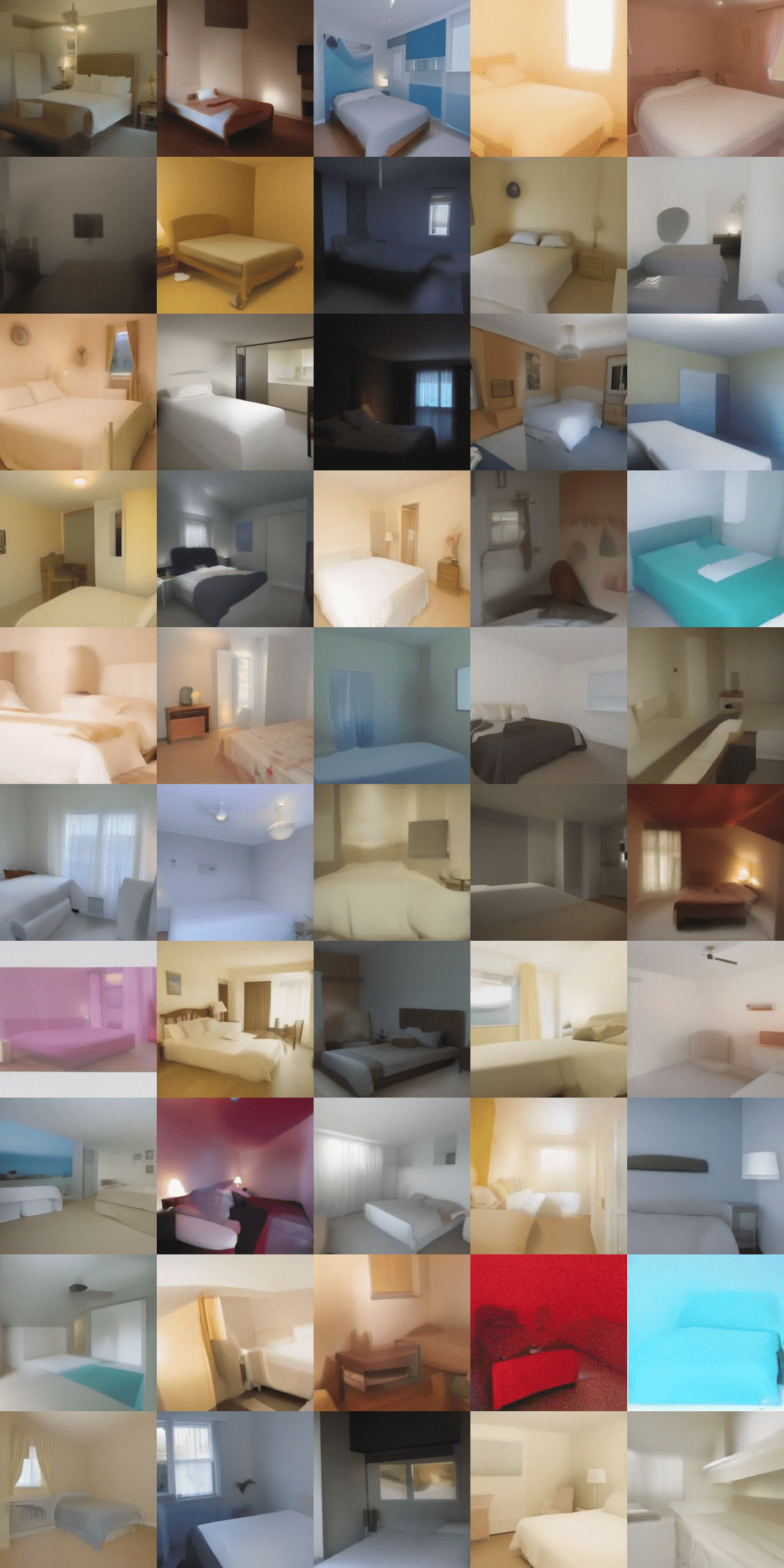}
    \put(5,102){\color{black}\small  LSUN Bedroom $\sigma=0.1$}
  \end{overpic} &
  \begin{overpic}[width=0.3\linewidth]{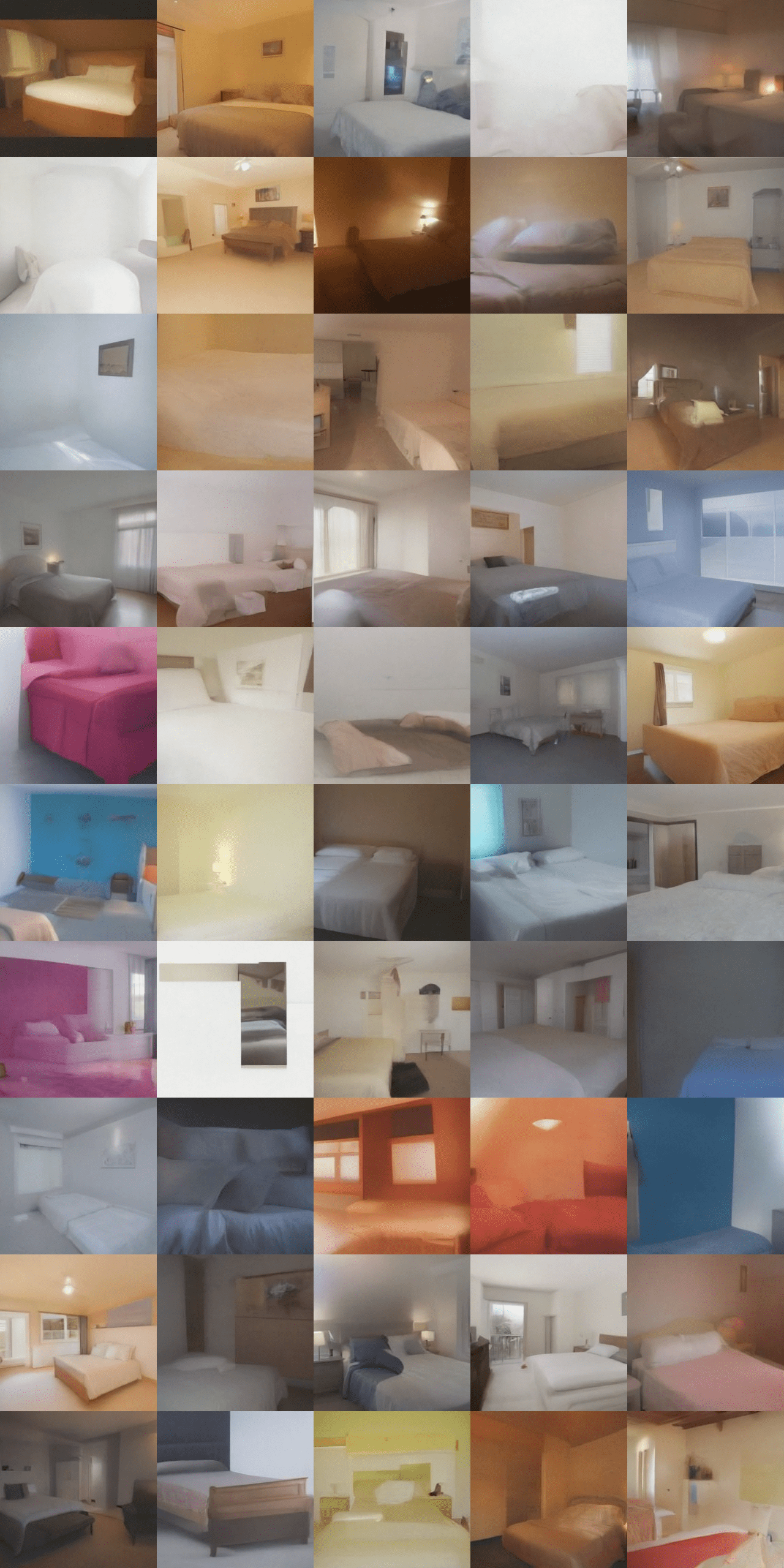}
    \put(5,102){\color{black}\small  LSUN Bedroom $\sigma=0.2$}
  \end{overpic} \\[25pt]
  \begin{overpic}[width=0.3\linewidth]{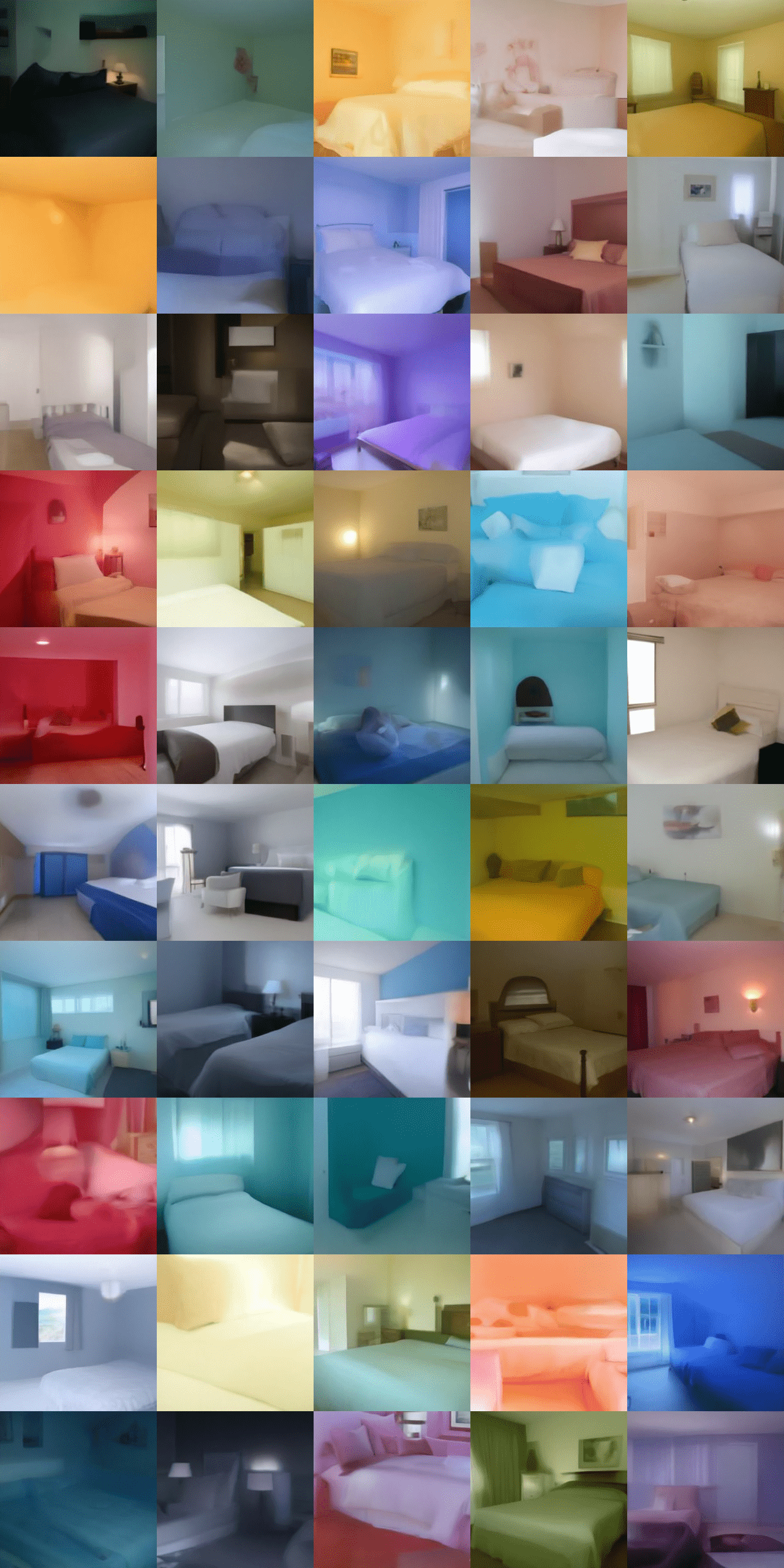}
    \put(13,102){\color{black}\small  LSUN Bedroom}
    \put(-5,35){\rotatebox{90}{\color{black}\small \texttt{late stage training}}}
  \end{overpic} &
  \begin{overpic}[width=0.3\linewidth]{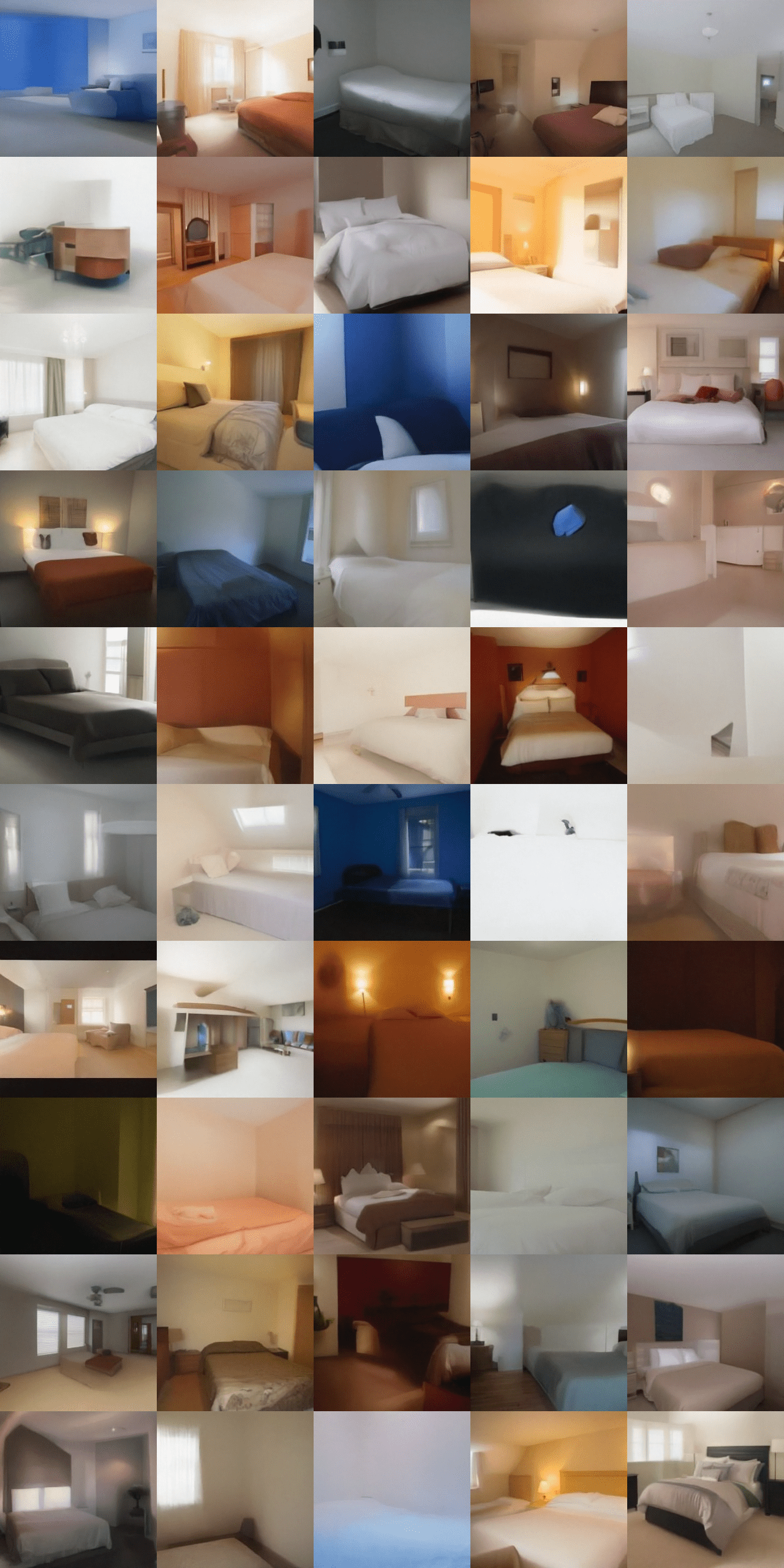}
    \put(5,102){\color{black}\small  LSUN Bedroom $\sigma=0.1$}
  \end{overpic} &
  \begin{overpic}[width=0.3\linewidth]{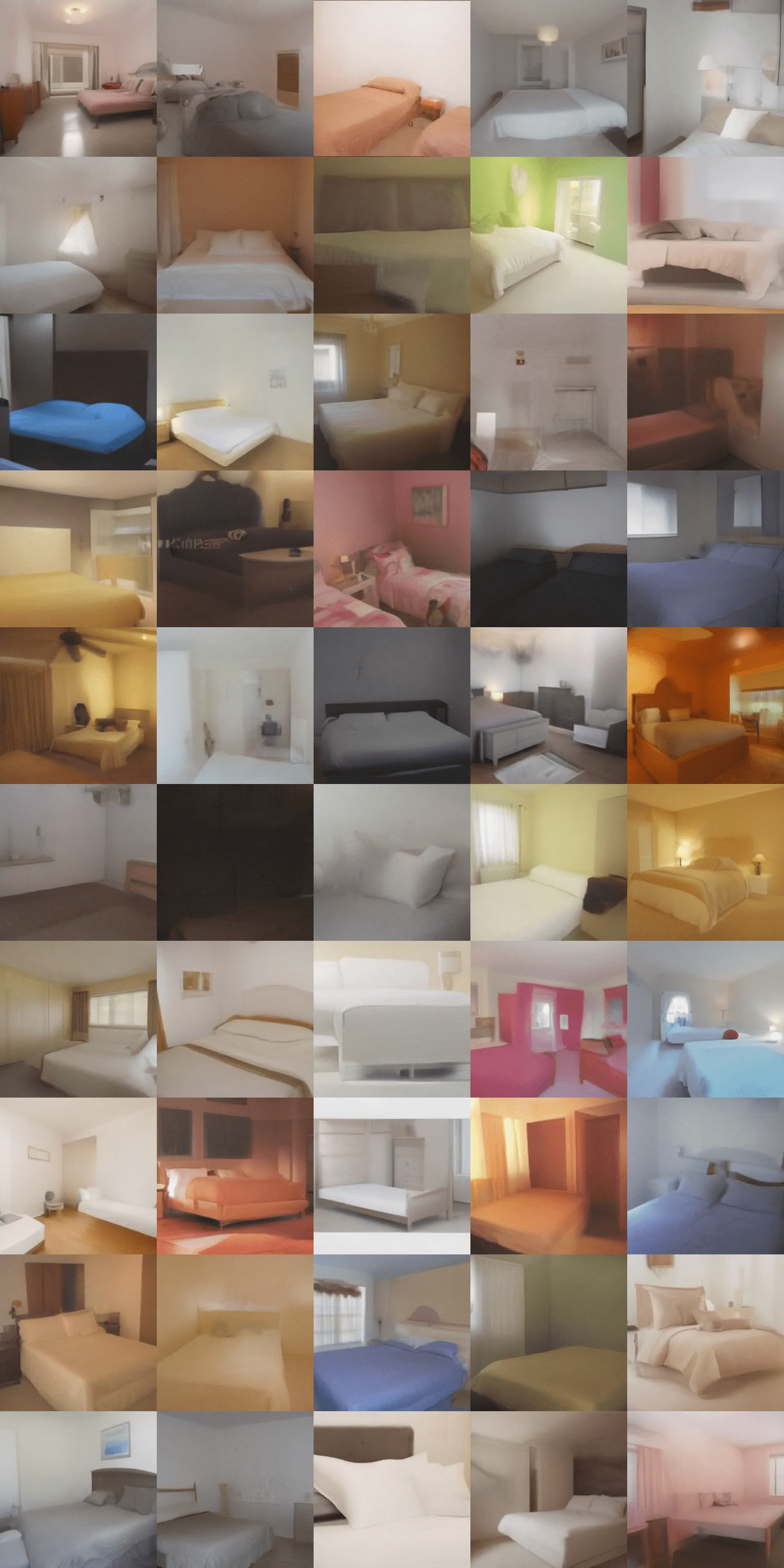}
    \put(5,102){\color{black}\small  LSUN Bedroom $\sigma=0.2$}
  \end{overpic}
\end{tabular}

\vspace{10pt}
\caption{Qualitative results analysis on samples generated by \robust{adv } at different training stages. }
\label{fig:ablation_grid}
\end{figure}

Adversarial training diffusion models inevitably influences DM training dynamics. Indeed, the proposed regularization acts as a smoothing factor for the diffusion process in the trajectory space. In order to evaluate the training dynamics, we propose an ablation on DM generated samples at different training iterations. 
\cref{fig:ablation_grid} is intended to show the evolution of generated samples by AT models at different training iterations. The first row shows samples generated by models trained in an early stage, while the second shows generations from models trained for longer. 
On the right column, the dataset has not been corrupted; the generations, after more training iterations, start losing the bright colors, tending towards more natural-looking colors. Moreover, the generated data starts acquiring its details.
The same effects can be seen for models trained on corrupted data, $\sigma =0.1$ in the middle and $\sigma = 0.2$ on the right (both with $p=0.9\%$). In those cases, it is also possible to see that some generated samples, which at earlier epochs still resulted in being noisy, are completely denoised.
This dynamic suggests that the model first focuses on fitting the overall data model, focusing more on the smoothing effect. Once done, the model goes back to learning the details of the data distribution, including some variability, but still not taking into account the noise present in the data. 
Furthermore, a clearer picture of the training dynamics can be obtained by examining images \cref{fig:qual-lsu-early-no-noise,fig:qual-lsu-early-01-noise,fig:qual-lsu-early-02-noise,fig:qual-lsu-late-no-noise,fig:qual-lsu-late-01-noise,fig:qual-lsu-late-02-noise},  that effectively compare the robust approach with the DDPM model at both early and late stages of training.

\subsubsection{How $\lambda$ in Eq. (9) of the main paper influences the model's denoising capability}
\begin{figure}[ht]
\vspace{10pt}
\centering
\setlength{\tabcolsep}{5pt} 
\renewcommand{\arraystretch}{1} 

\begin{tabular}{ccc}
  \begin{overpic}[width=0.3\linewidth]{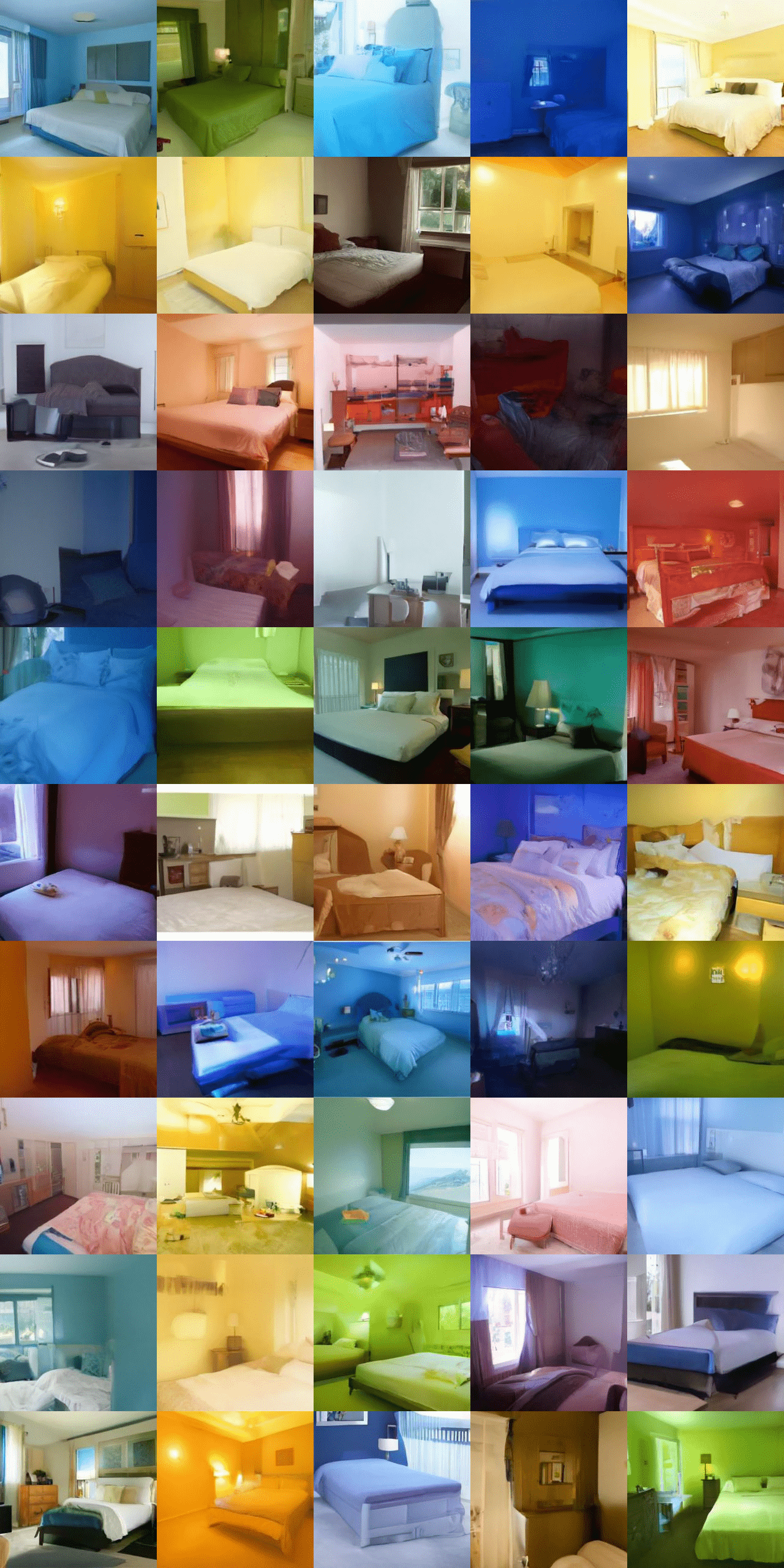}
    \put(13,102){\color{black}\small  LSUN Bedroom}
    \put(-5,45){\rotatebox{90}{\color{black}\small \texttt{$\lambda = 0.1$}}}
  \end{overpic} &
  \begin{overpic}[width=0.3\linewidth]{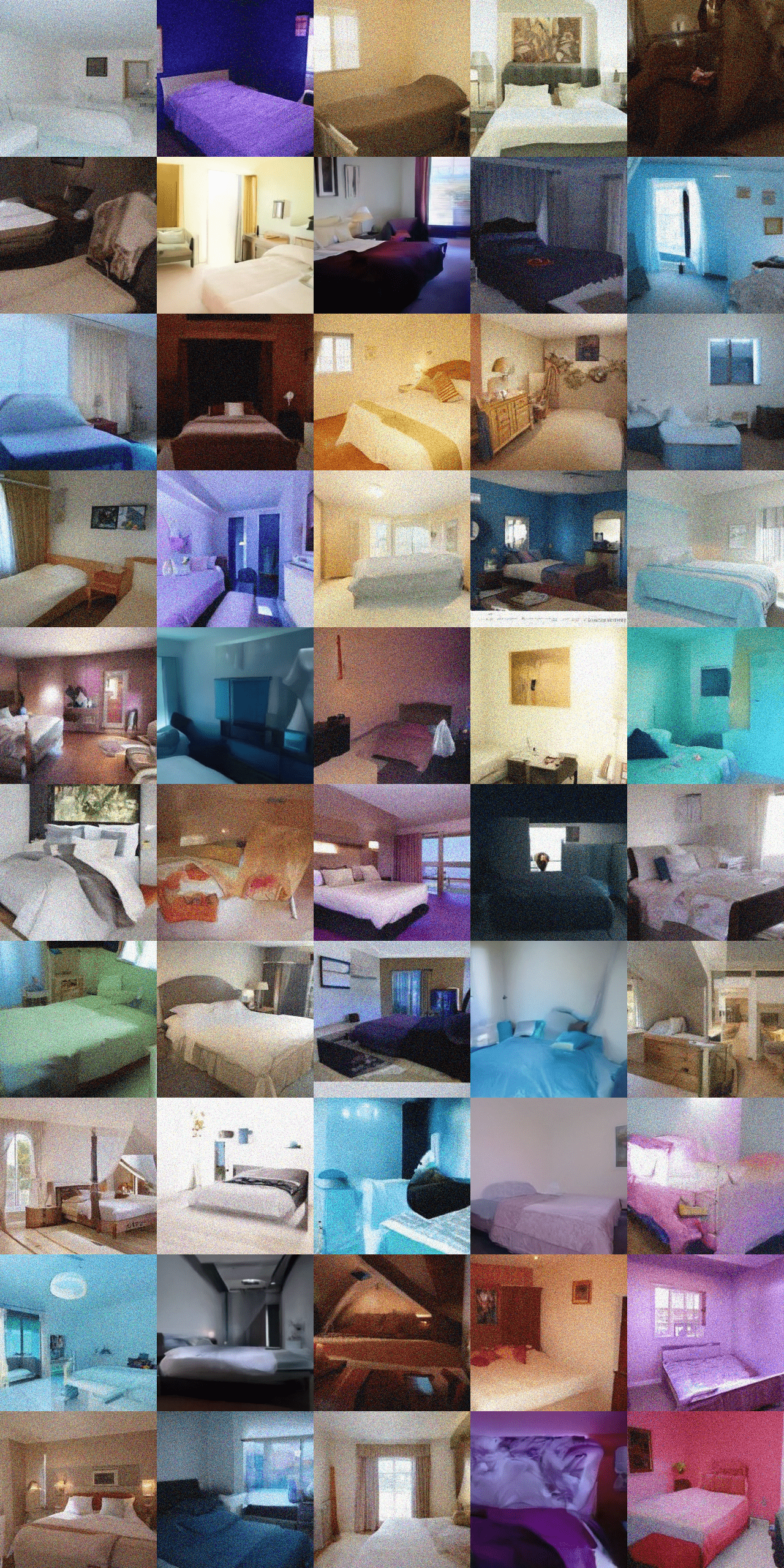}
    \put(5,102){\color{black}\small  LSUN Bedroom $\sigma=0.1$}
  \end{overpic} &
  \begin{overpic}[width=0.3\linewidth]{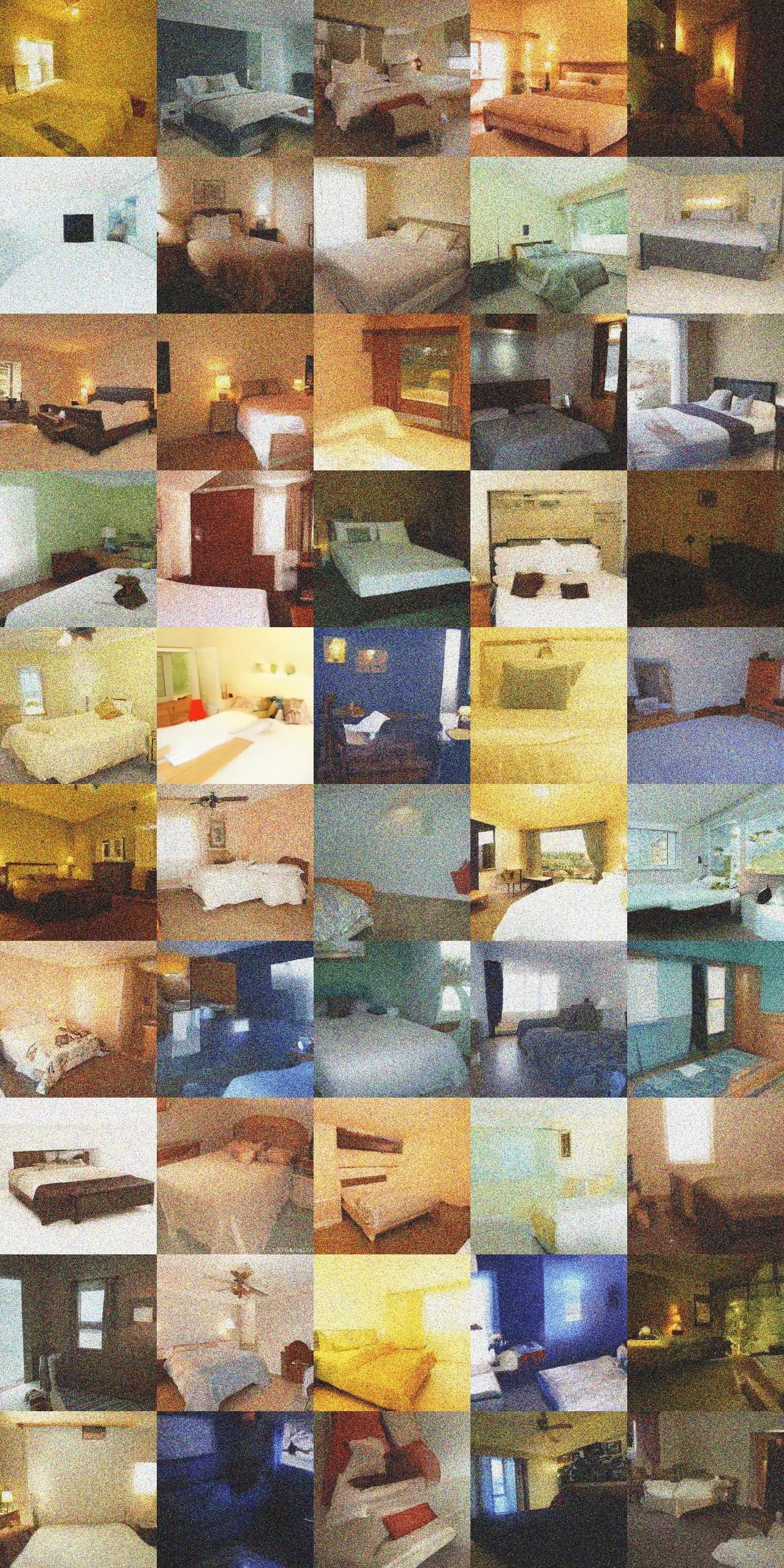}
    \put(5,102){\color{black}\small  LSUN Bedroom $\sigma=0.2$}
  \end{overpic} \\[15pt]
  \begin{overpic}[width=0.3\linewidth]{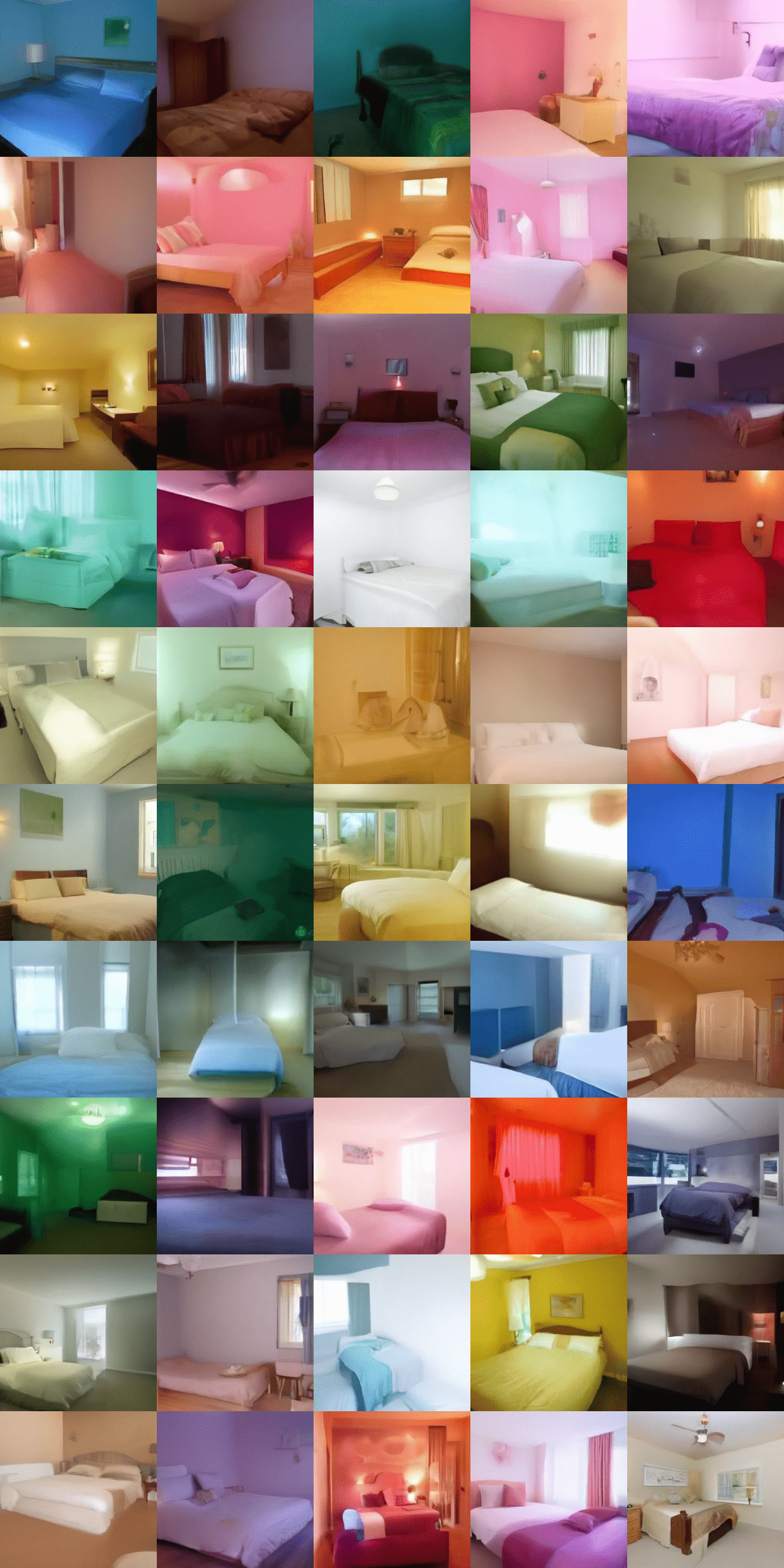}
    \put(-5,45){\rotatebox{90}{\color{black}\small \texttt{$\lambda = 0.2$}}}
  \end{overpic} &
  \begin{overpic}[width=0.3\linewidth]{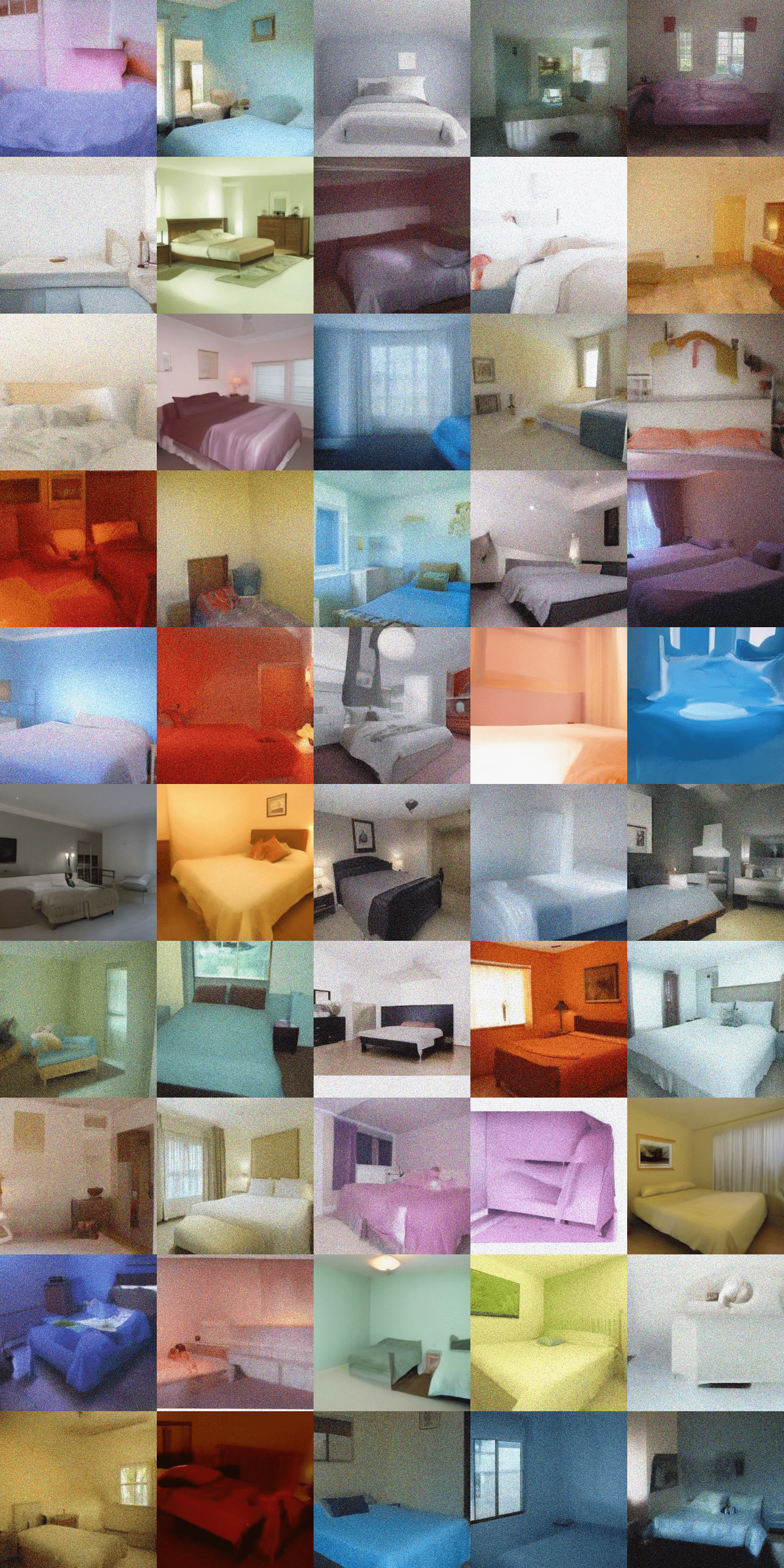}
  \end{overpic} &
  \begin{overpic}[width=0.3\linewidth]{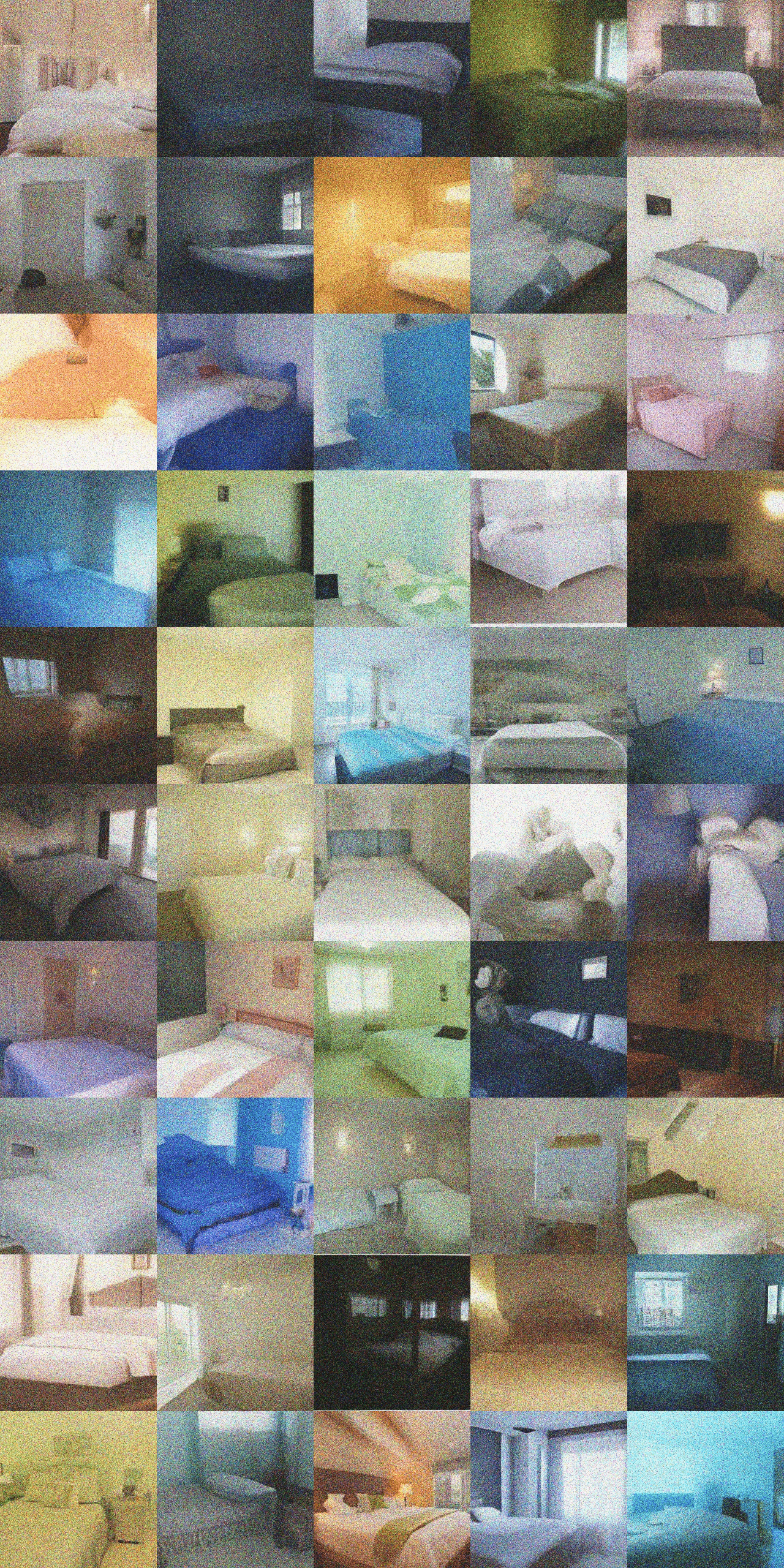}
  \end{overpic}
\end{tabular}

\vspace{10pt}
\caption{\textbf{\robust{adv}} trained on LSUN Bedroom dataset, with different noisy data ($p = 90\%$,  different $\sigma$ are visible in the image). The first row sets the regularization hyperparameter $\lambda$ to $0.1$, the second to $0.2$. }
\label{fig:lambd-ablation_grid}
\end{figure}

In the method section, we stress that the choice of the hyperparameter $\lambda$ heavily influences the model's smoothing ability. To further motivate the previous statement, we provide straightforward evidence of this by observing generated samples produced by different models, with the same architectures and minimum regularization ray among all the shown samples. The varying parameter is $\lambda$, which is set to the values $\{0.1, 0.2, 0.3\}$. 
\cref{fig:lambd-ablation_grid} shows the results at an early training stage of the models. Despite being at an early stage, the $\lambda$ influence in models' performance already appears clear. When the data is not noisy (first column), increasing its value results in oversmoothing data, losing subject details, due to the smoothing factor introduced by the regularization. When the data becomes noisy, the regularization becomes fundamental in learning the correct distribution. In the first row we see that the smoothing action is limited due to the small $\lambda=0.1$, indeed the noise is still present in the generated samples both in $\sigma=0.1$ and $\sigma=0.2$, whereas the noise decreases drastically when increasing $\lambda$ to $0.2$. In fact, the images shown in the bottom row show a minor presence of noise, which is expected to disappear in later training. 
On the other side, the increase of the parameter $\lambda$ also causes a loss of details in the image subjects. This phenomenon is due to the smoothing effect, which not only affects noise but also data variability. This smoothing effect becomes even more apparent when compared to \cref{fig:qual-lsu-early-no-noise,fig:qual-lsu-early-01-noise,fig:qual-lsu-early-02-noise,fig:qual-lsu-late-no-noise,fig:qual-lsu-late-01-noise,fig:qual-lsu-late-02-noise}, all generated with $\lambda = 0.3$. These comparisons further support the previous observations by extending the analysis across different levels of noise and training stages.

\clearpage

{
    \small
    \bibliographystyle{ieeenat_fullname}
    \bibliography{main}
}

\newpage

\end{document}